\documentclass[10pt]{article} 
\usepackage[accepted]{tmlr}

\usepackage{hyperref}

\hypersetup{colorlinks,linkcolor={magenta},citecolor={gray},urlcolor={magenta}} 
\usepackage{url}

\title{A Survey on Generative Modeling with Limited Data,  \\Few Shots, and Zero Shot}

\author{\name Milad Abdollahzadeh\thanks{Equal first-author contribution} \footnotemark[4]
\email milad@betterdata.ai \\
    \addr Singapore University of Technology and Design \\ 
    \addr Betterdata AI, Singapore\\
        \AND
    \name Guimeng Liu\footnotemark[1] \email guimeng\_liu@mymail.sutd.edu.sg  \\
    \addr Singapore University of Technology and Design\\
        \AND
    \name Touba Malekzadeh\thanks{Equal second-author contribution} 
    \email touba\_malekzadeh@sutd.edu.sg \\
    \addr Singapore University of Technology and Design\\
        \AND
    \name Christopher T.H Teo\footnotemark[2] \footnotemark[4]
    \email christopher.teo@sap.com \\
    \addr Singapore University of Technology and Design \\
    \addr SAP, Singapore
        \AND
    \name Keshigeyan Chandrasegaran\footnotemark[2] \footnotemark[4] \email keshik@cs.stanford.edu \\
    \addr Singapore University of Technology and Design\\
    \addr Stanford University\\
        \AND
    \name Ngai-Man Cheung\thanks{Corresponding author} \email ngaiman\_cheung@sutd.edu.sg \\
    \addr Singapore University of Technology and Design\\
}

\def\month{12}  
\def\year{2025} 
\def\openreview{\url{https://openreview.net/forum?id=u7GTHazuRp}} 

\newcommand{\wrt}{\textit{w.r.t.} }
\newcommand{\ie}{\textit{i.e. }}
\newcommand{\eg}{\textit{e.g. }}

\usepackage{amsmath} %
\usepackage{makecell}
\usepackage{graphicx} %
\usepackage{tabularx} %
\usepackage{longtable} %

\usepackage{enumitem}
\setlist[itemize]{leftmargin=*}
\setlist[enumerate]{leftmargin=*}

\usepackage{amsthm}

\usepackage{mathtools}
\usepackage{booktabs}

\theoremstyle{definition}
\newtheorem{definition}{Definition}

\theoremstyle{remark}

\usepackage{tabularx} 
\usepackage{amssymb}

\usepackage{pifont}
\usepackage{bbding} 
\usepackage{multirow}
\usepackage{colortbl}
\usepackage{CJK}
\usepackage{caption} %
\usepackage{subcaption}
\NewDocumentCommand{\rot}{O{45} O{1em} m}{\makebox[#2][l]{\rotatebox{#1}{#3}}}%
\newcommand{\cmark}{\scalebox{0.75}{\CircleSolid}}%
\newcommand{\xmark}{\scalebox{0.75}{\CircleShadow}}%
\newcolumntype{R}[1]{>{\RaggedLeft\arraybackslash}p{#1}}
\usepackage{setspace} 

\newcommand{\D}{\mathcal{D}} 
\newcommand{\Ds}{\mathcal{D}_s} 
\newcommand{\Dt}{\mathcal{D}_t} 
\newcommand{\Cs}{C_{seen}} 
\newcommand{\Cu}{C_{unseen}} 

\begin{document}

\maketitle

\begingroup
\renewcommand{\thefootnote}{\fnsymbol{footnote}}
\footnotetext[4]{Work was done while the authors were at SUTD.}
\endgroup

\begin{abstract}

Generative modeling in machine learning aims to synthesize new data samples that are statistically similar to those observed during training. While conventional generative models such as GANs and diffusion models typically assume access to large and diverse datasets, many real-world applications (\eg in medicine, satellite imaging, and artistic domains) operate under limited data availability and strict constraints. In this survey, we examine 
{\bf
Generative Modeling under Data Constraint (GM-DC)}, which includes limited-data, few-shot, and zero-shot settings. We present a unified perspective on the key challenges in GM-DC, including overfitting, frequency bias, and incompatible knowledge transfer, and discuss how these issues impact model performance.
To systematically analyze this growing field, we introduce two novel taxonomies: one categorizing GM-DC tasks (\eg  unconditional vs. conditional generation, cross-domain adaptation, and subject-driven modeling), and another organizing methodological approaches (\eg  transfer learning, data augmentation, meta-learning, and frequency-aware modeling). Our study reviews over 230 papers, offering a comprehensive view across generative model types and constraint scenarios. We further analyze task-approach-method interactions using a Sankey diagram and highlight promising directions for future work, including adaptation of foundation models, holistic evaluation frameworks, and data-centric strategies for sample selection. This survey provides a timely and practical roadmap for researchers and practitioners aiming to advance generative modeling under limited data. Project website: 
\url{https://sutd-visual-computing-group.github.io/gmdc-survey/}.

\end{abstract}


\section{Introduction}

Generative modeling is a field of machine learning that focuses on learning the underlying distribution of the training samples, enabling the generation of new samples that exhibit similar statistical properties to the training data. Generative modeling has profound impacts in various fields including computer vision \citep{ramesh2022dalle2, karras2020analyzing, brock2019biggan, rombach2022latentdiffusion, peebles2023scalable, guo2025got}, natural language 
processing \citep{yu2017seqgan, gulrajani2017improved, van2017neural, gat2024discrete, nie2025large} and data engineering 
\citep{antoniou2017dagan, karras2020ada, tran2021dag, wang2023diffusiongan, hou2024augselfgan}.
Over the years, significant advancements have been made in generative modeling.
Innovative approaches such as Generative Adversarial Networks (GANs) \citep{goodfellow2014GANs,karras2019style,brock2019biggan,arjovsky2017wasserstein,kang2023scaling, huang2024r3gan}, Variational Autoencoders (VAEs) \citep{kingma2013VAE,vahdat2020nvae,van2017neural}, and Diffusion Models (DMs) \citep{rombach2022latentdiffusion,song2020denoising,dhariwal2021diffusionvsGAN,nichol2021improveddenoisingDIM, peebles2023scalable, esser2024scaling, chandrasegaran2024grafting} have played a pivotal role in enhancing the quality and diversity of generated samples. 
The advancements in generative modeling have fueled the recent disruption in generative AI, unlocking new possibilities in various applications such as image synthesis \citep{nellis-2024, metz-image-2025}, text generation \citep{jamali-2025, metz-text-2025}, music composition \citep{lanre-2025, lamba-2025},
genomics \citep{schiff2024caduceus},
and more \citep{han2024vfusion3d, chen2024morphable}. The ability to generate realistic and diverse samples has opened doors to creative applications and 
novel
solutions \citep{shearing-2025, tong-2025}.

\begin{figure}[]
\centering
    \includegraphics[width=0.88\linewidth]{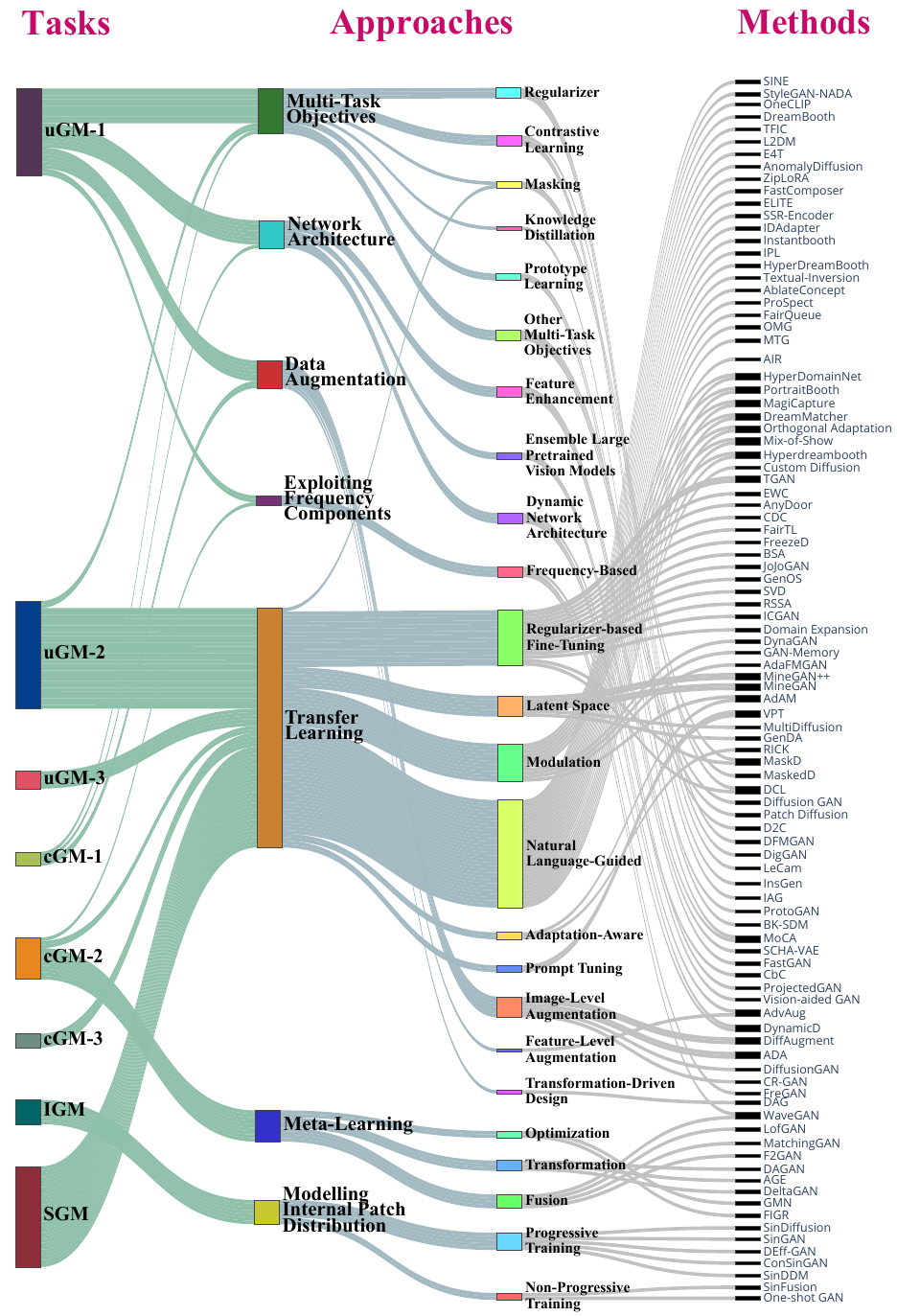}
  \vspace{-3mm}
\caption{{\bf Research Landscape of GM-DC.} 
The figure shows the interaction between GM-DC tasks and approaches (main and sub categories), and
representative
GM-DC methods. Tasks are defined in our proposed taxonomy in Tab.~\ref{tab:tasktaxonomy}, and approaches in our proposed taxonomy in Tab.~\ref{tab:approaches}. 
An interactive version of this diagram is available at our \href{https://sutd-visual-computing-group.github.io/gmdc-survey/}{project website}.
Best viewed in color and with zoom.
}
\label{fig:sankey}
\vspace{-0.5cm}
\end{figure}

Research
on generative modeling has been mainly focusing on setups with sizeable training datasets.
StyleGAN \citep{karras2019style} learns to generate  realistic and diverse face images using
Flickr-Faces-HQ (FFHQ), a high-quality dataset of 70k human face images collected from the photo-sharing website Flickr.
More recent text-to-image generative models and diffusion models (DMs) are  trained on millions of 
 image-text pairs, e.g.
Stable Diffusion \citep{rombach2022latentdiffusion} is trained on
 LAION-400M with 400 million samples \citep{schuhmann2021laion400m}.
However, in many domains (\eg medical), the collection of data samples is challenging and expensive.

{\bf In this paper}, we survey 
Generative Modeling under Data Constraint (GM-DC). 
This research area is important for many domains/ applications where challenges and constraints in data collection exist. We conduct a thorough literature review on learning generative models under limited data (given 50-5000 training samples), few shots (1-50 samples), and zero shot (no samples).  
{\em Our survey is the first to provide a comprehensive overview and detailed analysis of
all types of generative models, tasks, and approaches studied in GM-DC, offering an accessible guide on the research landscape}
(Fig.~\ref{fig:sankey}).
We cover the essential backgrounds, provide detailed analysis of unique challenges of GM-DC, discuss current trends, and present the latest advancements in GM-DC. 

\begin{figure}[!t]
 \centering
     \includegraphics[width=\textwidth]
     {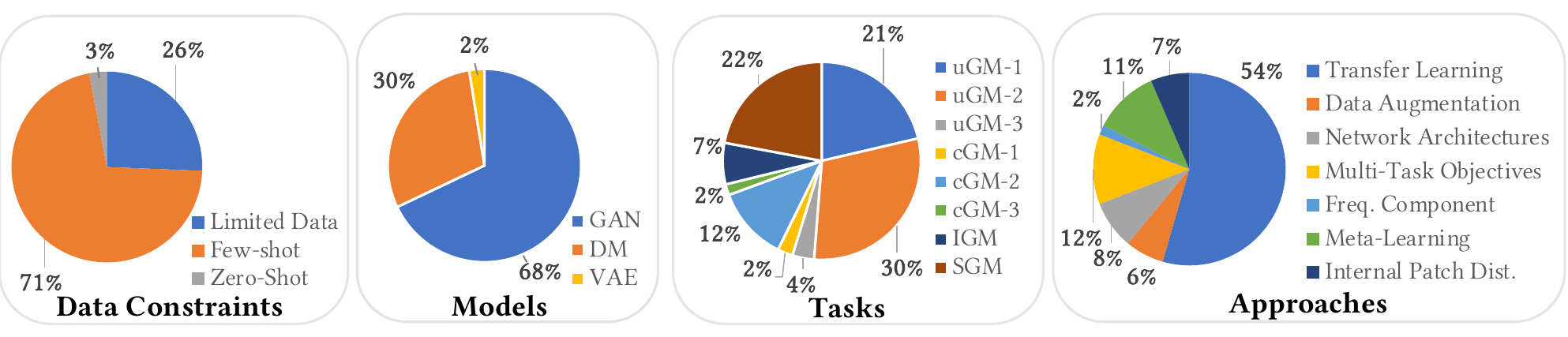}
     \vspace{-0.4cm}
     \caption{
     Analysis of publications in GM-DC. 
     {\bf Data Constraints:} 
     Different types of data constraints studied in GM-DC.
     See Sec.~\ref{sec:background} for more details on setups.
     {\bf Models:} Different types of models  are studied including Generative Adversarial Network (GAN), Diffusion Model (DM), and Variational Auto-Encoder (VAE).
     {\bf Tasks:} Different GM-DC tasks that are studied; See Sec.~\ref{ssec:tasks}, and Tab. \ref{tab:tasktaxonomy} for details on task definitions in our proposed task taxonomy.
     {\bf Approaches:} Different approaches that are applied for addressing GM-DC; More details on our proposed taxonomy of approaches can be found in Sec.~\ref{sec:comprehensive_review} and Tab.~\ref{tab:approaches}. Best viewed in color.
     }
\label{fig:works_statistics}
\vspace{-0.1cm}
\end{figure}

{\bf Our Contributions:}
i) Trends, technical evolution, and statistics of GM-DC (Fig.~\ref{fig:works_statistics};
Fig.~\ref{fig:timeline};
Sec.~\ref{ssec:landscape_analysis}); 
ii) New insights on GM-DC challenges (Sec.~\ref{ssec:challenges});
iii) Two novel and detailed taxonomies, one on GM-DC tasks (Sec.~\ref{ssec:tasks}) and another on GM-DC approaches (Sec.~\ref{sec:comprehensive_review});
iv) A novel Sankey diagram to visualize the research landscape and relationship between GM-DC tasks, approaches, and methods (Fig.~\ref{fig:sankey});
v) An organized summary of individual GM-DC works (Sec.~\ref{sec:comprehensive_review}),
critical analysis, and empirical comparison
(Sec.~\ref{Critical_Analysis_and_Empirical_Comparison}),
vi) Discussion of future directions (Sec.~\ref{ssec:future_direction}).
We further provide a \href{https://sutd-visual-computing-group.github.io/gmdc-survey/}{project website} with an interactive diagram to visualize GM-DC landscape.
Our survey aims to be an accessible guide 
to provide fresh perspectives on the current research landscape, organized pointers to comprehensive literature, and insightful trends on the latest advances of GM-DC.


\begin{figure}[!t]
 \centering
     \includegraphics[width=\textwidth]
     {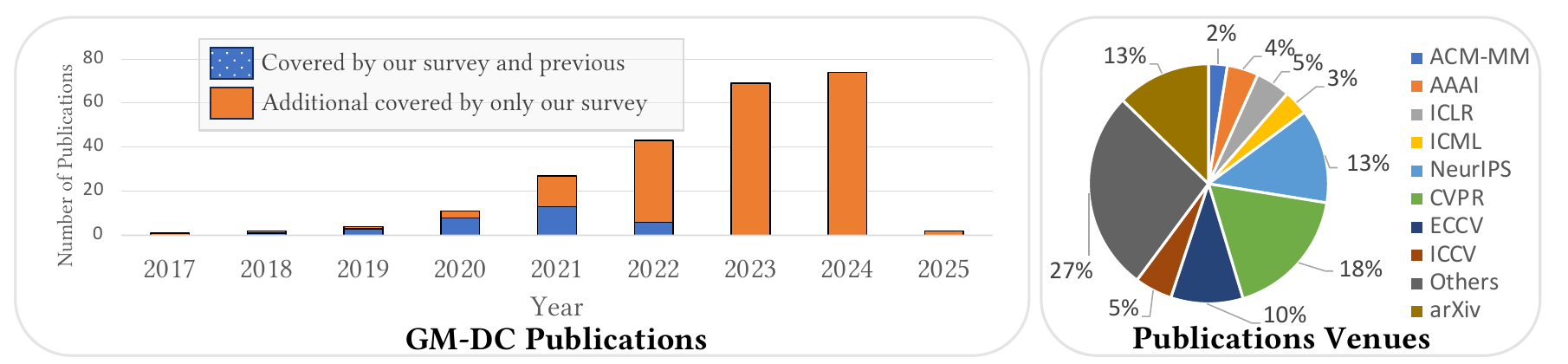}
\vspace{-0.6cm}
 \caption{
 Overall publications statistics in GM-DC.
 {\bf GM-DC Publications (Left):} 
 GM-DC publication trends indicate rising interest in this area.
 We remark that the previous survey \citep{li2022degan} only covers $\sim$13\% of publications discussed in our survey.
 {\bf Publication Venues (Right):} The distribution of publications in major machine learning and computer vision venues, other venues, and arXiv. Best viewed in color.
 }
\label{fig:gm-dc-publications}
\vspace{-0.2cm}
\end{figure}

\subsection{Paper Selection and Search Strategy}

\textbf{Search Strategy.} Our search strategy is based on an extensive list of keywords related to GM-DC. We began with a set of seed keywords derived from well-known papers in the field and selected several representative works. We then carefully examined these papers and their related work, expanding our initial list to form the main set of keywords used to search for relevant research papers for the GM-DC survey. The final list of 17 keywords is as follows: \emph{Transfer learning for generative models}; \emph{Transfer learning in GANs/DMs} (2 keywords); \emph{Fine-tuning generative models}; \emph{Fine-tuning GANs/DMs/VAEs} (3 keywords); \emph{Generative model adaptation}; \emph{Adaptation of GANs/DMs/VAEs} (4 keywords); \emph{Few-shot image generation}; \emph{Few-shot adaptation of generative models}; \emph{Data-efficient generative modeling}; \emph{Training GANs/DMs with limited data} (2 keywords).
To ensure comprehensive coverage, we searched for these keywords across seven major repositories commonly used for machine learning and computer vision research:
Google Scholar,
OpenReview, 
CVF Open Access,
IEEE Xplore Digital Library,
ACM Digital Library,
ScienceDirect (Elsevier), and
SpringerLink.

\textbf{Study Selection Criteria.} Among the collected papers, we applied three main criteria for inclusion, focusing on the problem setup, modality, and task type:
\begin{itemize}
    \item Problem Setup: We examined all experimental results (both qualitative and quantitative) in each paper to ensure that at least one of them satisfies a data-constrained setup ($0 \sim 5000$ available samples), i.e., the proposed approach can operate under such conditions.
    \item Modality: Our survey focuses exclusively on the image modality; therefore, we discarded studies dealing with other modalities such as sculpture, 3D mesh, or point cloud data.
    \item Task Type: We strictly focused on the image generation task and excluded papers addressing other tasks (e.g., image editing).\footnote{Note that some recent works, such as DreamBooth, can handle both image generation and image editing tasks. Since these methods support image generation and not only editing, they are included in our study.}
\end{itemize}

\textbf{Second Round of Search.} After the initial selection, we conducted a second round of review by carefully examining the related work and experimental setups of the selected papers to ensure that all relevant studies were included. In this step, we also added some works that did not explicitly mention the keywords in their title or abstract but nonetheless satisfied all inclusion criteria.

{\bf The rest of the paper is organized as follows.}
In Sec.~\ref{sec:related_work} we discuss related work.
In Sec.~\ref{sec:background} we provide the necessary background.
In Sec.~\ref{sec:taxonomy}, we discuss GM-DC tasks and unique challenges.
In Sec.~\ref{sec:comprehensive_review}, we analyze GM-DC approaches and methods.
In Sec.~\ref{Critical_Analysis_and_Empirical_Comparison}, we summarize, 
for each class of approaches, the key factors contributing to their success or failure, their fundamental limitations, and the deeper insights on their design principles. We further provide empirical comparison.
In Sec.~\ref{sec:discussion}, we discuss open research problems and future directions.
Sec.~\ref{sec:conclusion} concludes the survey.

\section{Related Work}
\label{sec:related_work}

{\bf Discriminative Modeling with Limited Data.} 
A conceptually similar task to GM-DC is discriminative learning under data constraints. Approaches in this research direction aim to learn classification, regression, or even reinforcement learning models using limited and sometimes few-shot data, often through techniques such as meta-learning \citep{finn2017maml, vinyals2016matching, snell2017prototypical} or knowledge transfer from a powerful, pretrained model \citep{radford2021CLIP, sun2019meta, tan2018survey}. This line of research is commonly referred to as few-shot learning for simplicity, and there are several surveys that cover the key concepts and methodologies in this area in considerable detail \citep{wang2020generalizing, hospedales2021meta, gharoun2024meta, zeng2024few, song2023comprehensive}. However, despite some conceptual overlap in addressing the challenge of learning with limited data, these works focus on discriminative tasks, which are fundamentally different from the generative learning tasks discussed in this survey. Consequently, the existing surveys in this domain do not cover the concepts, approaches, or taxonomies that are the focus of this work.

{\bf Generative Modeling.}
A broad line of research in machine learning has focused on developing powerful generative models that can synthesize high-quality and diverse data samples. Key paradigms include variational autoencoders 
\citep{kingma2013VAE,vahdat2020nvae}, generative adversarial models \citep{goodfellow2014GANs, karras2019style, kang2023scaling}, flow-based Models \citep{ho2019flow++, gat2024discrete}, diffusion models \citep{song2020denoising,rombach2022latentdiffusion,esser2024scaling}. Several surveys provide comprehensive overviews of these approaches and their advancements \citep{pouyanfar2018survey, jabbar2021survey, de2023review, hu2023diffusion, bie2024renaissance, li2025comprehensive}. These works highlight the remarkable progress in fidelity, diversity, and controllability. However, they typically assume access to abundant training data, and their performance can degrade significantly when trained with limited samples, manifesting in problems such as overfitting, mode collapse, or reduced generalization. Consequently, while these surveys form the foundation of modern generative modeling, they do not directly address the unique challenges of learning under data-constrained settings, which are the primary focus of this work.

{\bf Generative Modeling with Limited Data.} 
Previous survey on GM-DC \citep{li2022degan} has focused on only a subset of GM-DC papers, studying only  works with  
GANs as a generative model and a subset of technical tasks/ approaches.
Our survey differentiates itself from \citet{li2022degan} in: 
i) Scope  - Our survey is the first to cover all types of generative models and all GM-DC tasks and approaches (Fig. \ref{fig:works_statistics});
ii) Scale -
Our study includes 
233 papers and covers broad GM-DC works, 
while 
previous survey \citep{li2022degan}
covers only $\approx$13\% of works discussed in our survey (Fig. \ref{fig:gm-dc-publications}); 
iii) Timeliness - Our survey collects and surveys the most up-to-date papers in GM-DC; 
iv) Detailedness - Our paper includes detailed visualizations (Sankey diagram, charts) and tables to 
highlight 
interactions and
important attributes of GM-DC literature;
v) Technical evolution analysis - Our paper analyzes the evolution of GM-DC tasks and approaches, providing new perspectives on recent advances;
vi) Horizon analysis - Our paper discusses 
distinctive obstacles encountered in GM-DC and identifies avenues for future research.

\section{Background}
\label{sec:background}

In this section, we first define `domain' and `generative modeling', then we  discuss common approaches of generative modeling and  data constraints studied in 
GM-DC.

\noindent
{\bf Domain.} In this survey, a {\em domain} consists of two components: i) a sample space $\mathcal{X}$, and ii) a marginal probability distribution $P_{data}$, which models the probability of samples from $\mathcal{X}$ \citep{pan2009yang-qiang-transfer}.
This is written as $\mathcal{D}=\{\mathcal{X},P_{data} \}$,
and $x\sim P_{data} \in \mathcal{X}$ denoting a sample in this space. 
An example of a domain is the domain of image of human faces: $\mathcal{D}^{h}=\{\mathcal{X},P_{data}^{h} \}$.
Here $\mathcal{X}$ is the sample space of images, and $P_{data}^{h}$ is the probability distribution of human faces.

\noindent
{\bf Generative Modeling.}
Given a set of training sample $x$ of 
a domain 
$\mathcal{D}=\{\mathcal{X},P_{data} \}$, i.e., 
with an underlying probability distribution $P_{data}$, 
generative modeling  aims to learn to  capture $P_{data}$ ---sometimes also denoted as $P(x)$ in literature.
The result of generative modeling is a 
{\em generative model} $G$ encoding 
a probability distribution $P_{model}$.
The learning objective is to have $P_{model}$ similar to  $P_{data}$ statistically.
After the training, $G$ can generate samples following 
$P_{model}$.
For example, generative modeling with a training set of  human face images 
aims to learn to capture $P_{data}^{h}$, thereby the resulting $G^{h}$ can generate human face images
that are statistically similar to samples from $P_{data}^{h}$.
We also refer to the domain of training samples as {\em target domain}.

\noindent
{\bf Conditional vs Unconditional Sample Generation.}
After learning the underlying distribution of data $P_{data}$, the generative model can generate new samples by sampling from the learned distribution $P_{model}$.
Typically, generation starts with sampling a random vector $z$ ---also called latent code--- as input. Then, this input is passed into the generative model $G$ to transform the latent code into a new sample $G(z) \sim P_{model}$.
Ideally, a good generator is able to capture  the characteristics, quality and diversity of the training dataset, \ie, $P_{model}$ is similar to  $P_{data}$ statistically.
If an additional condition $c$ (like a class label or attribute) is used alongside with the latent code to steer the sample generation towards $c$, the sample generation is called conditional generation:  $G(z,c)$.

\begin{figure}[t]
 \centering
     \includegraphics[width=\textwidth]
     {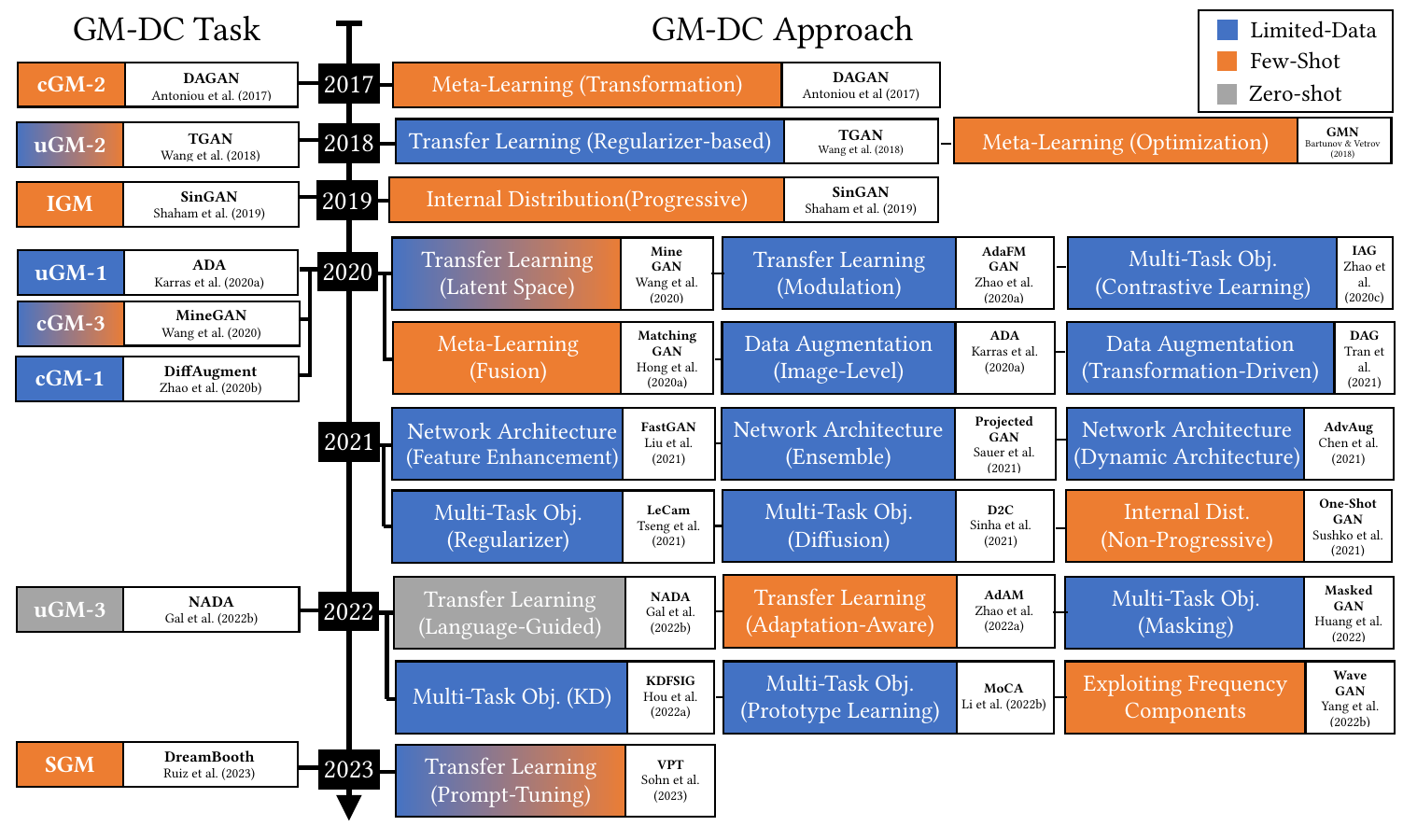}
     \vspace{-0.6cm}
     \caption{Illustration of the {\bf timeline} when a GM-DC task/approach was introduced based on our proposed taxonomies: task taxonomy (details in Sec.~\ref{ssec:tasks}, and Tab.~\ref{tab:tasktaxonomy}), and  approach taxonomy (details in Sec.~\ref{sec:comprehensive_review}, and Tab.~\ref{tab:approaches}). Best viewed in color.
     }
\label{fig:timeline}
\end{figure}
\vspace{-0.2cm}

\subsection{Approaches for Generative Modeling}
\label{ssec:generative_models}

Earlier works on generative models  
study
Gaussian Mixture Models \citep{reynolds2009gaussian}, Hidden Markov Models 
\citep{phung2005topic}, 
Latent Dirichlet Allocation 
\citep{chauhan2021topic}
and Boltzmann Machines \citep{ackley1985learning}.
With the introduction of deep neural networks, recent works study  powerful generative models,
particularly those for image generation, which most GM-DC works focus on.

\noindent
{\bf Variational Auto Encoders (VAE)} 
\citep{kingma2013VAE}.
VAE is a variant of Auto-Encoder (AE) \citep{zhai2018autoencoder}, where both consist of the encoder and decoder networks. AE focuses on dimensional reduction.
The encoder in AE learns to map an input $x$ into a latent (compressed) representation, $z=E(x)$.
Then, the decoder aims to reconstruct the image from that latent representation, $\hat{x}=D(z)$. 
Model parameters are optimized with the following reconstruction loss:
\begin{equation}
    \mathcal{L}_{rec}=||x-D(z)||_2
\end{equation}
AEs are notorious for latent space irregularity making them improper for sample generation \citep{kingma2019introduction}. VAE aims to address this problem by enforcing $E$ to return a normal distribution over latent space.
Assuming a distribution $z\sim \mathcal{N}(\mu,\sigma^2)$ for latent space, this is done by adding the KL-divergence term to the loss function:
\begin{equation}
    \mathcal{L}=||x-D(z)||_2 + KL(\mathcal{N}(\mu,\sigma^2), \mathcal{N}(0,I))
\end{equation} 
Due to the challenges of direct maximization of the likelihood in pixel space, Vector-Quantized VAE (VQ-VAE) 
proposes {\em tokenization} where a codebook $\mathbf{e}_k$, $k \in 1, \dots, K$ is used to quantize the embeddings $E(x)$ into visual tokens (indices), acting like a lookup table.
In addition, a latent prior of the visual tokens is predicted (usually using a transformer), and the decoder is modified to map the visual tokens into the image space.

\begin{table}[t]
    \centering
    \fontsize{7pt}{7pt}
    \selectfont
    \vspace{-0.4cm}
    \caption{List of common datasets used in GM-DC works. 
    Number of samples (\# Samples) refers to the sample size of the entire dataset. In GM-DC experiments, usually, only a subset of the dataset is used.
    We remark that \xmark/\cmark denotes the absence/presence of the dataset under the data constraint settings: {\bf LD}: \underline{L}imited-\underline{D}ata, {\bf FS}: \underline{F}ew-\underline{S}hot and {\bf ZS}: \underline{Z}ero-\underline{S}hot, and Labels indicate if training labels are available (but not necessarily used).
    }
    \label{tab:datasets}
    \begin{tabular}{c@{\hspace{2pt}}c@{\hspace{2pt}} c@{\hspace{3pt}}c @{\hspace{3pt}}c@{\hspace{3pt}}c@{\hspace{3pt}}c@{\hspace{3pt}}c}
    \toprule
        {\bf Dataset} &  {\bf Description} & 
        {\bf \# Samples} &
        {\bf Resolution} &
        {\bf LD}  & {\bf FS} & {\bf ZS}
        & {\bf Labels} \\
       \midrule
        \arrayrulecolor{black!10}
        \makecell[c]{Flickr-Faces-HQ\\ (FFHQ) \\ \citep{karras2019style}} & 
        \parbox[l]{0.45\linewidth}
        {Images with human faces, containing variation in terms of age, ethnicity, and image background.}
        & 70K
        & 1024$\times$1024
        &
        \cmark & \cmark & \cmark &        
        \xmark
        \\ \midrule
        \makecell[c]{Large-scale Scene\\ Understanding (LSUN) \\ \citep{yu2015lsun}  }       &
        \parbox[l]{0.45\linewidth}
          {
        Images with large-scale scene containing 10 scene and 20 object categories.
        }
        &
        3M
        &
        256$\times$256 &
        \cmark & \cmark & \xmark
        &
        \cmark
        \\ \midrule
        MetFace \\ \citep{karras2020ada}            & 
        \parbox[l]{0.45\linewidth}
        {Images depicting paintings, drawings, and statues of human faces 
        } 
        &
        1336 & 1024$\times$1024 &
        \cmark & \cmark & \xmark
        &
        \xmark
        \\ \midrule
        BreCaHAD \\ \citep{aksac2019brecahad}           &
        \parbox[l]{0.45\linewidth}
        {Images of breast cancer histopathology.
        }
        &
        162 & 1360$\times$1024 &
        \cmark & \xmark & \xmark
        &
        \xmark
        \\ \midrule
        \makecell[c]{Animal FacesHQ\\(AFHQ) \\ \citep{choi2020starganv2}} &
        \parbox[l]{0.45\linewidth}
        {Images of animal faces in the domains of cat, dog, and wildlife.}
        &
        15K & 512$\times$512 &
        \cmark & \cmark & \xmark &
        \xmark
        \\ \midrule
        CIFAR-10 \\ \citep{krizhevsky2009cifar100}           &
        \parbox[l]{0.45\linewidth}
        {
        Images including objects and animals.
        }
        &
        60K & 32$\times$32 &
        \cmark & \xmark & \xmark &
        \cmark
        \\ \midrule
        CIFAR-100 \\ \citep{krizhevsky2009cifar100}            & 
        \parbox[l]{0.45\linewidth}{
        A dataset similar to CIFAR-10, but with 100 classes}
        &
        60K & 32$\times$32 &
        \cmark & \xmark & \xmark 
        &
        \cmark
        \\ \midrule
        \makecell[c]{100-shot Obama/\\Gumpy Cat/Panda \\ \citep{zhao2020diffaug}}      &
        \parbox[l]{0.45\linewidth}{Colored images of Obama/Gumpy Cat/Panda}
        &
        100 & 256$\times$256 &
        \cmark & \xmark & \xmark 
        &
        \xmark
        \\\midrule
        Sketches \\ \citep{wang2008faceSketches}            & 
        \parbox[l]{0.45\linewidth}
        {Face sketches in frontal pose, normal lighting, and neutral expressions}
        &
        606 & 256$\times$256 &
        \xmark & \cmark & \xmark 
        &
        \xmark
        \\\midrule
        Sunglasses \\ \citep{ojha2021cdc}         &
        \parbox[l]{0.45\linewidth}
        {Images of human faces wearing sunglasses.
        }
        & 2700 & 256$\times$256 &
        \xmark & \cmark & \xmark 
        &
        \xmark
        \\\midrule
        Babies \\ \citep{ojha2021cdc}             &
        \parbox[l]{0.45\linewidth}
        {Images of baby faces.}
        &
        2500 & 256$\times$256 &
        \xmark & \cmark & \xmark 
        &
        \xmark
        \\\midrule
        Artistic-Faces\\ \citep{yaniv2019face} &
        \parbox[l]{0.45\linewidth}
        {Images containing 160 artistic portraits of 16 different artists.}
        &
        160 & 256$\times$256  &
        \xmark & \cmark & \xmark 
        &
        \xmark
        \\\midrule
        Haunted houses \\ \citep{yu2015lsun}       &
        \parbox[l]{0.45\linewidth}
        {Images of haunted houses}
        &
        1K & 256$\times$256 &
        \xmark & \cmark & \xmark 
        &
        \xmark
        \\\midrule
        Wrecked cars \\ \citep{yu2015lsun}       &
        \parbox[l]{0.45\linewidth}
        {Images of wrecked cars}
        &
        1K & 256$\times$256 &
        \xmark & \cmark & \xmark 
        &
        \xmark
        \\
        \arrayrulecolor{black!100}
        \bottomrule
    \end{tabular}
    \vspace{-10pt}
\end{table}

\begin{table}[!htbp]
    \centering
\fontsize{7pt}{7pt}
\selectfont
\caption{{\bf Our proposed taxonomy for tasks in GM-DC}. For each task, we extract their key characteristics.  
[Attributes] {\bf C}: \underline{C}onditional generation, {\bf P}: \underline{P}re-trained generator given, {\bf I}: \underline{I}mages (as input), {\bf TP}: \underline{T}ext-\underline{P}rompt (as input), {\bf X}: \underline{X}(Cross)-domain adaptation; [Data Constraint] {\bf LD}: \underline{L}imited-\underline{D}ata, {\bf FS}: \underline{F}ew-\underline{S}hot, {\bf ZS}: \underline{Z}ero-\underline{S}hot.
\xmark/\cmark denotes the absence/presence, respectively.
Best viewed in color.
}
\label{tab:tasktaxonomy}
\resizebox{\linewidth}{!}{
\begin{tabular}{
c
c@{\hspace{15pt}}c@{\hspace{15pt}}c@{\hspace{15pt}}c@{\hspace{15pt}}c 
@{\hspace{25pt}}
c@{\hspace{15pt}}c@{\hspace{15pt}}c
c}
\toprule 
\makecell[c]{\bf Task} & \multicolumn{5}{c}{\bf Attributes} & \multicolumn{3}{c}{\bf Data Constraint} & \makecell[c]{\bf Task Illustration}\\
\cmidrule(lr){1-1}\cmidrule(lr){2-6}\cmidrule(lr){7-9}\cmidrule(lr){10-10}
     &{\bf C}
     &{\bf P}
     &{\bf I}
     &{\bf TP}
     &{\bf X}
    &{\bf LD}
    &{\bf FS}
    &{\bf ZS}
 & \\
    \midrule
        \multirow{3}{*}{uGM-1} 
        &
        \xmark & \xmark & \cmark & \xmark & \xmark &
        \cmark &\xmark & \xmark
        &
        \multirow{3}{*}{
        \includegraphics[width=0.48\textwidth, clip, 
        trim={88pt 865pt 1099pt 60pt},]
        {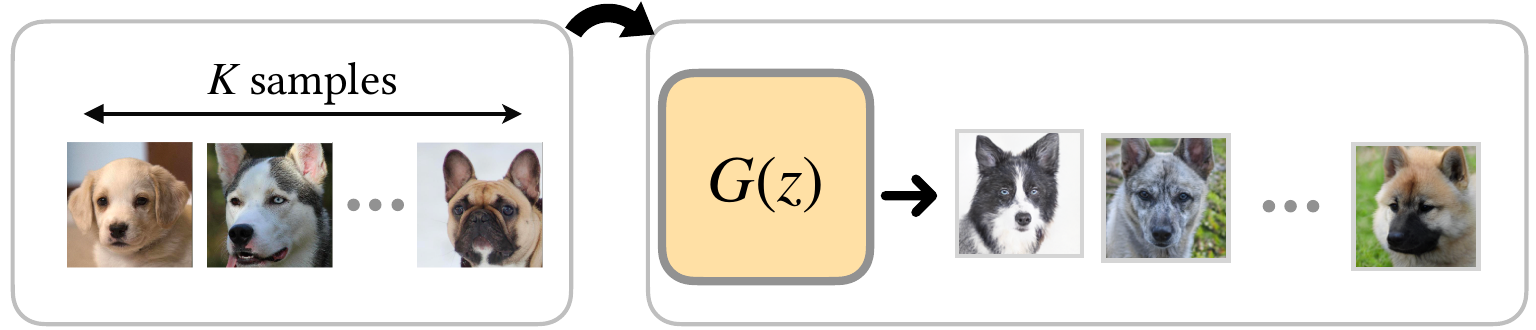}
        }
    \\
    \cmidrule(lr){2-9}
    \vspace{3pt}
    & 
    \multicolumn{8}{p{6cm}}{
    \textbf{Description:} Given $K$ samples from a domain $\D$,
    learn to generate diverse and  high-quality samples
    from $\D$} 
    \\
    &
    \multicolumn{8}{p{6cm}}{
    \textbf{Example:} ADA \citep{karras2020ada} learns a StyleGAN2 using 1k
    images from AFHQ-Dog}  
    \vspace{3pt}
    \\
    \midrule
        \multirow{3}{*}{uGM-2}
        &
        \xmark & \cmark & \cmark & \xmark &\cmark &
        \cmark & \cmark & \xmark &
        \multirow{3}{*}{
        \includegraphics[width=0.48\textwidth, clip, 
        trim={88pt 535pt 1099pt 338pt},]
        {figures/Tasks_v8_1.pdf}
        }
    \vspace{1pt}
    \\
    \cmidrule(lr){2-9}
    \vspace{1pt}
    & 
    \multicolumn{8}{p{6cm}}{
    \textbf{Description:} Given a pre-trained generator on a 
    source domain $\Ds$ and $K$ samples from a target
    domain $\Dt$, learn to generate diverse and
    high-quality samples from $\Dt$
    }
    \\
    &
    \multicolumn{8}{p{6cm}}{
    \textbf{Example:}
    CDC \citep{ojha2021cdc} adapts a pre-trained 
    GAN on FFHQ to Sketches using 10 samples}
    \vspace{2pt}
    \\
    \midrule
        \multirow{3}{*}{uGM-3} 
        &
        \xmark & \cmark & \xmark & \cmark & \cmark &
        \xmark & \xmark & \cmark &
        \multirow{3}{*}{
        \includegraphics[width=0.48\textwidth, clip, 
        trim={88pt 113pt 1099pt 698pt},]
        {figures/Tasks_v8_1.pdf}
        }
    \\
    \cmidrule(lr){2-9}
    \vspace{1pt}
    & 
    \multicolumn{8}{p{6cm}}{
    \textbf{Description:}  Given a pre-trained 
    generator on a 
    source domain $\Ds$
    and a text prompt 
    describing a target 
    domain $\Dt$, learn 
    to generate diverse 
    and high-quality 
    samples from $\Dt$
    }
    \vspace{2pt}
    \\
    &
    \multicolumn{8}{p{6cm}}{
    \textbf{Example:}
    NADA \citep{gal2022stylegannada}
    adapts pre-trained GAN on 
    FFHQ to the 
    painting domain 
    using {\em `Fernando 
    Botero
    Painting'} as input}
    \\
    \midrule
        \multirow{3}{*}{cGM-1} 
        &
        \cmark & \xmark & \cmark & \xmark & \xmark
        & \cmark & \xmark & \xmark &
        \multirow{3}{*}{
        \includegraphics[width=0.48\textwidth, clip, 
        trim={88pt 834pt 1099pt 68pt},]
        {figures/Tasks_v8_2}
        }
    \\
    \cmidrule(lr){2-9}
    \vspace{1pt}
    & 
    \multicolumn{8}{p{6cm}}{
    \textbf{Description:} 
    Given $K$ samples 
    with class labels  
    from a domain $\D$,
    learn to generate 
    diverse and 
    high-quality 
    samples 
    conditioning 
    on the class 
    labels from $\D$
    }
    \\
    &
    \multicolumn{8}{p{6cm}}{
    \textbf{Example:} CbC \citep{shahbazi2022collapse} trains conditional generator on 20 classes of ImageNet Carnivores using 100 images per class
    }
    \vspace{2pt}
    \\
    \midrule
        \multirow{3}{*}{cGM-2} 
        &
        \cmark & \cmark & \cmark & \xmark & \xmark
      & \xmark & \cmark & \xmark &
        \multirow{3}{*}{
        \includegraphics[width=0.48\textwidth, clip, 
        trim={88pt 492pt 1099pt 342pt},]
        {figures/Tasks_v8_2}
        }
    \\
    \cmidrule(lr){2-9}
    \vspace{3pt}
    & 
    \multicolumn{8}{p{6cm}}{
    \textbf{Description:} 
    Given a pre-trained 
    generator on the
    seen classes $\Cs$
    of a domain $\D$ and
    $K$ samples with class
    labels from unseen
    classes $\Cu$ of
    $\D$, learn to generate
    diverse and
    high-quality samples
    conditioning on
    the class labels
    for $\Cu$ from $\D$
    }
    \\
    &
    \multicolumn{8}{p{6cm}}{
    \textbf{Example:} LoFGAN \citep{gu2021lofgan} 
    learns from 85 classes of Flowers to generate 
    images for an unseen class with only 3 samples
    }
    \vspace{3pt}
    \\
    \midrule
        \multirow{3}{*}{cGM-3} 
        &
         \cmark & \cmark & \cmark & \xmark & \cmark &
    \cmark & \cmark & \xmark &
        \multirow{3}{*}{
        \includegraphics[width=0.48\textwidth, clip, 
        trim={88pt 113pt 1099pt 698pt},]
        {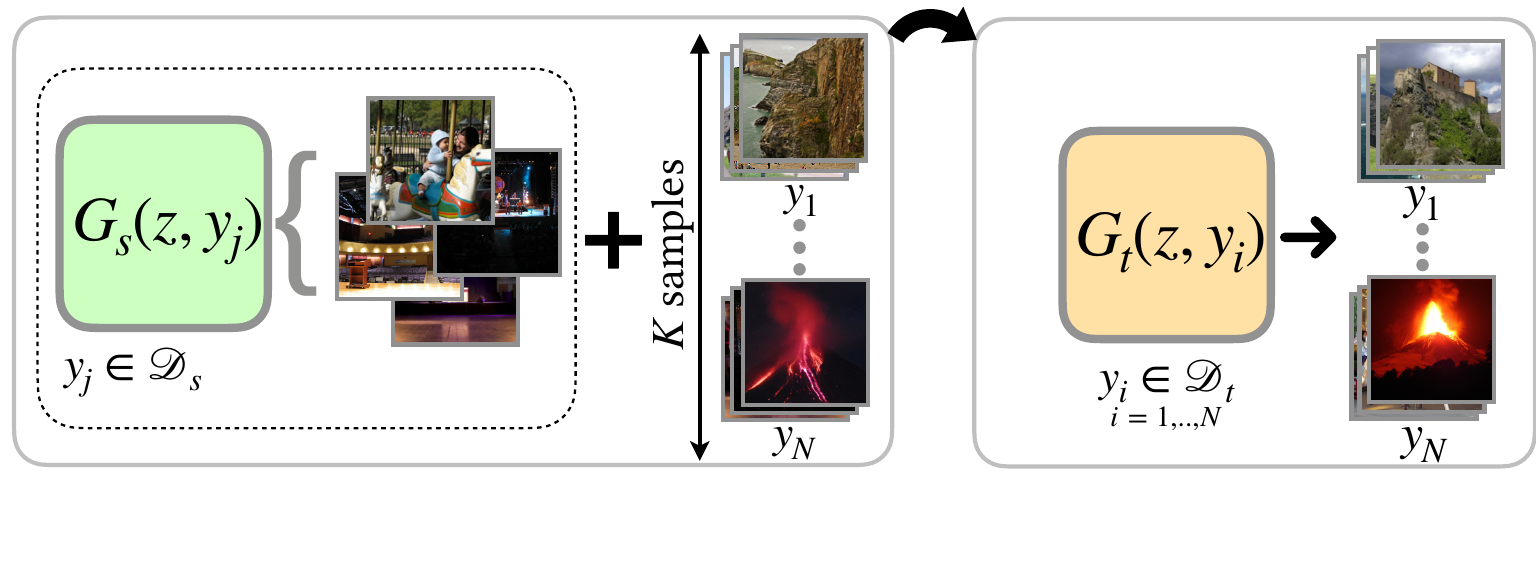}
        }
    \vspace{2pt}
    \\
    \cmidrule(lr){2-9}
    \vspace{3pt}
    & 
    \multicolumn{8}{p{6cm}}{
    \textbf{Description:} 
    Given a pre-trained 
    generator on a 
    source domain $\Ds$
    and $K$ samples 
    with class labels 
    from a target
    domain $\Dt$ , learn 
    to generate diverse 
    and high-quality 
    samples conditioning
    on the class 
    labels from $\Dt$
    }
    \\
    &
    \multicolumn{8}{p{6cm}}{
    \textbf{Example:}
     VPT \citep{sohn2023vpt} adapts
    a pre-trained 
    conditional 
    generator on
    ImageNet 
    to Places365 
    with 500 images per class
    }
    \vspace{3pt}
    \\
    \midrule
        \multirow{3}{*}{IGM} 
        &
         \xmark & \xmark & \cmark & \xmark & \xmark &
    \xmark & \cmark & \xmark &
        \multirow{3}{*}{
        \raisebox{-1.0cm}{
        \includegraphics[width=0.48\textwidth, clip, 
        trim={88pt 849pt 1099pt 20pt},]
        {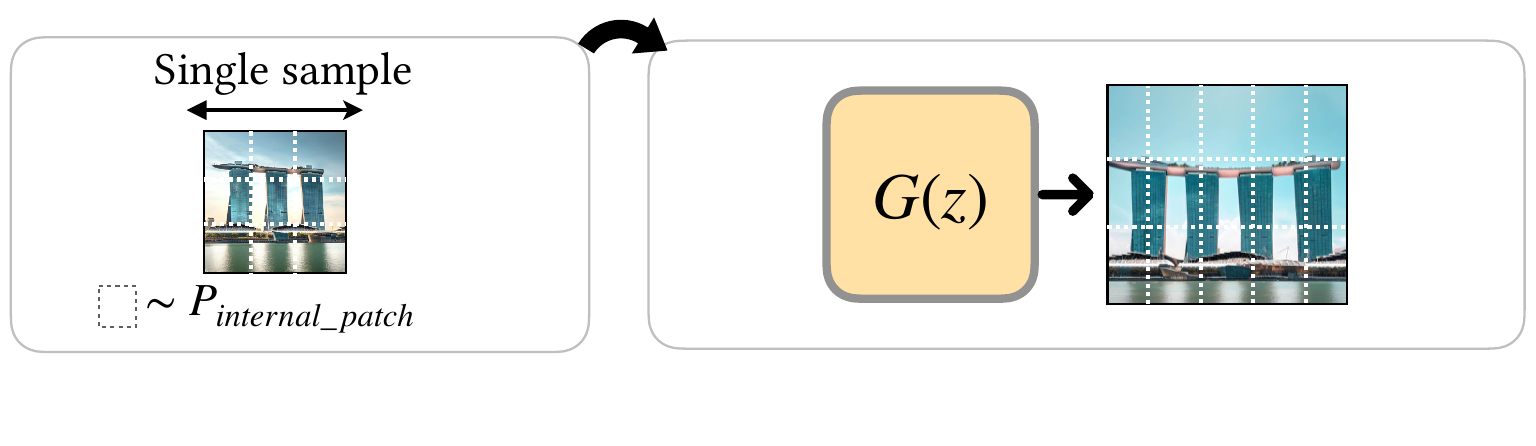}
        }
        }
    \\
    \cmidrule(lr){2-9}
    & 
    \multicolumn{8}{p{6cm}}{
    \textbf{Description:} 
     Given $K$ samples 
    (usually $K=1$) 
    and assuming rich internal
    distribution for
    patches within
    these samples,
    learn to generate
    diverse and
    high-quality
    samples with
    the same internal
    patch distribution
    }
    \\
    &
    \multicolumn{8}{p{6cm}}{
    \textbf{Example:}
    SinDDM \citep{kulikov2023sinddm}
    trains a 
    generator using 
    a single image 
    of
    Marina Bay Sands, and
    generates
    variants of it
    }
    \\
    \midrule
        \multirow{3}{*}{SGM} 
        &
        \xmark & \cmark & \cmark & \cmark & \xmark
        & \xmark & \cmark & \xmark &
        \multirow{3}{*}{
        \includegraphics[width=0.48\textwidth, clip, 
        trim={90pt 440pt 1099pt 365pt},]
        {figures/Tasks_v8_3.pdf}
        }
        \vspace{3pt}
    \\
    \cmidrule(lr){2-9}
    & 
    \multicolumn{8}{p{6cm}}{
    \textbf{Description:} 
    Given a pre-trained generator, $K$ samples
    of a particular
    subject, and a 
    text prompt,
    learn to generate
    diverse and
    high-quality
    samples containing
    the same subject
    }
    \\
    &
    \multicolumn{8}{p{6cm}}{
    \textbf{Example:}
    DreamBooth \citep{ruiz2023dreambooth}
    trains a  generator using 4 images of
    a particular
    backpack and
    adapts it with 
    a text-prompt
    to be in the {\em `grand canyon'}
    }
    \\
    \midrule
\end{tabular}
}
\end{table}

\noindent
{\bf Generative Adversarial Models (GAN)} \citep{goodfellow2014GANs}.
GAN applies an adversarial approach to learn the distribution of data $P_{data}$.
It consists of a generator $G$ and a discriminator $D$ playing a min-max game. 
Specifically, given the latent code $z$, the $G$ learns to generate the images $G(z)$, $z\sim P_z$, where $P_z$ is usually a Gaussian distribution.
Then, $D$ learns to distinguish the real images $x \sim P_{data}$ from the generated ones $G(z) \sim P_{model}$.
The $D$ and $G$ are optimized by respectively maximizing and minimizing the following value function:
\begin{equation}
    \mathcal{V}(D,G)= \mathbb{E}_{x\sim p_{data}} [ \log D(x)] + \mathbb{E}_{z\sim p_z} [\log (1-D(G(z)))] 
\end{equation}

\noindent
{\bf Flow-based Models} \citep{ho2019flow++}.
The flow-based model includes  a series of invertible yet differentiable functions $f$, between latent distribution $P_z$, and data distribution $P_{data}$.
The following log-likelihood function is maximized to train $f(.|\theta)$:
\begin{equation}
    \max_\theta \textstyle \sum_{i=1}^K \log P_{z} (f(x^{(i)}|\theta))+ \log |\det Df(x^{(i)}|\theta)|
\end{equation}
For ease of discussion, we simplify the model as a single flow and denote the training samples with $\{x^{(i)}\}^K_{i=1}$, and the Jacobian of $f(x)$ as $Df(x)$.
We remark that, unlike VAEs that estimate the lower bounds of the log-likelihood, flow-based models evaluate the exact log-likelihood in their loss function.

\noindent
{\bf Diffusion Models (DM)} \citep{ho2020denoising}. 
DM leverages the concept of the diffusion process from stochastic calculus and consists of forward diffusion and reverse diffusion processes.
In the forward diffusion process, based on the foundations of Markov chains, the noise $\epsilon \sim \mathcal{N}(0,I)$ is iteratively added to data samples until it approaches an isotropic Gaussian distribution.
Then, in the backward process, the DM learns to denoise the noisy vector $x_T$ and reconstruct the data samples $x_0$.
This is done by learning the noise estimation model $\epsilon_\theta$ with minimizing the following loss function \citep{ho2020denoising}:
\begin{equation}
    \mathcal{L}=\mathbb{E}_{t,x_0,\epsilon} [||\epsilon - \epsilon_{\theta}(\sqrt{\bar{\alpha}_t}x_0 + \sqrt{1-\bar{\alpha}_t}\epsilon, t)||_2]
    \label{eqn:DMLoss}
\end{equation}
Then, during the generation process, DM first samples a noise $x_T\sim \mathcal{N}(0,I)$, 
and utilizes the learned noise function $\epsilon_\theta$ to iteratively apply the following denoising process \citep{ho2020denoising}:
\begin{equation}
    x_{t-1}=\frac{1}{\sqrt{\alpha_t}}(x_t-\frac{1-\alpha_t}{\sqrt{1-\bar{\alpha}}}\epsilon_\theta(x_t,t)) +\sqrt{\beta_t}\epsilon, \quad t\in[0,T]
    \label{eqn:DMRec}
\end{equation}
Here, $x_t$ is the generated sample at step $T-t$, $\beta_t$ is variance scheduler, $\alpha_t=1-\beta_t$ and
$\bar{\alpha_t}=\prod^t_{s=1}\alpha_s$.


\noindent
{\bf Remark.} We remark that among discussed models, only GANs, DMs, and VAEs are adopted in the context of GM-DC.

\subsection{Data Constraints and Commonly Used Datasets}


In GM-DC, three data constraints have been considered in most works:
(i) \emph{Limited data (LD)}, 
when 50 to 5,000 training samples are given;
(ii) \emph{Few-Shot (FS)},
when 1 to 50 training samples are given;
(iii) \emph{Zero-Shot (ZS)},
when no training samples are given.
These ranges reflect the experimental setups repeatedly adopted in the literature. For example, representative LD works such as StyleGAN2-ADA and FastGAN report results on datasets with roughly $10^2$–$10^3$ samples (e.g., 60 to 1.9k images, up to 5k images), and later studies frequently follow similar scales. Works in the FS setting (e.g., EWC, CDC, AdAM, JoJoGAN) explicitly assume only 1–50 samples from the target domain. More recent ZS methods (e.g., NADA, IPL, AIR) assume no target-domain samples and instead rely on external guidance. We adopt these ranges to align with these widely used regimes. Training under such constraints often leads to issues such as overfitting and mode collapse. Tab.~\ref{tab:datasets} lists the most common datasets used in GM-DC with related details.



\newenvironment{myquote}
  {\begin{list}{}{%
     \setlength{\leftmargin}{1.6em}
     \setlength{\rightmargin}{0pt}
     }
   \item\relax}
  {\end{list}}

\section{Generative Modeling under Data Constraint: Task Taxonomy, Challenges}
\label{sec:taxonomy}
In this section, first, we present our proposed taxonomy on different GM-DC tasks (Sec.~\ref{ssec:tasks}) highlighting their relationships and differences based on their attributes, e.g. unconditional or conditional  generation. 
Then, we present the unique challenges of GM-DC (Sec.~\ref{ssec:challenges}), including new insights such as domain proximity, and incompatible knowledge transfer.
Later, in Sec.~\ref{sec:comprehensive_review}, we present our proposed taxonomy on approaches for GM-DC, with a detailed review of individual work organized under our proposed taxonomy.

\subsection{Generative Modeling under Data Constraint: A Taxonomy on Tasks}
\label{ssec:tasks}

The goal of GM-DC is to learn to generate diverse and high-quality samples given only a small number of training samples. A number of GM-DC setups have been studied in different works
(Fig.~\ref{fig:timeline}). In this section, we propose a {\bf GM-DC task taxonomy} to categorize  setups in different works.  Tab. \ref{tab:tasktaxonomy} tabulates our GM-DC task taxonomy.

\begin{enumerate}
\item {\bf Unconditional generative modeling under data constraint (uGM-1).}
\vspace{-0.6em}
\begin{definition}[uGM-1]
{\em Given $K$ samples from domain $\D$, learn to generate diverse and high-quality samples from $\D$.}
\end{definition}
\vspace{-0.7em}
Without leveraging other side information, existing work has studied uGM-1 under limited samples ranging from 100 to several thousands. uGM-1 is an important task, especially for a domain that is distant from common domains, e.g. medical images which are distant from common personal photos in terms of content and characteristics.
In such scenarios, leveraging from common domains would not  provide any advantage. We provide a comprehensive quantitative comparison of representative methods in Tab. \ref{tab:ugm1}.

\item {\bf Unconditional generative modeling under data constraint with pre-trained generator  and cross-domain adaptation} (uGM-2).
\vspace{-0.6em}
\begin{definition}[uGM-2]
{\em Given a pre-trained generator on a source domain $\Ds$  (with numerous and diverse samples) and $K$ samples from a target domain $\Dt$, learn to generate diverse and high-quality samples from $\Dt$.}
\end{definition}
\vspace{-0.7em}
uGM-2 is similar to uGM-1, except that 
a pre-trained generator on another source domain $\Ds$ is additionally given.
uGM-2 is a major task in GM-DC and has been studied in many works.
In most works, close proximity in semantic 
between $\Ds$ and $\Dt$ is assumed, \eg $\Ds$ is photos of human faces, $\Dt$ is sketches of human faces.
For uGM-2, 
transfer learning  has been a popular approach to tackle this task driving GM-DC into the few-shot regime, \eg only 10 samples from $\Dt$ are given \citep{li2020ewc} (See Sec.~\ref{sec:comprehensive_review}
for the taxonomy of GM-DC approaches).
Recent work has started to look into the challenging setup when $\Ds$ and $\Dt$ are more semantically apart \citep{zhao2022adam}, \eg $\Ds$ is photos of human faces, $\Dt$ is photos of cat faces. We provide a comprehensive quantitative comparison of representative methods in Tab. \ref{tab:ugm2}. See Sec.~\ref{ssec:challenges} for further discussion on domain proximity in GM-DC.

\item {\bf Unconditional generative modeling under data constraint with pre-trained generator  and cross-domain adaptation, using text prompt (uGM-3).}
\vspace{-0.5em}
\begin{definition}[uGM-3]
{\em Given a pre-trained generator on a source domain $\Ds$  (with numerous and diverse samples) and a text prompt describing  a target domain $\Dt$, learn to generate diverse and high-quality samples from $\Dt$.}
\end{definition}
\vspace{-0.7em}
uGM-3 is similar to uGM-2, except that a text prompt is provided to describe $\Dt$ instead of samples from $\Dt$.
Particularly, this task requires generating samples  from $\Dt$ without seeing any sample from that domain, \ie 
 zero-shot domain adaptation.
Important work to tackle this task leverages recent large vision-language models to provide textual direction to guide the adaptation of the pre-trained generator to $\Dt$ \citep{gal2022stylegannada}.

\item {\bf Conditional generative modeling under data constraint (cGM-1).}
\vspace{-0.5em}
\begin{definition}[cGM-1]
{\em Given $K$ samples with class labels from a domain $\D$, learn to generate diverse and high-quality samples  conditioning on the class labels  from $\D$.}
\end{definition}
\vspace{-0.7em}
cGM-1 is similar to uGM-1 but focuses on conditional generation, \ie inputs to the generator include a random latent vector and a class label.
Conditional generative models such as BigGAN \citep{brock2019biggan} could achieve high-quality image generation 
when they are trained on large-scale datasets \eg ImageNet. However, under limited data, 
it is 
challenging to 
 achieve diverse and high-quality conditional sample generation.
As a natural extension of uGM-1, 
data augmentation has been studied for 
cGM-1 among other approaches, see Sec.~\ref{sec:comprehensive_review}.

\item {\bf Conditional generative modeling under data constraint with pre-trained generator (cGM-2).}
\vspace{-1.5em}
\begin{definition}[cGM-2]
{\em Given a pre-trained generator on the seen classes $\Cs$ of a domain $\D$, and
$K$ samples with class labels from unseen classes $\Cu$ of $\D$, learn to generate diverse and high-quality samples conditioning on the class labels for $\Cu$ from $\D$.}
\end{definition}
\vspace{-0.7em}
cGM-2 is similar to cGM-1, except that a pre-trained generator on the seen classes $\Cs$ is additionally given.
Note that in cGM-2, $\Cs$ and $\Cu$  contain disjoint classes, but both of them are from the same domain $\D$.
For example, \citet{shahbazi2021efficient} 
studies the setup when CIFAR100
\citep{krizhevsky2009cifar100}
is partitioned into 80 seen classes for the pre-trained generator and 20 unseen classes as the target, 
with 100 samples per unseen class given for training.
Meta-learning and transfer learning (regularizer-based fine-tuning, etc.) have been effective approaches for cGM-2, see Sec.~\ref{sec:comprehensive_review}.

\item {\bf Conditional Generative Modeling under data constraint with pre-trained generator and cross-domain adaptation (cGM-3).} 
\vspace{-0.5em}
\begin{definition}[cGM-3]
{\em Given a pre-trained generator on a source domain $\Ds$ (with numerous and diverse samples) and $K$ samples with class labels from a target domain $\Dt$, learn to generate diverse and high-quality samples conditioning on the class labels from $\Dt$.}
\end{definition}
cGM-3 is similar to uGM-2 as cross-domain adaptation is required in both tasks, but cGM-3 focuses on conditional generation while uGM-2 focuses on unconditional generation.
Furthermore, cGM-3 is similar to cGM-2, but seen classes and unseen classes are from different domains in cGM-3.
For example, \citet{shahbazi2021efficient} has studied the setup when a pre-trained generator on ImageNet is adapted to generate samples for several classes from Places365 \citep{zhou2017places}.
Transfer learning is one of the effective approaches for cGM-3, see Sec.~\ref{sec:comprehensive_review}.

\item {\bf Internal patch distribution Generative Modeling  (IGM).}
\vspace{-0.5em}
\begin{definition}[IGM]
{\em Given $K$ samples and assuming rich internal distribution for patches within these samples, learn to generate diverse and high-quality samples with the same internal patch distribution.}
\end{definition}
\vspace{-0.7em}
IGM aims to  capture the internal distribution of
patches within the samples. 
With the model capturing the samples' patch statistics, it is then possible  to generate high
quality, diverse samples 
with the same content as the given training samples.
In most works, $K = 1$, and IGM focuses on images \citep{shaham2019singan},
learning to generate new images with 
significant variability while maintaining
both the global structure and fine textures of the training image.

\item {\bf Subject-driven Generative Modeling  (SGM).}
\vspace{-0.5em}
\begin{definition}[SGM]
{\em Given $K$ samples of a particular subject and a text prompt, learn to generate diverse and high-quality samples containing the same subject.}
\end{definition}
\vspace{-0.7em}
SGM is a 
recent GM-DC task introduced in \citet{ruiz2023dreambooth}.
Given  a few images (3-5 in most cases) of a subject
and leveraging a large text-to-image generative model, \citet{ruiz2023dreambooth} learns to generate diverse images of the subject in different contexts with the guidance of text prompts. 
The goals are: i) to achieve  natural interactions between the subject and diverse new contexts, and ii)  to maintain high fidelity to the key visual features of the subject. 
In \citet{ruiz2023dreambooth}, a natural language-guided transfer learning approach and a new prior preservation loss have been proposed to achieve SGM. 

\end{enumerate}
\clearpage %
\begin{spacing}{0.88}
\fontsize{7pt}{7pt}
\selectfont
\setlength\tabcolsep{2pt}
\begin{longtable}{cc}
\caption{
{\bf Our proposed taxonomy for approaches in GM-DC.}
For each  approach, 
the addressed GM-DC tasks (see Tab. \ref{tab:tasktaxonomy} for task definitions) and the data constraints 
are indicated. 
A detailed list of 
methods under 
each sub-category is also tabulated (some 
methods are under  multiple categories).
\xmark/\cmark ~denotes the absence/presence of the tasks commonly addressed by each approach, and  
the data constraints 
usually considered: {\bf LD}: \underline{L}imited-\underline{D}ata, {\bf FS}: \underline{F}ew-\underline{S}hot and {\bf ZS}: \underline{Z}ero-\underline{S}hot. 
}
\label{tab:approaches}
\\
\toprule
\rowcolor{gray!10}\multicolumn{2}{c}{{\bf \normalsize Transfer Learning} (Sec.~\ref{ssec:review_transferlearning})}
\\[2pt]
{\bf \small Description:}& \parbox[l]{0.90\linewidth}
{
Improve 
GM-DC on target domain by knowledge of a generator pre-trained on source domain (with numerous and diverse samples).} 
\\[2pt]
{\bf \small Task:} & \parbox[l]{0.92\linewidth}
{
uGM-1 \xmark \hspace{2pt} 
uGM-2 \cmark \hspace{2pt} 
uGM-3 \cmark \hspace{2pt} 
cGM-1 \xmark \hspace{2pt} 
cGM-2 \cmark \hspace{2pt} 
cGM-3 \cmark \hspace{2pt} 
IGM \xmark \hspace{2pt} 
SGM \cmark 
\hspace{5pt} 
{\bf \small Data constraint:} \hspace{-3pt} 
LD \cmark \hspace{1pt}
FS \cmark \hspace{1pt}
ZS \cmark 
}
\\
\arrayrulecolor{black!50}
\midrule
\multicolumn{2}{l}{
    \cellcolor{gray!0}\parbox[l]{\linewidth}{
    \textbf{1) Regularizer-based Fine-Tuning:} 
    Explore regularizers to preserve source generators' knowledge.}
    }
 \\[2pt]
\multicolumn{2}{l}{
    \parbox[l]{\linewidth}{
    \emph{Methods:} TGAN \citep{wang2018tgan}, BSA \citep{noguchi2019bsa}, FreezeD \citep{mo2020freezed}, EWC \citep{li2020ewc}, CDC \citep{ojha2021cdc}, cGANTransfer \citep{shahbazi2021efficient}, W\textsuperscript{3} \citep{grigoryev2022when}, C \textsuperscript{3}\citep{Lee2021C3}, DCL \citep{zhao2022dcl}, RSSA \citep{xiao2022rssa}, fairTL \citep{teo2023fairtl}, GenOS\citep{zhang2022generalizedoneshot}
           , SVD \citep{robb2020svd}, 
           D\textsuperscript{3}-TGAN \citep{wu2023d3tgan}, JoJoGAN \citep{chong2022jojogan}, 
           KDFSIG \citep{hou2022exploitingkd}, CtlGAN \citep{wang2022ctlgan}, 
           ICGAN \citep{casanova2021icgan}, 
           MaskD \citep{zhu2022few}, 
           F\textsuperscript{3} \citep{yuichi2023_fewshot},
           ICGAN \citep{casanova2021icgan},
           DDPM-PA \citep{zhu2022few_dm},
           DWSC \citep{hou_dynamic}, 
           CSR \citep{gou2023csr},
           ProSC \citep{moon2023prosc},
DOGAN \citep{hu2024DOGAN}, 
AnyDoor \citep{chen2024anydoor}, 
SmoothSim \citep{sushko2023smoothsim}, 
TAN \citep{wang2023efficient}, 
FPTGAN \citep{zhang2023fptgan}, 
CLCR \citep{zhang2023clcr}, 
FastFaceGAN \citep{kato2023faster}, 
FDDC \citep{hu2023phasic}, 
StyleDomain \citep{alanov2023styledomain}, 
DomainExpansion \citep{nitzan2023domain}, 
CVD-GAN \citep{jiang2023personalized}, 
Def-DINO \citep{zhou2024deformable}, 
HDA \citep{li2023hda}, 
FAGAN \citep{cheng2024frequency}, 
SSCR \citep{israr2024few}, 
DPMs-ANT \citep{wang2024bridging}, 
SACP \citep{he2024few},
DATID-3D \citep{kim2023datid3d}

       }
       }
\\
\midrule
\multicolumn{2}{l}{
    \cellcolor{gray!0}\parbox[l]{\linewidth}{
    \textbf{2) Latent Space:} 
   Explore latent space of source generator to identify suitable knowledge for adaptation.}
    }
\\[2pt]
\multicolumn{2}{l}{
    \parbox[l]{\linewidth}{
    \emph{Methods:} 
     MineGAN \citep{wang2020minegan}, MineGAN++ \citep{wang2021minegan++},
     GenDA \citep{yang2021genda},
     TF\textsuperscript{2} \citep{yu2024tf},
     LCL \citep{mondal2023lcl},
     SoLAD \citep{Mondal2024SoLAD},
     WeditGAN \citep{duan2023weditgan},
     CRDI \citep{cao2025CRDI},
     SiSTA \citep{thopalli2023targetaware}, MultiDiffusion \citep{bar2023multidiffusion},
     DiS \citep{everaert2023dis}
       }
       }
 \\
\midrule
\multicolumn{2}{l}{
    \cellcolor{gray!0}\parbox[l]{\linewidth}{
    \textbf{3) Modulation:} 
   Leverage trainable modulation weights on top of frozen weights of the source generator.}
    }
 \\[2pt]
\multicolumn{2}{l}{
    \parbox[l]{\linewidth}{
    \emph{Methods:} 
     AdaFMGAN \citep{zhao2020leveraging}, GAN-Memory \citep{cong2020ganmemory}, CAM-GAN \citep{varshney2021camgan}, AdAM \citep{zhao2022adam}, DynaGAN \citep{kim2022dynagan}, HyperDomainNet \citep{alanov2022hyperdomainnet}, NICE \citep{ni2023nice}, A\textsuperscript{3}FT \citep{moon2022finetuning}, LFS-GAN \citep{seo2023lfsgan}, OKM \citep{zhang2024few}, CFTS-GAN \citep{ali2025cfts}, HyperGAN-CLIP \citep{anees2024hypergan}, DPH \citep{li2024dual}, DoRM \citep{wu2024domain}, Mix-of-Show \citep{gu2023mixofshow}, Orthogonal Adaptation \citep{po2024orthogonal}, DreamMatcher \citep{nam2024dreammatcher}, PortraitBooth \citep{peng2024portraitbooth}, DisenDiff \citep{zhang2024disendiff}, RealCustom \citep{huang2024realcustom}
       }
       }
\\
\midrule
\multicolumn{2}{l}{
    \cellcolor{gray!0}\parbox[l]{\linewidth}{
    \textbf{4) Natural Language-guided:} 
      Use the feedback of vision-language models to adapt the source generator with text prompts.}
    }
 \\[2pt]
\multicolumn{2}{l}{
    \parbox[l]{\linewidth}{
    \emph{Methods:} 
       StyleGAN-NADA \citep{gal2022stylegannada}, MTG \citep{zhu2022mindthegap}, HyperDomainNet \citep{alanov2022hyperdomainnet}, DiFa \citep{zhang2022difa}, OneCLIP \citep{kwon2022oneclip}, IPL \citep{guo2023ipl}, SVL \citep{jeon2023svdm}, AIR \citep{liu2025air}, StyleGAN-Fusion \citep{song2024stylegan}, UniHDA \citep{li2024unihda},
       \textsc{ITI-Gen} \citep{zhang2023itigen},
       FairQueue \citep{teo2024fairqueue}, 
       SINE \citep{zhang2023sine}, DreamBooth \citep{ruiz2023dreambooth},
       Custom Diffusion \citep{kumari2023mcc}, Textual-Inversion \citep{gal2022textualinversion}, SpecialistDiffusion \citep{lu2023specialistdiffusion},
       BLIP-Diffusion \citep{li2023blipdiffusion}, AblateConcept \citep{kumari2023ablating}, StyO \citep{li2024styostylizefaceoneshot}, HyperGAN-CLIP \citep{anees2024hypergan}, DPH \citep{li2024dual}, ELITE \citep{Wei2023elite}, E4T \citep{gal2023e4t}, MoMA \citep{song2025MOMA}, SSR-Encoder \citep{zhang2024ssr}, MultiGen \citep{wu2025MultiGen}, MasterWeaver \citep{wei2025masterweaver}, Lego \citep{motamed2025Lego}, CGR \citep{jin2025CGR}, DreamBlend \citep{ram2025dreamblend}, Cross Initialization \citep{pang2023crossinitialization}, SAG \citep{chan2024sag}, ZipLoRA \citep{shah2025ziplora}, RealCustom \citep{huang2024realcustom}, PALP \citep{arar2024palp}, InstantBooth \citep{shi2023instantbooth}, IDAdapter \citep{cui2024idadapter}, Domain gallery \citep{duan2024domaingallery}, LogoSticker \citep{zhu2025logosticker}, ProSpect \citep{zhang2023prospect}, Dreambooth-CL \citep{zhu2024enhancing}, AnomalyDiffusion \citep{hu2024anomalydiffusion}, ComFusion \citep{hong2025ComFusion}, SuDe \citep{qiao2024facechain}, LFS-Diffusion \citep{song2024towards}, L\textsuperscript{2}DM \citep{sun2024create}, Omg \citep{kong2025OMG}, TFIC \citep{li2025TFIC}, MagiCapture \citep{hyung2024magicapture}, Mix-of-Show \citep{gu2023mixofshow}, Orthogonal Adaptation \citep{po2024orthogonal}, DBLoRA \citep{pascual2024enhancing}, HybirdBooth \citep{guan2025hybridbooth}, T2IRL \citep{wei2025T2IRL}, HyperDreamBooth \citep{ruiz2024hyperdreambooth}, FastComposer \citep{xiao2024fastcomposer}, PortraitBooth \citep{peng2024portraitbooth}, DreamMatcher \citep{nam2024dreammatcher}, DisenDiff \citep{zhang2024disendiff}, CII \citep{jeong2023cii}, DETEX \citep{cai2024decoupled}
       }
       }
\\
\midrule
\multicolumn{2}{l}{
    \cellcolor{gray!0}\parbox[l]{\linewidth}{
    \textbf{5) Adaptation-Aware:} 
      Preserve the source generator's knowledge that is important to the adaptation task.}
    }
 \\[2pt]
\multicolumn{2}{l}{
    \parbox[l]{\linewidth}{
    \emph{Methods:} 
        AdAM \citep{zhao2022adam}, RICK \citep{zhao2023rick}, OKM \citep{zhang2024few}
       }
       }
\\
\midrule
\multicolumn{2}{l}{
    \cellcolor{gray!0}\parbox[l]{\linewidth}{
    \textbf{6) Prompt Tuning:} 
      Freeze the source generator and add/ generate visual prompts to guide generation for the target domain.}
    }
 \\[2pt]
\multicolumn{2}{l}{
    \parbox[l]{\linewidth}{
    \emph{Methods:} 
       VPT \citep{sohn2023vpt}
       }
       }
\\
\arrayrulecolor{black!100} \midrule 
\rowcolor{gray!10}\multicolumn{2}{c}{{\bf \normalsize Data Augmentation}
(Sec.~\ref{ssec:review_dataaugmentation})}
\\[2pt]
{\bf \small Description:}& \parbox[l]{0.90\linewidth}{
Improve GM-DC by increasing coverage of the data distribution by applying various transformations on the given samples.}
\\[2pt]
{\bf \small Task:} & \parbox[l]{0.92\linewidth}
{
uGM-1 \cmark \hspace{2pt} 
uGM-2 \xmark \hspace{2pt} 
uGM-3 \xmark \hspace{2pt} 
cGM-1 \xmark \hspace{2pt} 
cGM-2 \xmark \hspace{2pt} 
cGM-3 \xmark \hspace{2pt} 
IGM \xmark \hspace{2pt} 
SGM \xmark 
\hspace{5pt} 
{\bf \small Data constraint:} \hspace{-3pt} 
LD \cmark \hspace{1pt}
FS \xmark \hspace{1pt}
ZS \xmark 
}
\\
\arrayrulecolor{black!50}
\midrule
\multicolumn{2}{l}{
    \cellcolor{gray!0}\parbox[l]{\linewidth}{
    \textbf{1) Image-Level Augmentation:} 
      Apply data transformations 
      on image space.}
    }
 \\[2pt]
\multicolumn{2}{l}{
    \parbox[l]{\linewidth}{
    \emph{Methods:} 
        ADA \citep{karras2020ada}, DiffAugment \citep{zhao2020diffaug}, IAG \citep{zhao2020imageaugmentation}, DiffusionGAN \citep{wang2023diffusiongan}, bCR \citep{zhao2021improved}, CR-GAN \citep{Zhang2020Consistency}, APA \citep{jiang2021deceive}, PatchDiffusion \citep{wang2023patchdiffusion},
ANDA \citep{zhang2024anda},
DANI \citep{zhang2024dani},
AugSelf-GAN \citep{hou2024augselfgan}

       }
       }
\\
\midrule
\multicolumn{2}{l}{
    \cellcolor{gray!0}\parbox[l]{\linewidth}{
    \textbf{2) Feature-Level Augmentation:} 
     Apply data transformations 
     on the feature space.
    }
    }
 \\[2pt]
\multicolumn{2}{l}{
    \parbox[l]{\linewidth}{
    \emph{Methods:} 
        AdvAug \citep{chen2021advaug}, AFI \citep{dai2021implicit},
FSMR \citep{kim2022feature}

       }
       }
\\
\midrule
\multicolumn{2}{l}{
    \cellcolor{gray!0}\parbox[l]{\linewidth}{
    \textbf{3) Transformation-Driven Design:} 
    Leverage the information of 
    individual transformations
    to design an efficient learning mechanism.
    }
}
 \\[2pt]
\multicolumn{2}{l}{
    \parbox[l]{\linewidth}{
    \emph{Methods:} 
        DAG \citep{tran2021dag}, SSGAN-LA \citep{hou2021labelaugmentation}
       }
       }
\\
\arrayrulecolor{black!100}
\midrule
\rowcolor{gray!10}\multicolumn{2}{c}{{\bf \normalsize Network Architectures} {(Sec.~\ref{ssec:review_networkarchitecture})}}
\\[2pt]
{\bf \small Description:}& \parbox[l]{0.90\linewidth}{Design specific architecture for the generator to improve its learning under data constraints.} 
\\[2pt]
{\bf \small Task:} & \parbox[l]{0.92\linewidth}
{
uGM-1 \cmark \hspace{2pt} 
uGM-2 \xmark \hspace{2pt} 
uGM-3 \xmark \hspace{2pt} 
cGM-1 \cmark \hspace{2pt} 
cGM-2 \xmark \hspace{2pt} 
cGM-3 \xmark \hspace{2pt} 
IGM \xmark \hspace{2pt} 
SGM \xmark 
\hspace{5pt} 
{\bf \small Data constraint:} \hspace{-3pt} 
LD \cmark \hspace{1pt}
FS \xmark \hspace{1pt}
ZS \xmark 
}
\\
\arrayrulecolor{black!50}
\midrule
\multicolumn{2}{l}{
    \cellcolor{gray!0}\parbox[l]{\linewidth}{
    \textbf{1) Feature Enhancement:} 
      Design additional modules/ layers to enhance/ retain the feature maps of the generator for better generative modeling.}
    }
 \\[2pt]
\multicolumn{2}{l}{
    \parbox[l]{\linewidth}{
    \emph{Methods:} 
    FastGAN \citep{liu2021fastgan},
    cF-GAN \citep{hiruta2022cfgan},
    MoCA \citep{li2022moca}, DFSGAN \citep{yang2023dfsgan},
    DM-GAN \citep{yan2024dm},
    FewConv \citep{liu2025fewconv},
    SCHA-VAE \citep{pmlr-v162-giannone22a}
       }
       }
\\
\midrule
\multicolumn{2}{l}{
    \cellcolor{gray!0}\parbox[l]{\linewidth}{
    \textbf{2) Ensemble Large Pre-trained Vision Models:} 
    Improve architecture by integrating pre-trained vision models to enable more accurate GM-DC.}
    }
 \\[2pt]

\multicolumn{2}{l}{
    \parbox[l]{\linewidth}{
    \vspace{0.1cm}
    \emph{Methods:} 
    ProjectedGAN \citep{sauer2021projectedgan},
    SPGAN \citep{hiruta2022cfgan},
    Vision-aided GAN \citep{kumari2022ensembling},
    P2D \citep{chong2024p2d},
    DISP \citep{mangla2022data}
       }
       }
\\
\midrule
\multicolumn{2}{l}{
    \cellcolor{gray!0}\parbox[l]{\linewidth}{
    \textbf{3) Dynamic Network Architecture:} 
     Improve generative learning with limited data by evolving the generator architecture during training.
    }
    }
 \\[2pt]
\multicolumn{2}{l}{
    \parbox[l]{\linewidth}{
    \vspace{0.05cm}
    \emph{Methods:} 
    CbC \citep{shahbazi2022collapse},
    PYP \citep{li2024pyp},
    DynamicD \citep{yang2022dynamicd}, AdvAug \citep{chen2021advaug}, Re-GAN \citep{saxena2023regan},
    RG-GAN \citep{saxena2024rggan},
    AutoInfoGAN \citep{shi2023autoinfogan}
       }
       }
\\
\arrayrulecolor{black!100}
\midrule
\rowcolor{gray!10}\multicolumn{2}{c}{{\bf \normalsize 
Multi-Task Objectives} 
{(Sec.~\ref{ssec:review_trainingtechniques})}}
\\[2pt]
{\bf \small Description:}& \parbox[l]{0.90\linewidth}{
Introduce additional task(s) to extract generalizable representations that are useful for all tasks, to reduce overfitting under data constraints.
} 
\\[2pt]
{\bf \small Task:} & \parbox[l]{0.92\linewidth}
{
uGM-1 \cmark \hspace{2pt} 
uGM-2 \cmark \hspace{2pt} 
uGM-3 \xmark \hspace{2pt} 
cGM-1 \cmark \hspace{2pt} 
cGM-2 \xmark \hspace{2pt} 
cGM-3 \xmark \hspace{2pt} 
IGM \xmark \hspace{2pt} 
SGM \xmark 
\hspace{5pt} 
{\bf \small Data constraint:} \hspace{-3pt} 
LD \cmark \hspace{1pt}
FS \cmark \hspace{1pt}
ZS \xmark 
}
\\
\arrayrulecolor{black!50}
\midrule
\multicolumn{2}{l}{
    \cellcolor{gray!0}\parbox[l]{\linewidth}{
    \textbf{1) Regularizer:} 
    Add an additional task objective as a regularizer to prevent an undesirable behaviour during training generative model.
    }
    }
 \\[2pt]
\multicolumn{2}{l}{
    \parbox[l]{\linewidth}{
    \vspace{0.05cm}
    \emph{Methods:} 
         LeCam \citep{tseng2021lecam}, 
         RegLA \citep{hou2023regularizing},
         DigGAN \citep{fang2022diggan},
         MICGAN \citep{zhai2024micgan},
         CHAIN \citep{ni2024chain},
         MDL \citep{kong2022mdl},
         DFMGAN \citep{duan2023few}
       }
       }
\\
\midrule
\multicolumn{2}{l}{
    \cellcolor{gray!0}\parbox[l]{\linewidth}{
    \textbf{2) Contrastive Learning:} 
    Introduce a pretext task to enhance the learning process of the generative model.
    }
    }
 \\[2pt]
\multicolumn{2}{l}{
    \parbox[l]{\linewidth}{
    \emph{Methods:} 
     InsGen \citep{yang2021insgen}, FakeCLR \citep{li2022fakeclr}, DCL \citep{zhao2022dcl}, C\textsuperscript{3} \citep{Lee2021C3}, ctlGAN \citep{wang2022ctlgan}, IAG \citep{zhao2020imageaugmentation}, CML-GAN  \citep{phaphuangwittayakul2022cmlgan},
     RCL \citep{gou2024few}
       }
       }
       \\
\midrule
\multicolumn{2}{l}{
    \cellcolor{gray!0}\parbox[l]{\linewidth}{
    \textbf{3) Masking:} 
    Mask a part of the image/ information to increase the task hardness and prevent learning the trivial solutions.
    }
    }
 \\[2pt]
\multicolumn{2}{l}{
    \parbox[l]{\linewidth}{
    \emph{Methods:} 
     MaskedGAN \citep{huang2022maskedgan}, MaskD \citep{zhu2022few},
     DMD \citep{zhang2023dmd}
       }
       }
\\
\midrule
\multicolumn{2}{l}{
    \cellcolor{gray!0}\parbox[l]{\linewidth}{
    \textbf{4) Knowledge Distillation:} 
    Add a task objective that enforces the generator to follow a strong teacher.
    }
    }
 \\[2pt]
\multicolumn{2}{l}{
    \parbox[l]{\linewidth}{
    \emph{Methods:} 
    KD-DLGAN \citep{cui2023kddlgan}, KDFSIG \citep{hou2022exploitingkd},
    BK-SDM \citep{kim2025bksdm}
       }
       }
\\
\midrule
\multicolumn{2}{l}{
    \cellcolor{gray!0}\parbox[l]{\linewidth}{
    \textbf{5) Prototype Learning:} 
    Emphasize learning prototypes for samples/ concepts within the distribution as an additional task objective.
    }
    }
 \\[2pt]
\multicolumn{2}{l}{
    \parbox[l]{\linewidth}{
    \emph{Methods:} 
     ProtoGAN \citep{yang2023protogan}, MoCA \citep{li2022moca}
       }
       }
\\
\midrule
\multicolumn{2}{l}{
    \cellcolor{gray!0}\parbox[l]{\linewidth}{
    \textbf{6) Other Multi-Task Objectives:} 
    Apply other
    types of multi-task objectives
    including co-training, patch-level learning, and diffusion.
    }
    }
 \\[2pt]
\multicolumn{2}{l}{    
    \parbox[l]{\linewidth}{
    \emph{Methods:} 
     GenCo \citep{cui2022genco}, PatchDiffusion \citep{wang2023patchdiffusion}, AnyRes-GAN \citep{chai2022anyresolution} , DiffusionGAN \citep{wang2023diffusiongan}, D2C \citep{sinha2021d2c},
     AdaptiveIMLE \citep{aghabozorgi2023adaptiveimle},
     RS-IMLE \citep{vashist2024rejection},
     FSDM \citep{giannone2022fsdm},
     SpiderGAN \citep{asokan2023spider}
       }
       }
\\
\arrayrulecolor{black!100}
\midrule
\rowcolor{gray!10}\multicolumn{2}{c}{{\bf \normalsize Exploiting Frequency Components} {(Sec.~\ref{ssec:review_exploitingfrequency})}}
\\[2pt]
{\bf \small Description:}& \parbox[l]{0.90\linewidth}{
Exploit frequency components to improve learning the generative model by reducing frequency bias.} 
\\[2pt]
{\bf \small Task:} & \parbox[l]{0.92\linewidth}
{
uGM-1 \cmark \hspace{2pt} 
uGM-2 \xmark \hspace{2pt} 
uGM-3 \xmark \hspace{2pt} 
cGM-1 \xmark \hspace{2pt} 
cGM-2 \cmark \hspace{2pt} 
cGM-3 \xmark \hspace{2pt} 
IGM \xmark \hspace{2pt} 
SGM \xmark 
\hspace{5pt} 
{\bf \small Data constraint:} \hspace{-3pt} 
LD \cmark \hspace{1pt}
FS \cmark \hspace{1pt}
ZS \xmark 
}
\\
\arrayrulecolor{black!50}
\midrule
\multicolumn{2}{l}{
    \parbox[l]{\linewidth}{
    \emph{Methods:} 
     FreGAN \citep{yang2022fregan}, WaveGAN \citep{yang2022wavegan}, MaskedGAN \citep{huang2022maskedgan}, Gen-co \citep{cui2022genco}, FAGAN \citep{cheng2024frequency}, SDTM \citep{yang2023sdtm}
       }
       }
\\
\arrayrulecolor{black!100}
\midrule
\rowcolor{gray!10}\multicolumn{2}{c}{{\bf \normalsize Meta-Learning} {(Sec.~\ref{ssec:review_metalearning})}}
\\[2pt]
{\bf \small Description:}& \parbox[l]{0.90\linewidth}{
Learn meta-knowledge from seen classes to improve generator learning for unseen classes.
} 
\\[2pt]
{\bf \small Task:} & \parbox[l]{0.92\linewidth}
{
uGM-1 \xmark \hspace{2pt} 
uGM-2 \xmark \hspace{2pt} 
uGM-3 \xmark \hspace{2pt} 
cGM-1 \xmark \hspace{2pt} 
cGM-2 \cmark \hspace{2pt} 
cGM-3 \xmark \hspace{2pt} 
IGM \xmark \hspace{2pt} 
SGM \xmark 
\hspace{5pt} 
{\bf \small Data constraint:} \hspace{-3pt} 
LD \xmark \hspace{1pt}
FS \cmark \hspace{1pt}
ZS \xmark 
}
\\
\arrayrulecolor{black!50}
\midrule
\multicolumn{2}{l}{
    \cellcolor{gray!0}\parbox[l]{\linewidth}{
    \textbf{1) Optimization:}
    Learn initialization weights from the seen classes as meta-knowledge
    to enable quick adaptation to unseen classes.
    }
    }
 \\[2pt]
\multicolumn{2}{l}{
    \parbox[l]{\linewidth}{
    \emph{Methods:} 
    GMN \citep{bartunov2018few}, FIGR \citep{clouatre2019figr}, Dawson \citep{liang2020dawson}, FAML \citep{phaphuangwittayakul2021faml}, CML-GAN \citep{phaphuangwittayakul2022cmlgan}
       }
       }
\\
\midrule
\multicolumn{2}{l}{
    \cellcolor{gray!0}\parbox[l]{\linewidth}{
    \textbf{2) Transformation:} 
    Learn sample transformations from the seen classes as meta-knowledge and use them for sample generation for unseen classes.
    }
    }
 \\[2pt]
\multicolumn{2}{l}{
    \parbox[l]{\linewidth}{
    \vspace{0.1cm}
    \emph{Methods:}
    \vspace{0.1cm}
    DAGAN \citep{antoniou2017dagan}, DeltaGAN \citep{hong2022deltagan}, Disco \citep{hong2022disco}, AGE \citep{ding2022age}, SAGE \citep{ding2023sage}, HAE \citep{li2022hae}, LSO \citep{zheng2023lso}, TAGE \citep{zhang2024tage}, CDM \citep{gupta2024conditional}, ISSA \citep{huang2021few}, MFH \citep{xie2022learning}
       }
       }
\\
\midrule
\multicolumn{2}{l}{
    \cellcolor{gray!0}\parbox[l]{\linewidth}{
    \textbf{3) Fusion:} 
    Learn to fuse the samples of the seen classes as meta-knowledge, and apply learned meta-knowledge to generation for unseen classes.
    }
    }
 \\[2pt]
\multicolumn{2}{l}{
    \parbox[l]{\linewidth}{
    \vspace{0.1cm}
    \emph{Methods:} 
    MatchingGAN \citep{hong2020matchinggan}, F2GAN \citep{hong2020f2gan}, LofGAN \citep{gu2021lofgan}, WaveGAN \citep{yang2022wavegan}, AMMGAN \citep{li2023ammgan}, MVSA-GAN \citep{chen2023iot}, EqGAN \citep{zhou2023eqgan}, SDTM \citep{yang2023sdtm}, SMR-CSL \citep{xiao2025semantic}, SAGAN \citep{aldhubri2024sagan}, F2DGAN \citep{zhou2024f2dgan}
       }
       }
\\
\arrayrulecolor{black!100}
\midrule
\rowcolor{gray!10}\multicolumn{2}{c}{{\bf \normalsize Modeling Internal Patch Distribution} {(Sec.~\ref{ssec:review_internalpatch})}}
\\[2pt]
{\bf \small Description:}& \parbox[l]{0.90\linewidth}{Learn the internal patch distribution within one image to generate diverse samples
with the
same visual content (patch distribution).} 
\\[2pt]
{\bf \small Task:} & \parbox[l]{0.92\linewidth}
{
uGM-1 \xmark \hspace{2pt} 
uGM-2 \xmark \hspace{2pt} 
uGM-3 \xmark \hspace{2pt} 
cGM-1 \xmark \hspace{2pt} 
cGM-2 \xmark \hspace{2pt} 
cGM-3 \xmark \hspace{2pt} 
IGM \cmark \hspace{2pt} 
SGM \xmark 
\hspace{5pt} 
{\bf \small Data constraint:} \hspace{-3pt} 
LD \xmark \hspace{1pt}
FS \cmark \hspace{1pt}
ZS \xmark 
}
\\
\arrayrulecolor{black!50}
\midrule
\multicolumn{2}{l}{
    \cellcolor{gray!0}\parbox[l]{\linewidth}{
    \textbf{1) Progressive Training:} 
    Train a generative model progressively 
    to learn the patch distribution at different scales/ noise levels.
    }
    }
 \\[2pt]
\multicolumn{2}{l}{
    \parbox[l]{\linewidth}{
    \emph{Methods:}
    SinDiffusion \citep{wang2022sindiffusion}, SinDDM \citep{kulikov2023sinddm}, Deff-GAN \citep{kumar2023deffGAN}, BlendGAN \citep{kligvasser2022blendgan}, SinGAN \citep{shaham2019singan}, ConSinGAN \citep{hinz2021consingan}, CCASinGAN \citep{wang2022ccasingan}, PromptSDM \citep{park2024promptsdm}, LatentSDM \citep{han2024latentsdm}, SD-SGAN \citep{yildiz2024sdsgan}, SA-SinGAN \citep{chen2021sa}, ExSinGAN \citep{zhang2021exsingan}, TcGAN \citep{jiang2023tcgan}, RecurrentSinGAN \citep{he2021recurrent}
       }
       }
\\
\midrule
\multicolumn{2}{l}{
    \cellcolor{gray!0}\parbox[l]{\linewidth}{
    \textbf{2) Non-progressive Training:} 
    Train a generative model
    on the same scale/ noise but with changes to the model’s architecture.
    }
    }
 \\[2pt]
\multicolumn{2}{l}{
    \parbox[l]{\linewidth}{
    \emph{Methods:}
        SinFusion \citep{nikankin2022sinfusion}, One-Shot GAN \citep{sushko2021oneshotgan}, PetsGAN \citep{zhang2022petsgan}
       }
       }
\\
\arrayrulecolor{black!100}
\bottomrule
\end{longtable}
\end{spacing}

\def\checkmark{\tikz\fill[scale=0.4](0,.35) -- (.25,0) -- (1,.7) -- (.25,.15) -- cycle;} 

\subsection{Generative Modeling under Data Constraint: Challenges}
\label{ssec:challenges}

\subsubsection{Challenges for Training Generative Models under Data Constraint}
Data constraints typically introduce additional challenges and 
amplify existing ones when training generative models. 
Here, we delve into the challenges of training GM-DC.
These limitations include pervasive issues of overfitting and frequency bias which are commonly observed across various approaches.
Additionally, knowledge transfer between domains brings forth specific problems including the proximity between source and target domains and the transfer of incompatible source knowledge.
As shown in Fig.~\ref{fig:works_statistics}, 
a lot of 
works directly rely on knowledge transfer as a mainstream method to tackle GM-DC, and 
a number of  
works propose methods based on other approaches that are compatible with transfer learning.

\textbf{Overfitting to Training Data.}
In machine learning, overfitting is a common issue
when powerful models start to 
 memorize the training data instead of learning the
generalizable semantics \citep{santos2022avoiding}.
In generative modeling, 
the overfitting problem exacerbates 
under data constraints due to the high capacity of current generative models \citep{noguchi2019bsa, liu2021fastgan, karras2020ada}. 
When limited training data is available,  generative models may simply remember the training data \citep{li2020ewc, ojha2021cdc} and learn to generate the exact training samples \citep{zhao2022adam} instead of capturing the data distribution.
Furthermore, under data constraints, generative modeling is more prone to mode collapse \citep{tran2021dag}, i.e., the generators learn only a limited set of modes and fail to capture other modes of the data distribution, resulting in limited diversity 
in generated samples \citep{yu2022understanding, nguyen2023rethinking}.

\textbf{Frequency Biases.}
Generative models are notorious for their spectral bias \citep{rahaman2019nnfrequncybias, khayatkhoei2022ganfrequencybias}, \ie tendency to prioritize fitting low-frequency components while disregarding high-frequency components within a data distribution \citep{durall2020watchupconvolution, tancik2020fourier, chandrasegaran2021closer}. 
The exclusion of these high-frequency components which encode intricate image details \citep{gonzales1987digital} can significantly impact the quality of generated samples, \ie, accurate modeling of high-frequency details is critical in various fields including medical imaging (X-rays, CT-scans, MRIs), satellite/ aerial imaging, astrophotography, and art restoration.
This issue becomes more severe under limited data \citep{yang2022fregan, yang2022wavegan} as even advanced network structures tailored for such scenarios \citep{liu2021fastgan} struggle to maintain the desired level of details in generated samples.

\begin{figure}
    \centering
    \includegraphics[trim={0 263pt 0 210pt},clip,width=\textwidth]{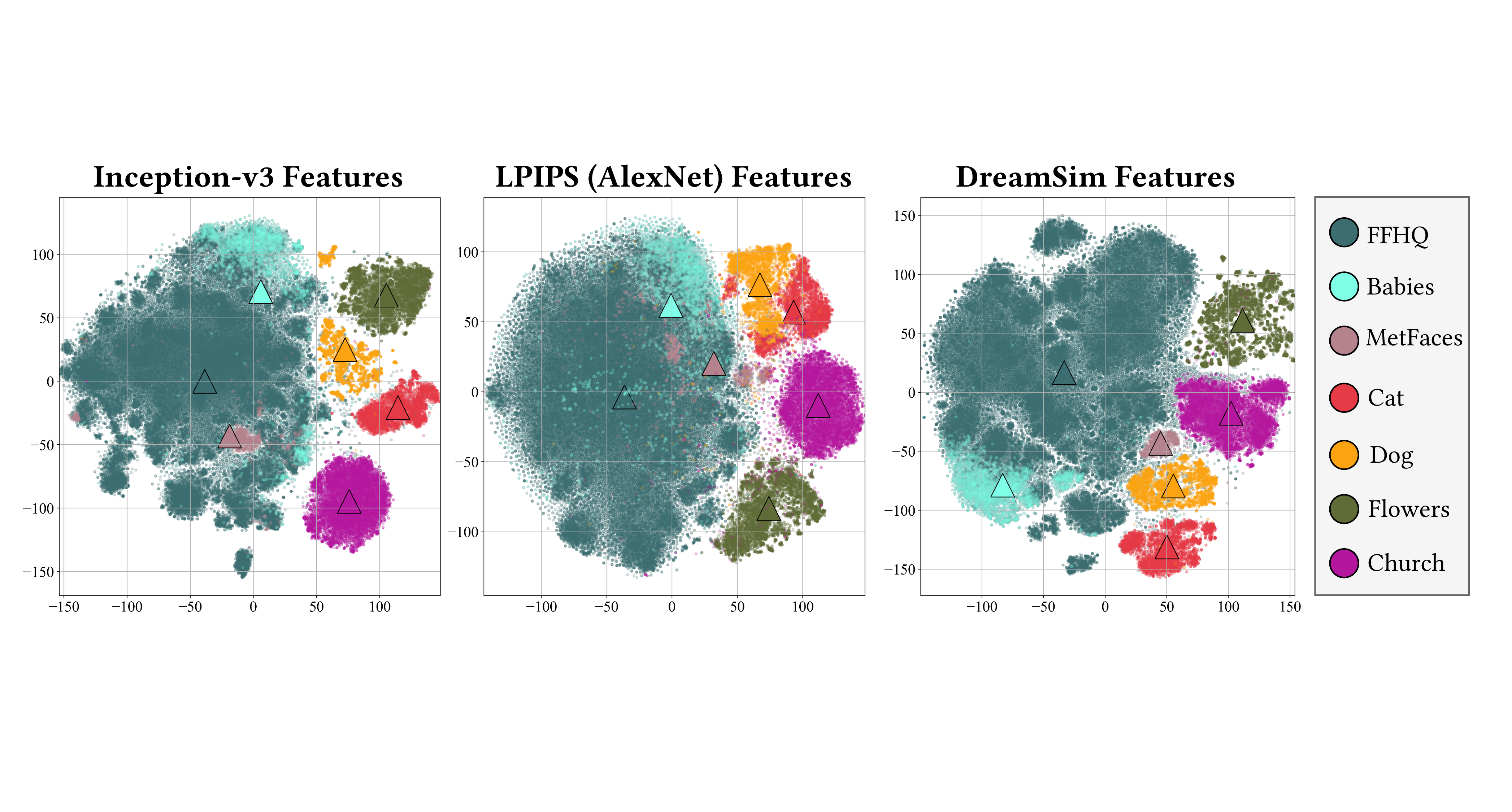}
\caption{
\textbf{Source-target domain proximity visualization indicates that distant/ remote target domains have not been explored in GM-DC setups and are very challenging.}
We use FFHQ \citep{karras2019style} as the source domain.
We show 
source-target
domain proximity qualitatively by
visualizing {\bf Inception-v3 (Left)} \citep{szegedy2016labelsmoothing}, {\bf LPIPS (Middle)} \citep{zhang2018lpips} 
and {\bf DreamSim (Right)} \citep{fu2023dreamsim}  features. 
For feature visualization, we use t-SNE \citep{JMLR:v9:vandermaaten08a_tsne} and show centroids ($\bigtriangleup$) for all domains. 
We clearly show using feature visualizations 
that additional setups -- Flowers \citep{nilsback2008oxford_flower} and Church \citep{yu2015lsun} -- represent target domains that are remote from the source domain (FFHQ) compared to target domains used in the literature. 
This indicates that the exploration of distant/ remote target domains under GM-DC setups has not been pursued and poses notable challenges (Fig. \ref{fig:proximity-measurements-and-10-shot-flowers}).
Best viewed in color.
  }
\label{fig:proximity-visualization}
\vspace{-0.7cm}
\end{figure}

\begin{figure}
   \centering
     \includegraphics[width=1.0\linewidth]{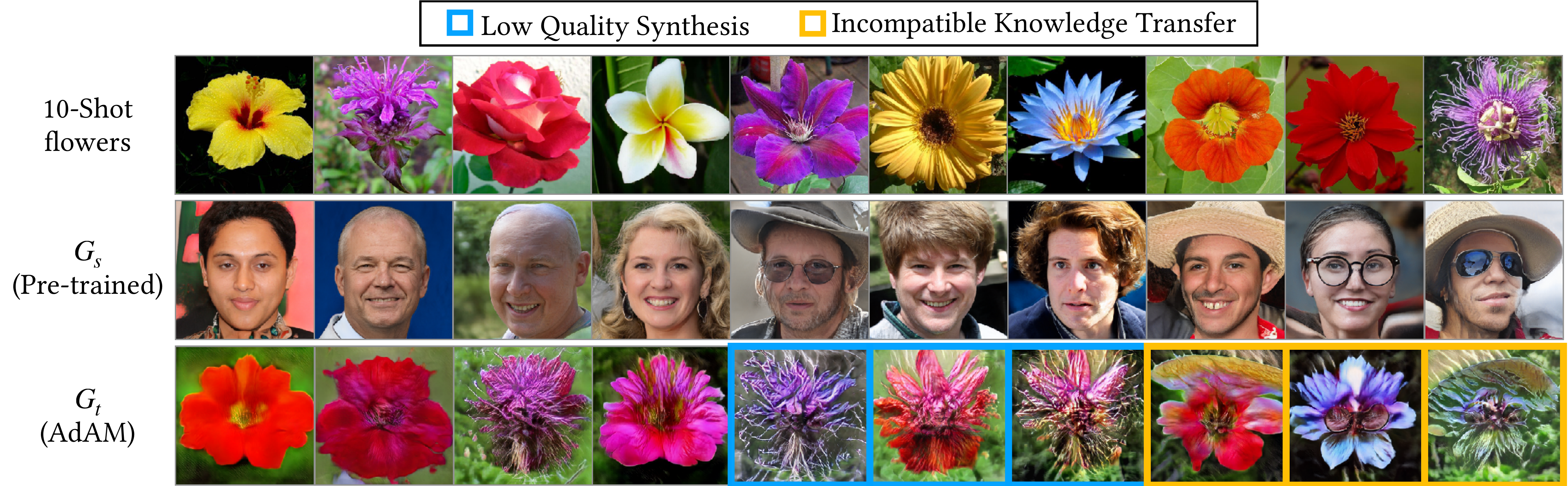}
\caption{
\textbf{
Knowledge transfer under distant/ remote target domain (Human face $\rightarrow$ Flowers) suffers from low synthesis quality and incompatible knowledge transfer.
}
We show 10-shot adaptation results for FFHQ $\rightarrow$ Flowers using AdAM \citep{zhao2022adam}. 
The FID for the 10-shot adaptation using AdAM is 124.46.
We highlight multiple instances of {low synthesis quality} and {incompatible knowledge transfer} (\ie glasses, hat from FFHQ to flowers), showing that GM-DC modeling of remote target domains poses significant challenges.
Best viewed in color.
}
\label{fig:proximity-measurements-and-10-shot-flowers}
\end{figure}

\textbf{Modeling Distant/ Remote Target Domains under GM-DC Setups.}~
Substantial number of GM-DC tasks rely on the transfer learning principle (uGM-2, uGM-3, cGM-2, cGM-3, SGM),  
which aims to enhance the generative capabilities for a target domain by leveraging the knowledge of a generator pre-trained on a large and diverse source domain (See Fig. \ref{fig:sankey}).
A significant amount of research has been
focused on target domains that are semantically/ perpetually similar to the source domain, \eg, learn to generate Baby faces using a pre-trained generator trained on Human faces \citep{gal2022stylegannada, li2023hda, liu2025air}. 
In particular, when dealing with GM-DC setups involving significant domain shifts between the source and target domains (Human Faces$\rightarrow$Animal Faces), many proposed methods fail to outperform a basic fine-tuning approach \citep{zhao2022adam}.  This is due to these methods prioritizing knowledge preservation from the source domain/ task, overlooking the adaptation step to the target domain \citep{zhao2022adam}.
Adaptation-aware algorithms 
have characterized source$\rightarrow$target domain proximity \citep{zhao2022adam} and addressed GM-DC setups with pronounced domain shifts between the source and target domains (Human Faces$\rightarrow$Animal Faces) \citep{zhao2022adam, zhao2023rick}.
To understand the concept of distant/ remote target domains,
we additionally introduce two remote target domains that further exhibit a considerable degree of domain shifts: 
i) Human Faces (FFHQ) \citep{karras2019style}$\rightarrow$ Flowers \citep{nilsback2008oxford_flower}, 
ii) Human Faces (FFHQ) \citep{karras2019style}$\rightarrow$Church \citep{yu2015lsun}.
Domain proximity visualization 
is shown in Fig. \ref{fig:proximity-visualization}.
In particular, we conducted a GM-DC experiment (uGM-2) to adapt a pre-trained Human face (FFHQ) generator to Flowers under 10-shot setup using AdAM \citep{zhao2022adam}, 
obtaining a FID value of 124.46. 
Adaptation results are shown in Fig. \ref{fig:proximity-measurements-and-10-shot-flowers}.
As one can observe, multiple instances of {low quality synthesis} are observed in AdAM \citep{zhao2022adam}.
In summary, we remark that modeling distant/ remote target domains remains an important and challenging area for GM-DC.

\textbf{Identifying and Removing Incompatible Knowledge Transfer.}~
Another challenge with 
leveraging source domain's knowledge for GM-DC tasks is  
incompatible knowledge transfer, which is discovered in \citet{zhao2023rick}.
In particular, many methods may transfer knowledge that is  
incompatible with the target domain, \eg hat from source domain FFHQ to target domain flowers, significantly degrading the realisticness of the generated samples.
In Fig. \ref{fig:proximity-measurements-and-10-shot-flowers}, 
we show multiple examples of {incompatible knowledge transfer} using AdAM for 10-shot flower adaptation.
Although some recent effort has been invested in identifying and proactively truncating incompatible knowledge transfer \citep{zhao2023rick} in Human Faces $\rightarrow$ Animal Faces adaptation setups, it is worth noting that identifying and removing incompatible knowledge remains a critical and demanding area in GM-DC.

\subsubsection{Challenges on Selecting Samples for GM-DC}
\begin{figure}[!t]
    \includegraphics[width=0.99\linewidth]{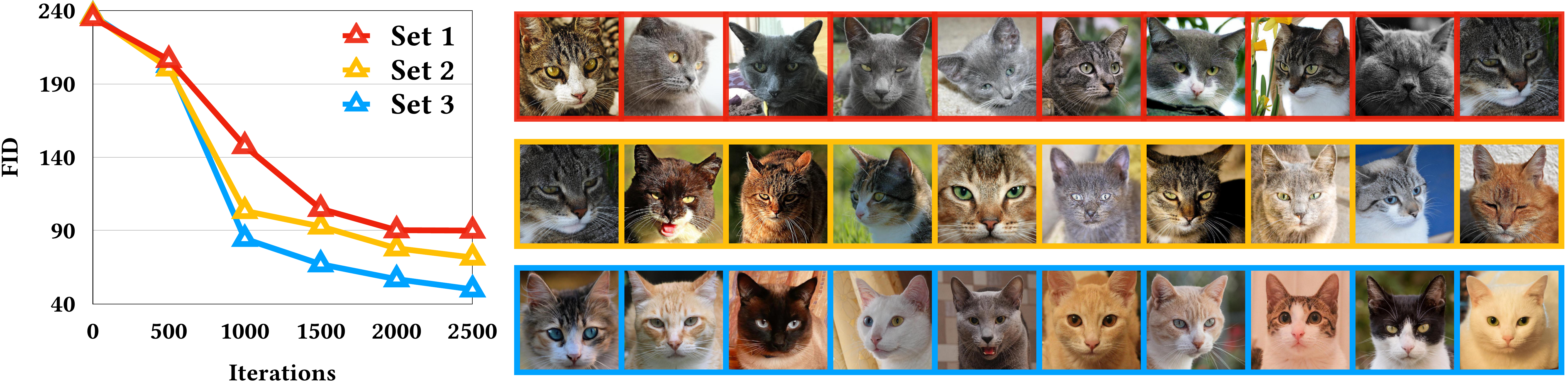}
\caption{
\textbf{
Sample selection for GM-DC remains challenging and relatively unexplored.}
We use AdAM \citep{zhao2022adam} to adapt a pre-trained StyleGAN2 on FFHQ to AFHQ-Cat dataset \citep{choi2020starganv2} using 3 random sets of 10-shot data \textit{(Right)}.
We report FID results during training \textit{(Left)} for these sets of 
data.
Following \citet{zhao2022adam}, FID is measured between 5000 generated samples and the entire AFHQ-Cat dataset consisting of 5115 samples.
We use clean-FID library \citep{Parmar_2022_CVPR}, obtaining
FID values of {90.0}, {71.6} and {49.9} 
for Sets 1, 2, and 3 respectively at iteration=2500. 
As indicated by FID trends,
the generative capabilities of GM-DCs are drastically influenced by sample selection.
Best viewed in color.
}
\label{fig:challenges-data-selection} 
\vspace{-0.3cm}
\end{figure}

~
Although considerable research effort has been invested in developing algorithms for GM-DC, the task of sample selection for GM-DC remains a challenging and relatively unexplored area.
It is essential that the samples selected for GM-DC should represent the target domain.
In particular, we observe significant variation in performance with different selection of target samples as the training datasets in  GM-DC.
We perform a 10-shot \textit{data-centric} GM-DC experiment using AdAM \citep{zhao2022adam} to emphasize the importance of sample selection in GM-DC. Following \citet{zhao2022adam, zhao2023rick}, we use AFHQ-Cat dataset \citep{choi2020starganv2} and select 3 random sets of 10-shot cat data for GM-DC. 
Data and 10-shot adaptation FID results are shown in Fig. \ref{fig:challenges-data-selection}.
We obtain FID values of 90.0, 71.6 and 49.9 for Sets 1, 2 and 3 respectively (iteration=2500). 
This study provides evidence that sample selection plays a vital role in determining the capabilities of GM-DC.
Specifically, due to cost/ privacy concerns, the role of sample selection is critical in applications including biomedical imaging, satellite/ aerial imaging and remote sensing. In summary, sample selection for GM-DC holds significant importance and remains an area with limited investigation thus far.

\subsubsection{Challenges in Evaluating Generative Models under Data Constraint}
The assessment of generative modeling capabilities presents 
lots of challenges, encompassing both objective and subjective evaluation \citep{kynkaanniemi2023the}.
These issues are aggravated under low-data regimes resulting in the evaluation of GM-DC to be challenging and an active topic of research.
In contemporary GM-DC literature, sample quality and diversity are used as the main attributes for evaluating generation capability.
A summary of prominent metrics for GM-DC is included in Tab. \ref{tab:evaluation-metrics}.

Existing GM-DC evaluation metrics present multiple challenges:
i) Statistical measures including FID, KID, IS, FID\textsubscript{CLIP} lose their significance when dealing with 
setups
where there is an extreme scarcity (Few-shots) or complete absence (Zero-shot) of target domain data. For example, when the reference distribution contains only 10 real images, the mean and trace components of FID are not statistically significant.
ii) Although human judgment/ user feedback is used for the subjective evaluation of GM-DC, the absence of a unified framework/ protocol for such evaluation strategy results in inadequacy when comparing the generative capabilities of different GM-DC models.
iii) The over-reliance on objective GM-DC measures  on deep features extracted from pre-trained networks remains challenging and relatively unexplored. 
For example, FID, KID, and IS use features extracted from an Inception model trained on ImageNet-1K \citep{deng2009imagenet}; LPIPS, and Intra-LPIPS, 
use features extracted from models trained on BAPPS \citep{zhang2018lpips} dataset.
Although these pre-trained models effectively function as general-purpose feature extractors, their ability to capture properties/ attributes of out-of-domain data to objectively quantify the capabilities of GM-DC requires more investigation, \eg medical images.

In summary, the area of evaluation measures for GM-DC cannot be overstated, as it remains critical and challenging.

\begin{table}[!t]
    \centering
    \fontsize{7pt}{7pt}\selectfont
    \caption{Common metrics used for evaluating GM-DC works \citep{zhao2022adam, teo2023fairtl, liu2025air}.
    {\bf LD}: \underline{L}imited-\underline{D}ata, {\bf FS}: \underline{F}ew-\underline{S}hot, {\bf ZS}: \underline{Z}ero-\underline{S}hot. 
    \xmark/\cmark denotes the absence/presence, respectively.
    }
    \begin{tabularx}{0.7\columnwidth}{Xccc}
    \toprule
    \textbf{Metric} & \textbf{LD} & \textbf{FS} & \textbf{ZS} \\ 
    \midrule
    FID \citep{heusel2017twotimescale}/ FID\textsubscript{CLIP} \citep{kynkaanniemi2023the} & \cmark & \cmark & \xmark \\
    \midrule
    KID \citep{binkowski2018demystifying} & \cmark & \cmark & \xmark \\
    \midrule
    IS \citep{salimans2016improved} & \cmark & \cmark & \xmark \\
    \midrule
    Intra-LPIPS \citep{ojha2021cdc} & \xmark & \cmark & \cmark \\
    \midrule
    SIFID \citep{shaham2019singan} & \xmark & \cmark & \cmark \\
    \midrule
    Image/ Text Similarity \citep{gal2022textualinversion} & \xmark & \cmark & \cmark \\
    \midrule
    User Feedback & \cmark & \cmark & \cmark \\
    \bottomrule
    \end{tabularx}
    \label{tab:evaluation-metrics}
\end{table}

\section{Comprehensive Review}
\label{sec:comprehensive_review}

In this section, 
first, we  will present our proposed taxonomy of approaches for GM-DC which systematically categorizes and organizes GM-DC methods under seven approaches (Tab.~\ref{tab:approaches}) based on the principal ideas of these methods.
Then, we will discuss individual GM-DC methods organized under our proposed taxonomy.

\noindent
{\bf Our Proposed Taxonomy of Approaches for GM-DC} categorizes  GM-DC methods into seven groups:
\begin{enumerate}
    \item {\bf Transfer Learning:} 
    In GM-DC, transfer learning (TL) aims to improve the learning of the generator for the target domain using the knowledge of a generator pre-trained on a source domain (with numerous and diverse samples). 
    For example, some methods under this category use the knowledge of a Stable Diffusion \cite{rombach2022latentdiffusion} pre-trained on LAION-400M \cite{schuhmann2021laion400m} to learn to generate diverse and high-quality samples of particular subject(s),
    given only a few images of the subject(s) \cite{ruiz2023dreambooth, gal2022textualinversion, kumari2023mcc}.
    Major challenges for TL-based GM-DC are to identity, select, and preserve suitable knowledge of the source generator for the target generator.
    Along this line, there are six subcategories: i) {\em Regularization-based Fine-tuning}, explores regularizers to preserve suitable source generator's knowledge to improve learning target generator; 
    ii) {\em Latent Space},
    explores transforming/ manipulating the source generator's latent space to identify suitable knowledge for adaptation; 
    iii) {\em Modulation},  freezes and transfers weights of the source generator to the target generator and adds trainable modulation weights on top of frozen weights to increase the adaptation capability to the target domain; iv) {\em Natural Language-guided}, uses natural language prompt and supervision signal from language-vision models to adapt source generator to target domain; v) {\em Adaptation-Aware}, identifies and preserves the source generator's knowledge that is important to the adaptation task; vi) {\em Prompt Tuning}, is an emerging idea that freezes the weights of the source generator and learns to generate visual prompts (tokens) to guide generation for the target domain.

    \item {\bf Data Augmentation:} Augmentation aims to improve GM-DC by increasing coverage of the data distribution with applying various transformations $\{ T_k \}_{k=1}^{K}$ to available data. 
    For example, within this category, some works augment the available limited data to train an unconditional StyleGAN2 \citep{karras2020analyzing} using the 100-shot Obama dataset or train a conditional BigGAN \citep{brock2019biggan} with only 10\% of the CIFAR-100 dataset.
    A major challenge of these approaches is augmentation leakage, 
    where the generator learns the augmented distribution, \eg, generating rotated/ noisy samples.
    There are three representative categories: 
    i) {\em Image-Level Augmentation}, applies the transformations on the image space; 
    ii) {\em Feature-Level Augmentation}, applies the transformations on the feature space; 
    iii) {\em Transformation-Driven Design}, leverages the information on each individual transformation $T_k$
    to design an efficient learning mechanism.

    \item {\bf Network Architectures:} These approaches design specific architectures for the generators to improve their learning under data constraints. 
    Some works in this category design shallow/ sparse generators to prevent overfitting to training data due to over-parameterization.
    The primary challenge is that when endeavoring to design a new architecture, the process of discovering the optimal hyperparameters can be laborious.
    There are three major types of architectural designs for GM-DC: 
    i) {\em Feature Enhancement}, introduces additional modules to enhance/ retain the knowledge within feature maps; 
    ii) {\em Ensemble Large Pre-trained Vision Models}, leverages large pre-trained vision models to aid more accurate generative modeling; iii) {\em Dynamic Network Architecture}, evolves the architecture of the generative model during training to compensate for data constraints.
    
    \item {\bf 
    Multi-Task Objectives:} 
    These approaches modify the learning objective of the generative model by introducing additional task(s) to extract generalizable representations and reduce overfitting under data constraints.
    As an example, some works define a pretext task based on contrastive learning \citep{he2020momentum} to pull the positive samples together and push negative ones away in addition to the original generative learning task.
    The efficient integration of the new objective with the generative learning objective could be challenging under data constraints.
    These works can be categorized into several approaches: 
    i) {\em Regularizer}, adds an additional learning objective as a regularizer to prevent an undesirable behavior during training a generative model under data constraints. Note that this category is different from regularizer-based fine-tuning, as the latter aims to preserve source knowledge, but the former is for training without a source generator; 
    ii) {\em Contrastive Learning}, adds the learning objective related to a pretext task to enhance the learning process of the generative model using an additional supervision signal from solving this pretext task; 
    iii) {\em Masking},
    introduces alternative learning objective by masking a part of the image/ information to improve generative modeling by increasing the task hardness and preventing learning the trivial solutions; 
    iv) {\em Knowledge Distillation}, introduces an additional learning objective that enforces the generator to follow a strong teacher; 
    v) {\em Prototype Learning}, emphasizes learning prototypes for samples/ concepts within the distribution as an additional objective; 
    vi) {\em Other Multi-Task Objectives}, include co-training, patch-level learning, and using diffusion to enhance 
    generation.

    \item {\bf Exploiting Frequency Components:} Deep generative models exhibit frequency bias tending to ignore high-frequency signals as they are hard to generate \citep{schwarz2021frequency}. Data constraints can exacerbate this problem \citep{yang2022fregan}. 
    The approaches in this category aim to improve frequency awareness of the generative models by leveraging frequency components during training. 
    For instance, certain approaches employ Haar Wavelet transform to extract high-frequency components from the samples. 
    These frequency components are then fed into various layers using skip connections, aiming to alleviate the challenges associated with generating high-frequency details.
    Despite its effectiveness, utilizing frequency components for GM-DC has not been thoroughly investigated. The performance can be enhanced by incorporating more advanced techniques for extracting frequency components.
    
    \item {\bf Meta-Learning:} These approaches create sample generation tasks with data constraints for the seen classes, and learn the meta-knowledge---knowledge that is shared between all tasks---across these tasks during meta-training. This meta-knowledge is then used in improving generative modeling for the unseen classes with data constraints.
    For instance, some studies, as meta-knowledge, learn to fuse the samples from 
    the seen categories $C_{seen}$ of the Flowers dataset \citep{nilsback2008oxford_flower} for sample generation.
    This meta-knowledge enables the model to generate new samples from unseen classes $C_{unseen}$ within the same dataset ($C_{seen} \cap C_{unseen} = \emptyset$) by fusing only 3 samples from each class.
    Note that as these works employ episodic learning within a generative framework, the training stability can be impacted.
    Approaches within this line can be classified into three categories: 
    i) {\em Optimization}, initializes the generative model with weights learned on the seen classes as meta-knowledge, to enable quick adaptation to unseen classes with limited steps of the optimization;
    ii) {\em Transformation}, learns cross-category transformations from the samples of the seen classes as meta-knowledge and applies them to available samples of the unseen classes to generate new samples; 
    iii) {\em Fusion}, learns to fuse the samples of the seen classes as meta-knowledge, and applies learned meta-knowledge to sample generation by fusing samples of the unseen classes.

    \item {\bf Modeling Internal Patch Distribution:} These approaches aim to learn the internal patch distribution within one image (in some cases a few images), and then generate diverse samples that carry the same visual content (patch distribution) with an arbitrary size, and aspect ratio. 
    As an example, some works train a Diffusion Model using a single image of the ``Marina Bay Sands'', and after training, the Diffusion Model can generate similar images, but include additional towers topped by the similar ``Sands Skypark''.
    However, a major limitation of these methods lies in the fact that for every single image, usually a separate generative model is trained from scratch, 
    neglecting the potential for efficient training through knowledge transfer in this context.
    Approaches proposed along this line can be categorized into two major groups: i) {\em Progressive Training}, progressively trains a generative model to learn the patch-distribution at different scales or noise levels; ii) {\em Non-Progressive Training}, learns a generative model at a single scale by implementing additional sampling techniques or new model architectures.  
\end{enumerate}
In what follows we delve into detailed descriptions of the approaches within each category.
\vspace{-0.3cm}
\subsection{Transfer Learning}
\label{ssec:review_transferlearning}
Transfer Learning (TL) is a major approach for GM-DC. 
Given a source generator $G_s$ (for GANs, both $G_s$ and $D_s$) pre-trained on a large and diverse source domain $\mathcal{D}_s$, these approaches aim to learn an adapted generator to the target domain $G_t$ by initializing the weights to those of the source generator.

\subsubsection{Regularizer-based Fine-Tuning}
\label{ssec:Regularizer-based Fine-Tuning}

{\bf Early evidence of transfer learning benefits.} 
Early works in this category explore the effectiveness of transfer learning in the context of generative modeling with limited data, analyzing different aspects such as the effect of source and target domain distance, the size of the target domain dataset, and even the statistics of the distribution on which the source model was trained.
TGAN \citep{wang2018tgan} is the first systematic study to evaluate transfer learning in GANs. 
TGAN shows that transfer learning reduces the convergence time and improves generative modeling under limited data. 
The knowledge transfer is performed by using the source GAN for initializing the weights of the target GAN, followed by fine-tuning the weights on target data.
TGAN \citep{wang2018tgan} demonstrates that: i) transferring $D$ is more important than $G$, while transferring both $G$ and $D$ gives the best results; ii) transfer learning performance degrades by increasing the distance between source and target domains or decreasing the number of samples from target domain; iii) to select a pre-trained GAN for a target domain, in addition to a smaller distance, more dense source domains are preferable. 
As an example, for the Flower \citep{nilsback2008oxford_flower} target domain, surprisingly, a GAN pre-trained on semantically unrelated LSUN Bedrooms \citep{yu2015lsun} is shown to be among the best sources \citep{wang2018tgan}.
W\textsuperscript{3} \citep{grigoryev2022when} revisits the transfer learning in GANs with a modern structure---StyleGAN2-ADA \citep{karras2020analyzing, karras2020ada} instead of WGAN-GP \citep{gulrajani2017improved} used in TGAN.
Results in \citet{grigoryev2022when} suggest that for SOTA GANs, it is beneficial to transfer the knowledge from sparse and diverse sources (pre-trained StyleGAN2 on ImageNet) rather than dense and less diverse ones.
A major limitation of TGAN is that under limited data, simply fine-tuning the whole generator destroys a considerable portion of the general knowledge obtained on the source domain. Almost all of the following works aim to address this by using different approaches to preserve the knowledge of the source generator.

{\bf Preserving source knowledge via parameter-efficient fine-tuning.}
Approaches immediately after TGAN performed this knowledge preservation through parameter-efficient fine-tuning, where a small set of parameters was fine-tuned using data from the target domain while the remaining parameters were kept frozen.
BSA \citep{noguchi2019bsa} only updates scale and shift parameters in batch normalization (BN) layers during fine-tuning to prevent overfitting to limited data.
FreezeD \citep{mo2020freezed} hypothesizes that as $D$ performs the discriminative task during training a GAN, based on common knowledge in discriminative learning \citep{yosinski2014transferable}, its early layers encode general knowledge which is shared between source and target domains. 
Therefore, this general knowledge is preserved during adaptation by freezing early layers of $D$.
cGANTransfer \citep{shahbazi2021efficient} assumes that the pre-trained $G$ is conditioned on class labels using BN parameters, \ie each class has its own BN parameters \citep{brock2019biggan}. Then, explicit knowledge propagation from seen classes to unseen classes is enforced by defining the BN parameters of the unseen classes to be the weighted average of the BN parameters of seen classes.
SVD \citep{robb2020svd} applies singular value decomposition \citep{van1976generalizing}, and only updates the singular values that are related to changing entanglement between different attributes within data. 

{\bf Constraining adaptation to prevent source-knowledge distortion.}
Later, more advanced approaches were proposed by exploring the causes of source knowledge distortion during fine-tuning and then proposing constraints to prevent it.
EWC \citep{li2020ewc} aims to preserve the diversity of the source GAN during adapting to a target domain with only a few samples, \eg, 10-shot. The importance of each parameter in source GAN is measured by Fisher Information (FI), and the change on each parameter during adaptation is penalized based on its importance, \eg, change over important parameters is penalized more.
CDC \citep{ojha2021cdc} aims to keep the diversity of the generated samples using a cross-domain correspondence loss.
Specifically, first, a batch of $N+1$ latent codes are sampled for image generation: $\{ G(z_0), ..., G(z_N) \}$.
Then, using $G(z_0)$ as a reference, the similarity of generated samples to the reference is measured for the generator before and after adaptation, resulting in two $N-way$ probability vectors. The diversity is preserved by adding the KL divergence between these two probability vectors to the standard loss as a regularizer.
MaskD~\citep{zhu2022few} applies random masks to extracted features of $D$, on top of CDC \citep{ojha2021cdc}, to prevent overfitting. 
DDPM-PA \citep{zhu2022few_dm} uses a pairwise adaptation method similar to CDC for adapting diffusion models to the new domain.
RSSA \citep{xiao2022rssa} extends the cross-domain consistency idea of the CDC \citep{ojha2021cdc} to a more constrained form by preserving the structural similarity of the samples before and after adaptation.
ProSC \citep{moon2023prosc} extends RRSA by performing a progressive adaptation to the target domain in $N$ iterations instead of a single adaptation to reduce the gap between pairs of domains.
CSR \citep{gou2023csr} uses a similar idea to CDC but applies semantic loss directly to the spatial space, \ie, generated images with $G_s$ and $G_t$.
DWSC \citep{hou_dynamic} proposes the dynamic weighted semantic correspondence between the source and target generator during adaptation to preserve the diversity.
SSCR \citep{israr2024few} extends CDC by computing more accurate similarity scores by leveraging frozen Siamese networks \citep{chen2020simple}.

{\bf Contrastive objectives to preserve diversity.}
A line of work employs contrastive learning during fine-tuning to prevent knowledge distortion when adapting generative models to the target domain. 
For instance, $C^3$ \citep{Lee2021C3}, DCL \citep{zhao2022dcl}, and CtlGAN \citep{wang2022ctlgan} aim to preserve diversity by applying contrastive learning.
Assuming $G_s(z_i)$ as an anchor point, the generated sample for the same latent code with the adapted generator ($G_t(z_i)$) is considered a positive pair, and the generated samples with the adapted generator for other latent code values ($G_t(z_j)$, $i \neq j$) are considered as negative pairs. 
Additionally, DCL applies similar contrastive learning to the $D$.
Based on the contrastive learning idea of DCL \citep{zhao2022dcl} and LeCam regularizer \citep{tseng2021lecam}, 
 CLCR \citep{zhang2023clcr} proposes a transfer learning approach for generating diverse and high-quality COVID-19 CT images for a target group.

{\bf Latent inversion for stronger preservation.}
Some works invert images from the target domain into the latent space of the pre-trained generator to develop a more accurate regularizer for knowledge preservation during fine-tuning.
JoJoGAN \citep{chong2022jojogan} addresses one-shot image generation using the style space of StyleGAN2.
First, GAN inversion is used to find the corresponding style code of the reference image. Then, style mixing is used to generate a set of style codes, and generated images with these styles are used for GAN adaptation.
GenOS \citep{zhang2022generalizedoneshot} includes entity transfer with some related entity mask using an auxiliary network.
D\textsuperscript{3}-TGAN
\citep{wu2023d3tgan} first inverts each target sample into the latent code space of the source GAN.
Then, the maximum mean discrepancy between the features of the source $G$ for inverted code and features of the adapted GAN for a random latent code is used as a regularizer.
F\textsuperscript{3} \citep{yuichi2023_fewshot} proposes a faster method for image generation with features of a specific group. First, a GAN inversion of target images is applied, and then PCA is leveraged as a feature extraction strategy to render features of the target group.

{\bf Diverse and emerging regularization ideas.}
More recent advances in regularizer-based fine-tuning employ a wide array of diverse and scattered ideas for knowledge preservation.
Focusing on a somewhat different goal,
FairTL \citep{teo2023fairtl, teo2024fairtl} adopts transfer learning in GANs to train a fair generative model \wrt a sensitive attribute (SA) using a limited fair dataset.
To model complex distributions like ImageNet, IC-GAN \citep{casanova2021icgan} learns data distribution as a mixture of conditional distributions.
This enables IC-GAN to generate images from unseen distributions, by just changing the conditioning instances on the target samples.
KDFSIG \citep{hou2022exploitingkd} exploits the knowledge distillation idea by treating the source model as a teacher and the target model as a student.
DOGAN \citep{hu2024DOGAN} proposes frequency-based 
segregation of the generation and the discrimination process. 
FAGAN \citep{cheng2024frequency} introduces two frequency regularizers between the source and adapting generator in a one-shot adaptation setup.
AnyDoor \citep{chen2024anydoor} proposes the use of a discriminative-ID and frequency-aware extractor to characterize the subject. 
SmoothSim \citep{sushko2023smoothsim}  proposes to preserve  the smoothness of the source GAN in target GAN via a regularizer that estimates the Jacobian matrix of the generator in different layers.
DPMs-ANT \citep{wang2024bridging} uses KL-divergence between the noisy output of the source and fine-tuned diffusion model at time step $t$ (not final output) to prevent overfitting when adapting a diffusion model to a target domain in a few-shot setup.
It also introduces an adversarial noise selection mechanism to reduce training iterations.
Focusing on
efficient transfer learning in diffusion models, 
TAN \citep{wang2023efficient} proposes a KL divergence minimization 
between the gap 
between the source and target domains.
To address the overfitting problem caused by multiple denoising steps in diffusion models, they also propose an adaptive noise selection scheme to reduce the number of denoising steps.
FDDC \citep{hu2023phasic} gives more weight for content information in early steps of the denoising process by fusing the content from the images generated by the source generator.
To address catastrophic forgetting, FPTGAN \citep{zhang2023fptgan} proposes a trust-region optimization to smooth the fine-tuning dynamics by adjusting the noise distribution and a SVD-based approach to detect and mitigate mode collapse.
StyleDomain \citep{alanov2023styledomain} analyzes the structure of StyleGAN2 \citep{karras2020analyzing} for lightweight adaptation of a pre-trained model to target domains.

{\bf Beyond single-domain adaptation: expansion and hybrid targets.}
Finally, several works extend the setting itself. In \citet{nitzan2023domain}, domain expansion (DE) of Image Generators is studied.
Instead of transforming the entire generator from
a source domain to a target domain, DE expands the generator to include new data domains, based on their discovery that traversing the latent space of generative models along some directions changes the image significantly while traversing
others has no perceptible effect (the dormant direction).
HDA \citep{li2023hda} studies a task beyond domain adaptation to a certain concept, aiming to adapt to a hybrid domain that integrates attributes from several concepts. Specifically, the hybrid domain is represented by the mean embedding of several domains. They utilized the directional loss and distance loss to fine-tune the pre-trained GAN.
Def-DINO \citep{zhou2024deformable} suggests that the features learned in the DINO vision transformer \citep{caron2021dino} via self-distillation are more powerful for one-shot adaptation than the features learned by vision-language models like CLIP.
In \citet{jiang2023personalized}, CVD-GAN, a personalized image generation is proposed for color vision deficiency (CVD).
SACP \citep{he2024few} trains a translation module to detect the content and style and then adds a regularizer to preserve the content while adapting the style during fine-tuning a pre-trained GAN.
In \citet{kim2023datid3d}, DATID-3D is proposed for domain
adaptation tailored for 3D generative models. The method leverages text-to-image diffusion models to synthesize diverse images per text prompt. The dataset is refined with their
CLIP and pose reconstruction-based filtering process, and the refined dataset is used to fine-tune the 3D generator.

\subsubsection{Latent Space} 

MineGAN \citep{wang2020minegan} trains a miner network $M$ during adaptation, to map the latent space $z$ of the source GAN to another space $u = M(z)$ more appropriate for the target domain. 
MineGAN++ \citep{wang2021minegan++} extends MineGAN by only updating important parameters.
GenDA \citep{yang2021genda} proposes a lightweight attribute adaptor in the form of scaling and shifting latent codes
to adapt the latent space of the source GAN to the target GAN.
TF\textsuperscript{2} \citep{yu2024tf} aligns the latent codes of the defect-free and defect images after denoising to enable robust transfer of the defects to the defect-free images and improve the diversity of the few-shot defect image generation.

LCL \citep{mondal2023lcl} freezes $G$ and learns a network to map the latent codes from the $\mathcal{Z}$ space to the extended intermediate space $\mathcal{W}^+$ of a pre-trained StyleGAN2 during adapting GAN.
SoLAD \citep{Mondal2024SoLAD} introduces a sample-specific latent mapper to transform the sampled latent code before feeding it into the generator.
WeditGAN \citep{duan2023weditgan} proposes to learn a constant offset parameter ($\Delta w$) for the target domain in the intermediate latent space of StyleGAN2 to relocate source latent codes to the target domain.
After fine-tuning the generator to a target domain, SiSTA \citep{thopalli2023targetaware} perturbs latent representations of the fine-tuned generator that fall below a threshold, either by replacing them with zero or reverting them back to the pre-trained generator's weights.
CRDI \citep{cao2025CRDI} avoids the need for fine-tuning or retraining by proposing reconstruction and diversity enhancements to first estimate the generation path to construct the target few-shot samples, while annealing the noise perturbation scheduler for better diversity.

MultiDiffusion \citep{bar2023multidiffusion} freezes the whole parameters of the source diffusion model and optimizes the latent code as a post-processing method to generate the desired output based on a conditioned input.
DiS \citep{everaert2023dis} observed that the style of the images generated by Stable Diffusion is tied to the initial latent code. Therefore, they sample noise from the style-specific latent distribution (which is obtained by encoding the target style images to the VAE latent space) and fine-tune Stable Diffusion.

\subsubsection{Modulation}
In signal processing literature, modulation varies some key attributes of a signal to add the desired information to it \citep{oppenheim1997signals}. 
Similarly, in deep neural networks, modulation is used to add some desired information to a base network by adding modulation parameters to the parameters/ features of the base network.
AdaFMGAN \citep{zhao2020leveraging} shows that layers closer to the sample (earlier layers in $D$, and later layers in $G$) encoder general knowledge.
This general knowledge is conceptually shared between source and target domains and aimed to be preserved by Adaptive Filter Modulation which trains a scale and shift parameter for each $k \times k$ kernel.
GAN-Memory \citep{cong2020ganmemory}, CAM-GAN \citep{varshney2021camgan}, LFS-GAN \citep{seo2023lfsgan}, and CFTS-GAN \citep{ali2025cfts} use similar modulation ideas to modulate a pre-trained GAN for generative continual learning.
NICE \citep{ni2023nice} proposed to prevent overfitting by introducing noise into $D$ to modulate its hidden features. The noise is adaptively controlled by the overfitting degree of
$D$, which balance the discriminator's discrimination ability and potential overfitting issue.
AdAM \citep{zhao2022adam} and OKM \citep{zhang2024few} uses kernel modulation \citep{milad2021revisit} for few-shot generative modeling by aiming to preserve the important wights of a pre-trained GAN during a few-shot adaptation to a target domain.
HyperDomainNet \citep{alanov2022hyperdomainnet}, HyperGAN-CLIP \citep{anees2024hypergan}, DPH \citep{li2024dual}, and DoRM \citep{wu2024domain} adds an additional modulation to StyleGAN2 \citep{karras2020analyzing} for adapting to a new domain.
Similarly, Mix-of-Show \citep{gu2023mixofshow}, Orthogonal Adaptation \citep{po2024orthogonal}, DreamMatcher \citep{nam2024dreammatcher}, PortraitBooth \citep{peng2024portraitbooth}, DisenDiff \citep{zhang2024disendiff}, and RealCustom \citep{huang2024realcustom} adapt pre-trained diffusion models to learn customized concepts through modulation.
A\textsuperscript{3}FT \citep{moon2022finetuning} proposes to learn a time-aware adapter (modulation parameters) on top of frozen weights of a pre-trained diffusion model to enable different outputs in different denoising steps when adapting the diffusion model to a target domain with limited data.

\subsubsection{Adaptation-Aware}
Adaptation-aware transfer learning approaches propose that different parts of the knowledge encoded on a pre-trained generative model could be important based on the target domain. 
AdAM \citep{zhao2022adam} proposes a probing step before the main adaptation, where the importance of each kernel for adapting a source GAN to the target domain using a few samples is measured using FI.
Then, during the main adaptation, the important kernels are preserved using modulation, and the other kernels are simply fine-tuned.
OKM \citep{zhang2024few} exactly follows AdAM for importance probing. However, in the probing stage, instead of using standard discriminator loss, they use a relaxed version of the Optimal Transport \citep{villani2009optimal} idea to measure the distance of the distribution of the generated images from that of the real images.
RICK \citep{zhao2023rick} shows that incompatible knowledge from a source domain to a target domain is related to the kernels with the least importance to this adaptation, and this knowledge can not be removed by simple fine-tuning. Therefore, RICK proposes a dynamic probing schedule during adaptation where it gradually prunes the kernels with the least importance.

\subsubsection{Natural Language-Guided}
Vision-Language models like CLIP \citep{radford2021CLIP} are usually trained on large-scale image-text pairs and learn to encapsulate the generic information by combining image and text modalities.
This generic information is shown to be helpful in various downstream tasks, including zero-shot and few-shot image generation.

{\bf Leveraging offset alignment in CLIP space for zero-shot image generation.}
StyleGAN-NADA (NADA) \citep{gal2022stylegannada} is the first work that leverages CLIP's joint image and text embedding space for zero-shot image generation.
It proposes to use the embedding offset of the textual description in the CLIP space to describe the difference between source and target domains. 
Specifically, assuming a text prompt $T_s$ that describes the source domain (\eg "Photo" for a StyleGAN2 pre-trained on the FFHQ), and a given $T_t$ (\eg "Fernando Botero Painting"), CLIP's text encoder $E_T$ is used to find the update direction in the embedding space: $\Delta T = E_T(T_t) - E_T(T_s)$. 
Similarly, the direction of the update/ change for the images can be computed using generated images with source and target generators: $\Delta I = E_I(G_t(z)) - E_I(G_s(z))$, where $E_I$ denotes CLIP's image encoder.
By assuming the text offset and image offset are well-aligned in CLIP space, NADA proposes to update the generator's parameters in a way to align $\Delta I$ and $\Delta T$ leading to the directional loss guidance $\mathcal{L}_{directional} = 1 - \Delta I\cdot\Delta T/(|\Delta I||\Delta T|)$.
IPL \citep{guo2023ipl} points out that adaptation directions in NADA for diverse image samples are computed from one pair of manually designed prompts, which will cause mode collapse.
Therefore, they learns a specific prompt for each generated image, and produce different adaptation directions for each sample.
Similarly, to prevent mode collapse, SVL \citep{jeon2023svdm} uses embedding statistics (mean and variance) for producing adaptation direction instead of only the mean of embeddings in NADA. 
While these works assumed a perfect alignment between text and image offset, AIR \citep{liu2025air} discovered that the offset misalignment exists in the CLIP embedding space and it increases as the source and target domains are more distant. To mitigate the offset misalignment issue, it proposes to sample anchor points closer to the target iteratively during adaptation.

{\bf Extending directional guidance for one-shot setting.}
Directional loss proposed in NADA \citep{gal2022stylegannada} can be easily extended to one-shot image generation, by replacing $\Delta T$ with the direction obtained by target image $I_t$ and a batch of generated images by the source generator: $\Delta I' = E_I(I_t) - \mathbb{E}_i \{E_I(G_s(z_i))\}$, where $\mathbb{E}_i \{E_I(G_s(z_i))\}$ denotes the mean of the CLIP embedding for a batch of generated images.
MTG \citep{zhu2022mindthegap} extends the idea for one-shot image generation by replacing the mean embedding with the projection of the target image on the source generator denoted as $I^*_s$. Specifically, MTG uses GAN inversion to get the corresponding $z^{ref}$ for $I_t$, and uses it to generate the projected image: $I^*_s=G_s(z^{ref})$. 
HyperDomainNet improves the performance of the NADA and MTG by freezing the weights of the source generator and training modulation weights for the synthesis part inside the generator.
DiFa \citep{zhang2022difa} adds an attentive style loss to directional loss of NADA \citep {gal2022stylegannada} as a local-level adaptation which aligns the intermediate tokens of the generated image with source and pre-trained GAN.
OneCLIP \citep{kwon2022oneclip} exploits the CLIP embedding for three major modules in one-shot learning: i) inverting sample into latent space, ii) preserving the diversity of the GAN during adaptation, and iii) 
a patch-wise contrastive learning approach for preserving local consistency.

{\bf Language guidance for diffusion control.}
The connection between the image embeddings and natural language is also used for adapting generative models to new domains through techniques like diffusion guidance and prompt learning. 
StyleGAN-Fusion \citep{song2024stylegan} aligns a pre-trained StyleGAN2 \citep{karras2020analyzing} with target domains by mapping generated images into the latent diffusion space \citep{rombach2022latentdiffusion} and optimizing based on text-conditioned denoising loss. Similarly, HyperGAN-CLIP \citep{anees2024hypergan}, UniHDA \citep{li2024unihda}, and DPH \citep{li2024dual} leverage CLIP embeddings of image and text to guide domain adaptation, either via hypernetworks that modulate GAN weights or by computing direction-based latent updates. StyO \citep{li2024styostylizefaceoneshot} targets one-shot face stylization by learning disentangled tokens for source and target styles, along with image-specific tokens to preserve identity. To address fairness in generation, \textsc{ITI-Gen} \citep{zhang2023itigen} and FairQueue \citep{teo2024fairqueue} propose prompt learning from a limited number of target images to enhance the representation of under-sampled sensitive attributes in text-to-image diffusion models.

{\bf Language-enabled subject-driven generative modeling.}
A recent and highly popular research direction involves leveraging the connection between images and natural language to embed a concept directly into the embedding space of image generators (SGM in our proposed GM-DC task taxonomy in Sec.~\ref{ssec:tasks}; this is also called as personalization of generative models in some recent works).
DreamBooth \citep{ruiz2023dreambooth} addresses subject-driven sample generation by fine-tuning a text-to-image diffusion model e.g., Imagen \citep{saharia2022imagen}, or Stable Diffusion \citep{rombach2022latentdiffusion}. 
Input images are paired with a text prompt that contains a unique identifier and the subject class (e.g., ``A [V] dog''), and the pair is used to fine-tune the model. 
They further propose a class-specific prior preservation regularizer to encourage diversity and to mitigate {\em language drift}, i.e., the model progressively loses syntactic and semantic knowledge during fine-tuning.
To support continual subject learning without overwriting prior knowledge, LFS-Diffusion \citep{song2024towards}, and L\textsuperscript{2}DM \citep{sun2024create} introduce lifelong personalization frameworks based on knowledge distillation. 
Dreambooth-CL \citep{zhu2024enhancing} extends Dreambooth to learn the differences between various input concepts by leveraging a multimodal contrastive learning loss through the CLIP vision encoder.
T2IRL \citep{wei2025T2IRL} proposes the utilization of the diffusion model as a deterministic policy that can be guided by a learnable reward policy for personalized sample generation.
Contrary to DreamBooth \citep{ruiz2023dreambooth}, AblateConcept \citep{kumari2023ablating} aims to prevent the generation of specific concepts (e.g., copyrighted content) in diffusion models. It achieves this by introducing a KL-divergence loss that aligns the conditional distribution of the target concept with its core class, effectively erasing distinctive details (e.g., converging "A photo of a Grumpy cat" to "A photo of a cat").

{\bf Efficient tuning for SGM.}
Instead of fine-tuning the generator,
Textual-Inversion \citep{gal2022textualinversion} optimizes a word vector for the new subject given a few images and uses that word vector for SGM. 
DBLoRA \citep{pascual2024enhancing} uses the same framework as DreamBooth, but only updates LoRA weights instead of fine-tuning the whole generator to reduce the computation and memory requirements.
It also uses an objective-specific token and a style token to enable more controlled subject-driven generation.
HyperDreamBooth \citep{ruiz2024hyperdreambooth} proposes improved memory and time efficiency to DreamBooth \citep{ruiz2023dreambooth}. Specifically, it includes a Hyper-network for efficient approximation of the network-weight, followed by fast fine-tuning for better image fidelity. 
HybridBooth \citep{guan2025hybridbooth} combines optimization-based and regression-based methods in a two-stage process, where an initial domain-agnostic word embedding is refined based on the subject image.

{\bf Tuning-free SGM.}
A line of research focuses on learning explicit subject representations through encoder-based methods for fast and controllable SGM. BLIP-Diffusion \citep{li2023blipdiffusion}, SINE \citep{zhang2023sine}, ELITE \citep{Wei2023elite}, E4T \citep{gal2023e4t}, SSR-Encoder \citep{zhang2024ssr}, and MasterWeaver \citep{wei2025masterweaver} leverage encoders or mapping networks to extract subject features, enabling efficient generation without iterative token optimization. Expanding beyond single-modal encoders, recent works like MoMA \citep{song2025MOMA}, MultiGen \citep{wu2025MultiGen}, and AnomalyDiffusion \citep{hu2024anomalydiffusion} incorporate multi-modal inputs—such as text, image, and spatial information—to facilitate fine-grained control and broaden the scope of subject-driven generation.
In parallel, another line of work seeks to eliminate test-time fine-tuning by directly learning subject-aware modules. InstantBooth \citep{shi2023instantbooth} achieves this through three auxiliary networks that encode subject identity and inject relevant features into the UNet. 
IDAdapter \citep{cui2024idadapter} follows a similar goal using adapter layers and fused facial features, enabling efficient and identity-consistent generation without iterative optimization.

{\bf Improving localization and disentanglement for controllable SGM.}
Some of the recent methods aim to improve subject consistency and region-aware control in SGM. FastComposer \citep{xiao2024fastcomposer}, PortraitBooth \citep{peng2024portraitbooth}, DreamMatcher \citep{nam2024dreammatcher}, and DisenDiff \citep{zhang2024disendiff} focus on enhancing subject localization through embedding extraction, attention calibration, or alignment of attention maps. In contrast, CII \citep{jeong2023cii} and DETEX \citep{cai2024decoupled} address subject disentanglement from pose, background, and other semantics, ensuring consistent and controllable generation across diverse scenarios.

{\bf Balancing identity fidelity with prompt controllability.}
A key challenge in SGM work is balancing subject fidelity with prompt controllability. Recent works such as Lego \citep{motamed2025Lego}, CGR \citep{jin2025CGR}, DreamBlend \citep{ram2025dreamblend}, Cross Initialization \citep{pang2023crossinitialization}, SAG \citep{chan2024sag}, ZipLoRA \citep{shah2025ziplora}, RealCustom \citep{huang2024realcustom}, and PALP \citep{arar2024palp} focus on disentangling subject identity from other semantic attributes while improving alignment with prompts.
Building on the idea of leveraging structured semantic knowledge, SuDe \citep{qiao2024facechain} models subjects as subclasses of broader categories, capturing both public and subject-specific attributes through reconstruction and semantic constraints. Similarly, ComFusion \citep{hong2025ComFusion} enhances semantic alignment by integrating scene priors and visual-text matching to fuse personalized subject representations with scene-level descriptions.

{\bf Beyond SGM}, other works explore broader personalization tasks. SpecialistDiffusion \citep{lu2023specialistdiffusion} and Domain Gallery \citep{duan2024domaingallery} generalize diffusion models to hard-to-describe or unseen domains, while ProSpect \citep{zhang2023prospect} and LogoSticker \citep{zhu2025logosticker} introduce fine-grained control over generation attributes and identity-aware logo insertion, respectively. Multi-subject and multi-concept customization is tackled by Custom Diffusion \citep{kumari2023mcc}, along with modular LoRA-based frameworks such as OMG \citep{kong2025OMG}, TFIC \citep{li2025TFIC}, MagiCapture \citep{hyung2024magicapture}, Mix-of-Show \citep{gu2023mixofshow}, and Orthogonal Adaptation \citep{po2024orthogonal}, which collectively balance scalability with fidelity and modular control.


\subsubsection{Prompt Tuning}

VQ-VAEs (Sec.~\ref{ssec:generative_models}) can be broadly categorized into two types from the perspective of predicting the latent prior of visual tokens. AutoRegressive (AR) approaches like DALL$\cdot$E \citep{ramesh2021dalle} and VQ-GAN \citep{esser2021vqgan}, learn an AR predictor that follows a raster scan order and predicts the visual tokens from left to right, line-by-line. 
Non-AutoRegressive (NAR) approaches like DALL$\cdot$E2 \citep{ramesh2022dalle2}, MaskGIT \citep{chang2022maskgit}, Latent Diffusion \citep{rombach2022latentdiffusion}, or Imagen \citep{saharia2022imagen} usually resort to masking techniques \citep{devlin2019bert} to predict the visual tokens in a series of refinement or denoising steps.
VPT \citep{sohn2023vpt} is the first work that adopts the prompt tuning idea for image generation with generative knowledge transfer.
It uses a VQ-VAE framework where a MaskGIT\citep{chang2022maskgit}/ VQ-GAN\citep{esser2021vqgan} on the ImageNet dataset (as an example of NAR/ AR approach) is used as a pre-trained network.
Then, during adaptation, all the parameters of the VQ Encoder, VQ decoder, and transformer are frozen, and a generator is learned to minimize the adaptation loss by generating and appending a set of visual tokens to the predicted prior.
These visual tokens guide the generation process for the target domain by helping the transformer to predict proper tokens to the VQ decoder.

\subsection{Data Augmentation}
\label{ssec:review_dataaugmentation}
Data augmentation increases the quantity and diversity of the training data 
which is shown to be beneficial for GM-DC.
However, if it is not deployed correctly, augmentations can leak into the generator resulting in generating samples with similar augmentations, \eg noisy or rotated images, which is undesirable.

\subsubsection{Image-Level Augmentation}
Early work
CR-GAN \citep{Zhang2020Consistency} and bCR \citep{zhao2021improved} apply various transformations on images and enforce the output of the generator to be the same for original and transformed images.
Even though not developed specifically for GM-DC, experimental results in \citet{karras2020ada} show that CR-GAN and bCR are beneficial for limited data scenarios.
ADA proposes applying the transformations to real and fake images but with a probability $p < 1$.
The central design in ADA \citep{karras2020ada} is that the strength of the augmentation ($p$) is being adapted based on the training dynamics.
Specifically, the portion of the real images that get positive output from the discriminator, \ie, $r = \mathbb{E}[Sign(D)]$, is used as an indicator of the discriminator overfitting ($r=0$ no overfitting, and $r=1$ complete overfitting).
Then, during training, $p$ is adjusted to keep $r$ low.
DiffAugment \citep{zhao2020diffaug}, and IAG \citep{zhao2020imageaugmentation} use a similar idea to ADA, but without the adaptive component ($p=1$).
APA \citep{jiang2021deceive} uses the same adaptive augmentation mechanism in ADA, but instead of using transformations like rotation, it randomly labels generated images as pseudo-real ones to prevent an overconfident discriminator.

DiffusionGAN \citep{wang2023diffusiongan} applies the gradual diffusion process on real and generated images during training GAN. 
DANI \citep{zhang2024dani}, similar to DiffusionGAN \citep{wang2023diffusiongan}, prevents $D$ from overfitting by injecting noise into images (exactly like the Diffusion process) to augment both real and fake samples. Similar to NICE \citep{ni2023nice}, the noise is adaptive and controlled by the overfitting degree of $D$.
Training starts with real and generated images, and each diffusion step is applied after a certain number of training epochs, making the bi-classification task harder for the discriminator to prevent its overfitting.

PatchDiffusion \citep{wang2023patchdiffusion} 
augments the data during training diffusion models by sampling patches with random locations and random sizes alongside the full image and conditioning the denoising score function \citep{karras2022elucidating} on the patch size and the location information. 
To address the problem of 
generating out-of-distribution samples,
ANDA \citep{zhang2024anda} applies negative data augmentation \citep{sinha2021negative} to real data to create out-of-distribution samples as fake samples for the discriminator $D$.
AugSelf-GAN \citep{hou2024augselfgan} applies 
self-supervised learning to learn augmentation-aware information.

\subsubsection{Feature-Level Augmentation}
AdvAug \citep{chen2021advaug} computes the adversarial perturbation $\delta$ for the feature maps of the discriminator and generator using the projected gradient descent \citep{madry2018towards}. Denoting the discriminator as $D = D_2 \circ D_1$, the adversarial augmentation is applied on the intermediate feature maps ($D_1$) of both real and generated images, resulting in $D_{1}(x) + {\delta}$, and $D_{1}(G(z)) + {\delta}$. The adversarial loss is then added to the loss function of $D$ during GAN training to maximize the score of the perturbed real image and minimize the score of the perturbed generated image:
\begin{equation}
    \mathcal{L}_{D}^{adv} \coloneqq \max_{\| \delta \|_{\infty} < \epsilon} \mathbb{E}_{x \sim p_{data}} [f_{D}(-D_2(D_1(x)+\delta))] 
    +
    \max_{\| \hat{\delta} \|_{\infty} < \epsilon} \mathbb{E}_{z \sim p_{z}} [f_{D}(D_2(D_1(G(z))+\hat{\delta}))] 
\end{equation}
As AdvAug is performed on the feature level, it is shown to be complementary to image-level augmentations like ADA \citep{karras2020ada} and DiffAug \citep{zhao2020diffaug}.
AFI \citep{dai2021implicit} observes a flattening effect in discriminators with multiple output neurons, and takes advantage of this observation by proposing feature interpolation as implicit data augmentation.
Meanwhile,
inspired by the improved performance of classifiers by debasing them regarding texture, FSMR \citep{kim2022feature} aims to improve GM-DC by applying a similar idea to GAN's discriminator.
It augments the style of each image in the feature space of discriminator and enforces the prediction for these augmented samples to be similar to original sample.

\subsubsection{Transformation-Driven Design}
DAG \citep{tran2021dag} uses a separate discriminator $D_k$ for discriminating real and fake images that are augmented by transformation $T_k$. 
A weight-sharing mechanism between all discriminators is used to prevent overfitting.
Additionally, DAG provides a theoretical ground for training convergence under augmentation. As mentioned in \citet{goodfellow2014GANs}, for an optimal discriminator $D^*$, optimizing $G$ is equivalent to minimizing the Jensen-Shannon (JS) divergence between the real data distribution $P_{data}$ and generated data distribution $P_{model}$, \ie, $JS(P_{data}, P_{model})$. 
Denoting $P_{data}^{T}$, and $P_{model}^{T}$ as the distribution of the real and generated data under augmentation $T$, \citet{tran2021dag} shows that JS divergence between two distributions is invariant under differentiable and invertible transformations:
\begin{equation}
    \label{eq:JS_invariant}
    JS(P_{data}, P_{g}) = JS(P_{data}^{T}, P_{g}^{T})
\end{equation}
This means that as long as the augmentation is differentiable and invertible, training convergence is guaranteed.
SSGAN-LA \citep{hou2021labelaugmentation} extends DAG by merging all discriminators to a single discriminator and augmenting the label space of the discriminator, \ie, asking $D$ to detect the type of augmentation in addition to conventional real/ fake detection.

\subsection{Network Architectures}
\label{ssec:review_networkarchitecture}

\subsubsection{Feature Enhancement}
FastGAN \citep{liu2021fastgan} proposes a light-weight GAN structure ---shallower $G$ and $D$ compared to SOTA GANs like StyleGAN2--- to decrease the risk of overfitting.
Inspired by skip connections \citep{he2016resnet}, and squeeze-and-excitation module \citep{hu2018squeeze}, FastGAN fuses features with different resolutions in $G$ through proposed skip-layer excitation modules.
An additional reconstruction task is defined for $D$.
cF-GAN \citep{hiruta2022cfgan} is a typical CGAN advanced on FastGAN to perform cGM-1.
MoCA \citep{li2022moca} learns some prototypes for each semantic concept within a domain, \eg, railroad, or sky in a photo of a train.
Then, by attending to these prototypes during image generation, some residual feature maps are produced to improve image generation.
DFSGAN \citep{yang2023dfsgan} proposes to preserve the content and layout information in intermediate layers of $G$ by extracting channel-wise and pixel-wise information and using them to scale corresponding feature maps.
DM-GAN \citep{yan2024dm} proposes an encoder-decoder-based design for the generator, which consists of CNNs and ViTs for efficiently capturing both global and local information and enhancing image generation under limited data.
FewConv \citep{liu2025fewconv} proposes to replace traditional convolutions in GANs with a new design to independently learn the spatial and channel information and therefore reduce the amount of information that needs to be learned for generative modeling.
SCHA-VAE \citep{pmlr-v162-giannone22a} extend latent variable models for sets to a fully hierarchical approach and propose Set-Context-Hierarchical-Aggregation VAE for few-shot generation.

\subsubsection{Ensemble Pre-trained Vision Models}
ProjectedGAN \citep{sauer2021projectedgan} proposes to project real and generated images into the feature space of a pre-trained vision model to enhance $D$'s performance in discriminating real and fake images by adding two modules.
First, the output from multiple layers is used with separate discriminators.
Then a {\em random projection} is used to dilute the features and encourage the discriminator to focus on a subset of the features.
SPGAN \citep{hiruta2022cfgan}, built on ProjectedGAN, improves the generation quality of uGM-1 by simply introducing style mapping of StyleGAN and Skip Layer Excitation of FastGAN.
Vision-aided GAN \citep{kumari2022ensembling} uses an ensemble of the original discriminator $D$ and additional discriminators $\{ \hat{D}_n \}_{n=1}^{N}$ to perform the classification task.
The additional discriminators $\{ \hat{D}_n \}_{n=1}^{N}$ have a set of pre-trained feature extractors $\mathcal{F} = \{ F_n \}_{n=1}^{N}$ (extracted form pre-trained vision models) with a small trainable head $C_n$ added on top: $\hat{D}_n =  F_n \circ C_n$. 
P2D \citep{chong2024p2d}, similar to Vision-aided GAN \citep{kumari2022ensembling}, ensembles multiple pre-trained vision models to improve $D$. It proposed to include an R1 regularizer to prevent the classification heads from overfitting.
DISP \citep{mangla2022data} also follows a similar idea to vision-aided GAN to leverage a pre-trained classifier $C$, but it conditions generation by $G$ on the extracted features of a real image with this classifier ($C(x)$), and enforces $G(z|C(x))$ to be similar to input image $x$ in discriminator's feature space.

\subsubsection{Dynamic Network Architecture}
CbC \citep{shahbazi2022collapse} shows that under data constraints, 
where an unconditional GAN can generate satisfactory performance, training the 
conditional GANs (cGANs) result in mode collapse. 
To mitigate this issue, CbC \citep{shahbazi2022collapse} starts training from an unconditional GAN and slowly transitions to a cGAN using a transition function $0 \leq \lambda_t \leq 1$.
Considering the conditioning variable as $c$, this transition is implemented in $G$ as $G(z,c,\lambda_t)=G(S(z)+\lambda_t \cdot E(c))$, with $S$ and $E$ as neural networks that transform the latent code and the conditioning variable.
PYP \citep{li2024pyp}, using a similar architecture as CbC \citep{shahbazi2022collapse}, addresses few-shot image generation by generalizing from large pillar datasets during training. Different from CbC, the class embedding is injected into $G$ in parallel to the style code $w$. It further improves the generation diversity by using directional loss.

DynamicD \citep{yang2022dynamicd} dynamically reduces the capacity of $D$ by randomly sampling a subset of channels of $D$ during each training iteration to prevent overfitting.
Inspired by the lottery ticket hypothesis \citep{frankle2019lotteryticket}, AdvAug \citep{chen2021advaug} and Re-GAN \citep{saxena2023regan} have shown that a much sparse subnetwork of the original generator can be useful for GM-DC.
RG-GAN \citep{saxena2024rggan} proposed a new weight pruning method. Beyond standard network pruning, to prevent model from over-pruning, a regeneration step is implemented to reintroduce some weights if they gain importance during training.
AutoInfoGAN \citep{shi2023autoinfogan} applies a reinforcement learning-based neural architecture search to find the best network architecture for the generator.


\subsection{
Multi-Task Objectives
}
\label{ssec:review_trainingtechniques}

\subsubsection{Regularizer} 
LeCam \citep{tseng2021lecam} uses two moving average values to track $D$'s prediction for real and generated images, denoted by $\alpha_R$ and $\alpha_F$, respectively.
Then the distance between the $D$'s prediction for real (fake) images and $\alpha_F$ ($\alpha_R$) is decreased by adding a regularizer to prevent overfitting. 
Analysis in \citet{tseng2021lecam} shows that this regularizer enforces WGAN \citep{arjovsky2017wasserstein}/ BigGAN\citep{brock2019biggan} to minimize the LeCam-divergence which is beneficial for GM-DC.
Reg-LA \citep{hou2023regularizing} uses a similar idea to regularize the label-augmented GANs discussed in Sec. \ref{ssec:review_dataaugmentation}.
DigGAN \citep{fang2022diggan} shows that the discriminator gradient gap between real and generated images increases when training GANs with limited data, and adds this gap as a regularizer to prevent this behavior.
MICGAN \citep{zhai2024micgan} observed the mutual information (MI) decay issue in high-resolution uGM-1, therefore, it proposed to explicitly optimize the MI between the features of each layer.
CHAin \citep{ni2024chain} revisits batch normalization (BN) for training GANs under limited data, and suggests replacing the conventional centering of BN with zero-mean regularization and leveraging Lipschitz continuity constraint \citep{gouk2021regularisation} for the scaling part of the BN.
MDL \citep{kong2022mdl} addresses the pre-training free few-shot image generation by adding a regularizer that aims to keep the similarities between the latent codes in $\mathcal{Z}$ space and corresponding generated images in image space.
DFMGAN \citep{duan2023few} proposes the first defect generation approach with limited data. In the first training stage, StyleGAN2-ADA \citep{karras2020ada} is trained on defect-free images. Then, in the second stage, defect-aware layers are added on top of it to generate the defect masks and fuse the defect features to the main backbone.
A variant of mode-seeking loss \citep{mao2019mode} is proposed as a regularizer to encourage the generation of different defects for similar objects.

\subsubsection{Contrastive Learning}
InsGen \citep{yang2021insgen} uses contrastive learning to improve learning $D$ by introducing a pretext task.
The pretext task is defined as instance discrimination, meaning that each sample should be mapped to a separate class. This is done by constructing the query and key from the same sample as positive pair, and all remaining images as negative pair. 
FakeCLR \citep{li2022fakeclr}  
analyze three different contrastive learning strategies, namely instance-real,  instance-fake, and instance-perturbation. It is shown that instance-perturbation contributes the most improvement in quality and can effectively alleviate the issue of latent space discontinuity.
RCL \citep{gou2024few} follows a similar idea to DCL \citep{zhao2022dcl} but for training a generator from scratch.

{\bf Remark.}
As discussed in Sec.~\ref{ssec:Regularizer-based Fine-Tuning}, constrastive learning is also used in works like C\textsuperscript{3} \citep{Lee2021C3}, DCL \citep{zhao2022dcl}, CtlGAN \citep{wang2022ctlgan}, CML-GAN \citep{phaphuangwittayakul2022cmlgan}, and IAG \citep{zhao2020imageaugmentation} as a regularizer during adapting a pre-trained source generator to the target domain.
However, approaches discussed in this section use contrastive learning to train a generative model from scratch using limited data.

\subsubsection{Masking}
MaskedGAN \citep{huang2022maskedgan} utilizes a masking idea for training GANs under limited data by masking both spatial and spectral information.
For spatial masking, they use a patch-based mask to enable random masking of all spatial parts. For spectral masking, they mask each frequency channel (extracted by the Fourier transform) based on the amount of information, \ie, channels with more information are more probable to be masked.
MaskD \citep{zhu2022few} randomly masks feature maps extracted by $D$ for a few-shot setup.
DMD \citep{zhang2023dmd} detects that the discriminator slows down learning and applies random masking to its features adaptively to balance its learning pace with the generator.

\subsubsection{Knowledge Distillation}
KD-DLGAN \citep{cui2023kddlgan} proposes a knowledge distillation (KD) \citep{hinton-distill,chandrasegaran2022revisiting} approach by leveraging CLIP \citep{radford2021CLIP} as the teacher model to distill text-image knowledge  to the discriminator.
They propose two designs: aggregated generative knowledge designs a harder learning task, and correlated generative knowledge distillation improves the generation diversity by distilling and preserving the diverse
image-text correlation from CLIP.
BK-SDM \citep{kim2025bksdm} prunes several residual and attention blocks (manually defined) from the U-Net of Stable Diffusion \cite{rombach2022latentdiffusion} and use feature distillation to compensate for the decline in performance.
As discussed before, KDFSIG \citep{hou2022exploitingkd} also uses KD in the context of transfer learning for few-shot image generation.

\subsubsection{Prototype Learning}
Inspired by the success of learning prototypes in few-shot classification, ProtoGAN \citep{yang2023protogan}, aims to improve the fidelity and diversity of the FastGAN under limited data \citep{snell2017prototypical}.
ProtoGAN has two main modules: prototype alignment for increasing the fidelity of the generated images, and diversity loss to improve the generation diversity.
MoCA \citep{li2022moca} also learns prototypes but for different semantic concepts through an attend and replace mechanism on the extracted feature maps of $G$.

\subsubsection{Other Multi-Task Objectives}
Gen-Co \citep{cui2022genco} uses multiple discriminators to extract diverse and complementary information from samples.
This {\em co-training} has two major modules: weight-discrepancy co-training, which trains separate 
$D$s with different weights, and data-discrepancy co-training 
which in addition to training separate $D$s also uses different information as inputs, \ie, spatial or frequency information.
AdaptiveIMLE \citep{aghabozorgi2023adaptiveimle} proposes an adaptive version of implicit maximum likelihood estimation \citep{li2018imle} to improve the mode coverage by assigning different boundary radii for each sample.
RS-IMLE \citep{vashist2024rejection} also leverages implicit maximum likelihood estimation but for choosing a different prior so that the samples selected for training have a distribution more similar to those sampled at inference.
PathcDiffusion \citep{wang2023patchdiffusion} and AnyResGAN \citep{chai2022anyresolution} show the effectiveness of {\em Patch-Level} learning of the generators.
Diffusion-GAN \citep{wang2023diffusiongan} leverages the {\em diffusion process} to improve training GANs by gradually increasing the task hardness for $D$.  D2C \citep{sinha2021d2c} uses a DM to improve the sampling process of VAEs by denoising the latent codes and feeding VAE with a clean latent code for sample generation.
FSDM \citep{giannone2022fsdm} uses an attentive conditioning mechanism and aggregates image patch information using a vision transformer for image generation for unseen classes.
SpiderGAN \citep{asokan2023spider} uses an image dataset as input for training a GAN under limited data instead of using latent codes. It argues that choosing a low-entropy dataset (instead of high-entropy latent codes from a Normal distribution) helps with faster and better training procedures. 
They further propose a signed Inception Distance metric to select a closer subset of data to the target domain.

\subsection{Exploiting Frequency Components}
\label{ssec:review_exploitingfrequency}
Approaches in this category aim to improve frequency awareness to improve GM-DC.
FreGAN \citep{yang2022fregan} extracts high-frequency information 
($HF$)
of images (related to details in images) using Haar Wavelet transform \citep{porwik2004haar} and uses three different modules to emphasize learning high-frequency information: high-frequency discriminator uses $HF$ as an additional signal to perform real/fake classification, frequency skip connection feeds the $HF$ information of each feature map to the next one in $G$ to prevent frequency loss, and a frequency alignment loss is used to make sure $G$ and $D$ are learning frequency information in the same pace.
WaveGAN \citep{yang2022wavegan} uses a similar idea, but in a different setup to address the cGM-2 task.
SDTM \citep{yang2023sdtm} applies Haar Wavelet transformation to decompose features of $D$, encouraging the model to distinguish high-frequency signals of real images from those of generated samples, therefore mitigating the model’s frequency bias.
Gen-Co \citep{cui2022genco} extracts some frequency information of the image and feeds it to a separate $D$ in addition to using original real and fake images.
MaskedGAN \citep{huang2022maskedgan} masks out some frequency bands of the input during training to enforce the generative model to focus more on under-represented frequency bands.
FAGAN \citep{cheng2024frequency} introduces two frequency regularizers for one-shot adaptation. It aligns low-frequency components of $G_t$ with $G_s$ to preserve general knowledge and matches high-frequency components of the generated image with the reference image to capture domain-specific details.

\subsection{Meta-Learning}
\label{ssec:review_metalearning}
Meta-learning shifts the learning paradigm from data level to task level to capture across-task knowledge as {\em meta-knowledge}, and then adapt this meta-knowledge to improve the learning process of unseen tasks in the future.
A plethora of recent works adopt meta-learning to tackle few-shot classification \citep{finn2017maml, snell2017prototypical, vinyals2016matching, sung2018relationnet, milad2021revisit} and few-shot semantic segmentation \citep{wang2019panet}.
These works usually follow the {\em episodic learning} setup which matches the way that the model is trained and tested.
Considering task distribution $P_{\mathcal{T}}$, a set of training tasks are constructed from seen classes $\mathcal{T}^{train} = \{ \mathcal{T}^{train}_i \}$, where $\mathcal{T}^{train}_i$ denotes $i^{th}$ training (meta-training) task. 
The model is trained on the meta-training tasks and later tested on the meta-test tasks $\mathcal{T}^{test} = \{ \mathcal{T}^{test}_j \}$ constructed from unseen classes. Usually, meta-training and meta-testing tasks follow the same distribution $P_{\mathcal{T}}$.
Similarly, the approaches in this category use meta-learning to address image generation: train a generative model on a set of few-shot image generation tasks constructed from seen classes of a domain, then test it on the few-shot image generation tasks from unseen classes of the same domain.

\subsubsection{Optimization} 
Optimization-based meta-learning algorithms are used in these approaches for learning meta-knowledge. 
Generative Matching Network (GMN) proposes a similar attention mechanism used in Matching Networks \citep{vinyals2016matching} for few-shot image generation with variational inference. 
FIGR \citep{clouatre2019figr} meta-trains a GAN using Reptile \citep{nichol2018reptile}.
Training has an inner loop that adapts the GAN weights based on a few-shot image generation task and an outer loop that updates the meta-knowledge using Reptile.
Dawson \citep{liang2020dawson} modifies the inner loop training to directly get the gradients for the generator from evaluation data.
FAML \citep{phaphuangwittayakul2021faml} uses a similar idea to FIGR, but instead of using the standard GAN structure, it uses an encoder-decoder architecture for the generator.
CML-GAN \citep{phaphuangwittayakul2022cmlgan} extends FAML \citep{phaphuangwittayakul2021faml} by leveraging contrastive learning to learn quality representations.

\subsubsection{Fusion} 
MatchingGAN \citep{hong2020matchinggan} learns to generate new images for a category by fusing the available images of that category.
A set of encoders is used to estimate the similarity between the embedding of the latent code and input images. Then, these similarities are used as interpolation coefficients by an auto-encoder to extract the embeddings of the training images and fuse them for generating new images.
F2GAN \citep{hong2020f2gan} uses random coefficients for general information, and an attention module for details. 
The attention module takes the weighted average of the real image features and the corresponding features from the decoder to produce the image details.
LoFGAN \citep{gu2021lofgan} focuses on local features in the fusion process. 
Given a batch of images, one sample is selected as a base while the remaining are utilized as a reference set. This set acts as a feature bank for the fusing process.
F2DGAN \citep{zhou2024f2dgan} proposed to match the histogram of feature value instead of matching feature value directly in LoFGAN. It encode and reconstruct the features of real samples with a VAE to improve the diversity of fused features.
WaveGAN \citep{yang2022wavegan} adds frequency awareness to LofGAN by extracting and feeding the frequency components of feature maps to later layers of the generator.
SMR-CSL \citep{xiao2025semantic} extends LoFGAN by applying a mask to the fused feature during reconstruction. It further introduces a triplet loss to ensure generated images resemble real ones within the same category. This emphasizes category-specific features while enhancing inter-class distinction.
AMMGAN \citep{li2023ammgan} utilizes an adaptive fusion mechanism for learning pixel-wise metric coefficients during the fusion.
EqGAN \citep{zhou2023eqgan} balances structural and textural information by fusing multi-scale encoder features through a feature equalization module to improve the generation quality.
SDTM \citep{yang2023sdtm} proposed to improve the generation diversity by introducing out-of-distribution semantic information to the fused features.
MVSA-GAN \citep{chen2023iot} proposes a self-attention module alongside a multi-view feature fusion module to capture contextual information within the image and fuse it at the global and local levels for modeling complex scenes.
SAGAN \citep{aldhubri2024sagan} introduced an attention-based fusion to facilitate optimal integration of features from different encoder-decoder pairs.

\subsubsection{Transformation}
DAGAN \citep{antoniou2017dagan} leverages the task of the learning augmentation manifold task in the GAN learning process.
This is modeled as some transformations on the input, and these transformations are applied to the new sample from the unseen classes for sample generation. 
ISSA \citep{huang2021few} leverages the idea of implicit autoencoders to learn the transformation across datasets using an unsupervised representation in an adversarial manner, while each dataset distribution is learned using implicit distributions.
DeltaGAN \citep{hong2022deltagan} learns the difference between images (delta) in the feature space, and then uses this delta concept for diverse sample generation. 
MFH \citep{xie2022learning} aims to learn category-independent and category-related features during episodic training within different categories. 
Then, the generation network combines these two features to generate diverse images from a single input image.
Disco \citep{hong2022disco} learns a dictionary based on seen images to encode input images into visual tokens.
These tokens are then fed into the decoder with the style embedding of seen images to generate images from unseen classes.
AGE \citep{ding2022age} uses GAN inversion to invert the samples of a category to $\mathcal{W}^+$ space of StyleGAN2 \citep{karras2020analyzing}. The mean latent code for all samples of a category is used as a prototype, and all differences are considered as general attributes.
These attributes are then used to diversify sample generation for unseen classes.
TAGE \citep{zhang2024tage} extends AGE \citep{ding2022age} by learning a codebook to store category-agnostic features and using it to augment real samples through editing.
SAGE \citep{ding2023sage} addresses the class inconsistency in AGE by taking all given samples from unseen classes into account during inference.
HAE \citep{li2022hae} uses a similar idea to AGE \citep{ding2022age}, but hyperbolic space instead of using Euclidean distance, which allows more semantic diversity control.
LSO \citep{zheng2023lso} finds a prototype for each class similar to AGE \citep{ding2022age}. Then it adjusts the GAN to produce similar images to target samples using latent samples from the neighborhood of the prototype, followed by updating the prototype in latent space using the adapted GAN.
CDM \citep{gupta2024conditional} models the distribution of unseen classes in the latent space of Stable Diffusion \citep{rombach2022latentdiffusion} by leveraging the latent code of the most similar seen classes, it is then used as a conditional input to Stable Diffusion.

\subsection{Modeling Internal Patch Distribution}
\label{ssec:review_internalpatch}

\subsubsection{Progressive Training}
SinGAN \citep{shaham2019singan} is the pioneering work that makes use of the internal distribution of the patches within an image to train a generative model.
It trains a pyramid of generators $\{G_0, \dots, G_N \}$ against a pyramid of real images $\{x_0, \dots, x_N \}$, where $x_n$ is a downsampled version of input image $x$ by a factor of $r^n$.
The generator at scale $n$ uses random noise $z_n$ and an upsampled version of the generated image from the lower resolution $\tilde{x}_{n+1}$ as input:
$\tilde{x}_{n} = G_n(z_n, (\tilde{x}_{n+1})\uparrow^{r_{n}})$.
Similarly, a pyramid of discriminators is used where $D_n$ compares the $\tilde{x}_{n}$ and $x_{n}$ in patch-level for real-fake classification.
CCASinGAN \citep{wang2022ccasingan} improves SinGAN by introducing a network block to aggregate global image features, therefore avoiding the training being affected by the outliers in a single image.
ConSinGAN \citep{hinz2021consingan} stacks the new layers for a bigger scale on top of the previous layers used for a smaller scale instead of using separate generators for each scale. 
SD-SGAN \citep{yildiz2024sdsgan} enhances SinGAN by adding self-attention for global semantic control and DenseNet blocks to reduce computation while maximizing information flow.
SA-SinGAN \citep{chen2021sa} and TcGAN \citep{jiang2023tcgan} use a self-attention mechanism to enable modeling long-range correlations and local information for modeling internal patch distribution.
RecurrentSinGAN \citep{he2021recurrent} observes kernel similarities across SinGAN’s multi-scale generators and replaces them with a single recurrent generator to share parameters across scales.
BlendGAN \citep{kligvasser2022blendgan} and DEff-GAN \citep{kumar2023deffGAN} extend previous approaches for learning the internal distribution for $k$ images, thereby allowing for the potential mixing of different image semantics and improving diversity.
ExSinGAN \citep{zhang2021exsingan} compose three modular GANs to learn the structure, semantics, and texture of the internal paths within a single image in a successive manner.

SinDDM \citep{kulikov2023sinddm} applies the same idea of SinGAN but uses diffusion models with a fully convolutional lightweight denoiser.
PromptSDM \citep{park2024promptsdm} enhances SinDDM by incorporating text cross-attention into the diffusion model, where text inputs are generated by captioning blurred images. 
SinDiffusion \citep{wang2022sindiffusion} addresses artifacts in SinGAN due to progressive resolution growth by applying progressive denoising using a diffusion model architecture.
LatentSDM \citep{han2024latentsdm} accelerates inference by modeling internal patch distribution with a Stable Diffusion \citep{rombach2022latentdiffusion}, where the latent space is obtained by the fused embedding of a VAE and VQ-VAE trained by the given single image.

\subsubsection{Non-Progressive Training}
One-Shot GAN \citep{sushko2021oneshotgan} uses a standard generator (single-scale), but multiple paths for the discriminator to enforce learning objects' appearance and how to combine them.
Within the discriminator the low-level loss is defined on low-level features, and two different losses are defined to learn the content and the layout in image patches.
PetsGAN \citep{zhang2022petsgan} utilizes the semantic variation in GAN latent space through GAN inversion to enable large variations in the layout generation.
SinFusion \citep{nikankin2022sinfusion} explores learning the internal patch distribution from both single images and videos. 
SinFusion extends on DPPM \citep{ho2020denoising} and reduces the size of the receptive fields by first removing attention layers, then adopting ConvNext \citep{liu2022convnet} blocks in the U-Net \citep{ronneberger2015unet} architecture. 
To reconstruct videos, a series of images is fed into a series of 3 identical models. The first model predicts the next frame; the second model denoises and removes small artifacts from the generated images; the last model interpolates between the different frames.

\section{Critical Analysis and Empirical Comparison}
\label{Critical_Analysis_and_Empirical_Comparison}

\subsection{Critical Analysis and Design Principles}
In this section, we summarize, for each class of approaches, the key factors contributing to their success or failure, their fundamental limitations, and the deeper design principles that emerge from these observations.

\subsubsection{Transfer Learning}

{\em Regularizer-based fine-tuning} remains one of the most widely used strategies for leveraging transfer learning in GM-DC. These approaches tend to succeed when the source and target domains are closely related, because the regularizer constrains optimization to preserve useful semantic and structural priors for stabilizing adaptation. However, they often fail when the domain gap becomes large: the same constraint that stabilizes learning in similar domains can over-restrict adaptation in dissimilar ones, causing the generator to retain source-specific biases that are incompatible with the target distribution and hindering the learning of new features. This exposes a fundamental limitation: regularization strength inherently trades off between knowledge preservation and domain adaptability. Determining the optimal level of regularization requires estimating the transfer distance between domains, which is particularly difficult under limited data. A deeper design principle arising from this insight is that successful GM-DC methods should employ adaptive or data-aware regularization schemes that dynamically balance the reuse of transferable priors with the flexibility needed to capture novel characteristics of the target domain.

{\em Latent-space} methods provide a lightweight and parameter-efficient strategy for GM-DC. These approaches succeed when adaptation occurs within a compact and well-structured latent representation, allowing the model to reuse the pretrained backbone while updating only a small set of parameters. However, they often fail when the semantic gap between source and target domains becomes large. In such cases, the latent space may lack sufficient expressiveness to represent new concepts, leading to misalignment or entanglement of factors that distort the generated outputs. This highlights a fundamental limitation: adaptation performance is highly dependent on the quality of the pretrained latent representation. A deeper design principle emerging from this insight is that effective latent-space adaptation should include mechanisms for latent disentanglement and representation expansion, maintaining efficiency while enabling the model to flexibly capture novel characteristics of the target domain.

{\em Modulation-based} approaches have emerged as a practical middle ground for GM-DC. They succeed because they inject new information through lightweight modulation layers or parameters, effectively “writing” new concepts without overwriting pretrained base knowledge. However, they often fail when the required transformation between source and target domains is large or complex, as the limited modulation capacity may underfit distant domain adaptations. Performance also depends strongly on the granularity and location of modulation: inappropriate layer selection can lead to insufficient adaptation or degradation of important pretrained representations. This highlights a fundamental limitation: the method’s success is tied to the hierarchical representation of the specific model and is sensitive to how modulation interacts with that hierarchy. The deeper design principle is that effective modulation-based adaptation must align modulation granularity and placement with the model’s semantic representation hierarchy, ensuring sufficient modulation capacity to inject new concepts effectively while preserving pretrained knowledge.

{\em Adaptation-aware} methods provide a systematic and data-driven mechanism for selecting and preserving transferable knowledge in GM-DC. They succeed because they explicitly measure the importance or compatibility of model components, such as convolutional kernels or attention heads, often using criteria like Fisher Information. By freezing or pruning components of different importance, these methods retain useful pretrained knowledge while reducing negative transfer from irrelevant or harmful features. However, they often fail when the importance estimation are unreliable—especially under limited data—leading to the removal of components that are crucial for representing target-domain knowledge. This highlights a fundamental limitation: the quality of adaptation depends on the accuracy and stability of importance estimation under data scarcity. The deeper design principle is that effective adaptation-aware strategies should combine robust importance estimation with dynamic update mechanisms, allowing the model to preserve and refine relevant components during adaptation.

{\em Natural language-guided} approaches have shown strong performance in zero-shot generative modeling and personalization, especially in diffusion-based models where textual prompts enable flexible concept transfer with minimal training data. They succeed because they leverage the rich alignment between text and visual representations learned by large multimodal foundation models such as CLIP. However, these methods often fail when the textual encoder or base model has inadequate instruction-following ability or weak language grounding, leading to unstable or inaccurate generations. They can also amplify or transfer undesired semantic biases inherited from the pretrained text–image space. This highlights a fundamental limitation: performance depends heavily on the alignment quality and bias characteristics of the underlying multimodal model. The deeper design principle is that effective language-guided transfer requires robust semantic alignment and a reliable, bias-aware base model to ensure that linguistic guidance remains precise, consistent, and faithful to the intended target concept.

{\em Visual prompt-tuning} provides a highly parameter-efficient strategy for GM-DC and effectively mitigates catastrophic forgetting by keeping the pretrained backbone frozen. These methods succeed when the base model already encodes concepts and structures that overlap with the target domain, allowing lightweight visual tokens or prompts to guide generation without extensive retraining. However, they often fail under large domain shifts or fine-grained structural variations, where the limited capacity of guidance tokens cannot adequately steer the generation process toward new or complex target features. This highlights a fundamental limitation: visual prompts operate within the representational bounds of the pretrained model and have limited ability to extend its conceptual space. The deeper design principle is that effective visual prompt-tuning should incorporate mechanisms that enhance steering capacity while preserving the efficiency and stability benefits of frozen-backbone adaptation.

\subsubsection{Data Augmentation}

{\em Image-level augmentation} methods improve GM-DC by expanding the coverage of the data distribution through transformations applied directly to the image space. They succeed because augmentations such as flips, color jittering, and affine changes prevent the model from memorizing the limited training samples and enhance the data efficiency of generative models. However, they can fail when the applied transformations alter the semantic meaning of images. This mismatch leads to undesired invariance that weakens feature representation and ultimately affects the generator’s ability to model realistic variations. The fundamental limitation is their lack of semantic awareness: augmentations operate purely on pixel-level changes without considering their effect on class or concept meaning. The deeper design principle is that effective augmentation should preserve semantic consistency, using adaptive or learned strategies that respect the underlying data manifold.

{\em Feature-Level Augmentation} enhance GM-DC by applying transformations in the feature space rather than directly on images. They succeed because the feature manifold, being smoother and more compact than the raw image space, allows meaningful interpolation and mixing that improves regularization and generalization. By combining or interpolating features, these methods encourage invariance to stylistic variations and smooth the training landscape, helping the model learn more robust representations from limited data. However, their effectiveness depends strongly on the quality of the embedding space: if the features are poorly structured or lack semantic alignment, the augmented samples may lose meaning or introduce artifacts. The fundamental limitation is their reliance on the semantic quality and geometry of the learned feature space. The deeper design principle is that successful feature-level augmentation requires well-organized, semantically meaningful embeddings where interpolation preserves semantic consistency.

{\em Transformation-driven design} approaches leverage knowledge of each individual transformation to construct augmentation strategies that enhance learning without corrupting the true data distribution. They succeed because, unlike classical augmentation that can mislead the generator into modeling transformed data (e.g., rotated or flipped versions) instead of the original domain, these methods explicitly account for how each transformation affects the data and adjust the training objective accordingly. This enables the generator to benefit from augmented diversity while still learning the correct underlying distribution. However, their success depends heavily on the careful design and calibration of transformation types. They also introduce multiple hyperparameters that require extensive tuning for stability. The fundamental limitation lies in their design complexity and sensitivity to transformation choice. The deeper design principle is that effective transformation-driven augmentation should be guided by a clear understanding of how each transformation interacts with the generative process, balancing diversity enrichment with faithful distribution learning.

\subsubsection{Network Architecture}

{\em Feature enhancement} approaches aim to improve GM-DC by designing dedicated modules that strengthen or preserve specific features often overlooked by standard training objectives. They succeed because emphasizing such features, such as texture, edge, or structural cues, helps the model capture fine-grained details that would otherwise be lost when data diversity is low. This selective enhancement improves visual fidelity and stability in generation. However, these methods often fail to generalize across diverse domains because the enhanced features are typically task or dataset specific; what benefits one type of data, such as faces, may hinder another, such as natural scenes. This highlights a fundamental limitation: the effectiveness of feature enhancement depends heavily on correctly identifying which features to emphasize and where to apply them within the network, which is difficult to determine consistently across domains. The deeper design principle is that successful feature enhancement should balance specificity and generality.

{\em Ensemble pre-trained vision model} approaches enhance GMDC by leveraging multiple pre-trained vision networks to strengthen the discriminator with rich, complementary semantic representations. They succeed because these pre-trained models, each specialized in learning specific visual or semantic cues, provide a robust supervisory signal that stabilizes adversarial training and improves the generator’s ability to capture fine-grained structures. However, these methods often face practical and conceptual challenges when applied in data-constrained settings. Each pre-trained network requires its own set of trainable layers to align with the generative task, which significantly increases computational and data demands. Moreover, there is no principled way to evaluate or rank pre-trained vision models based on their suitability for a given target domain, leading to potential inefficiencies or suboptimal combinations. This reveals a fundamental limitation: ensemble-based enhancement depends on the compatibility and effective integration of heterogeneous pretrained features within the generative framework. The deeper design principle is that successful ensemble integration requires adaptive selection and alignment mechanisms that automatically identify and fuse the most relevant pretrained representations.

{\em Dynamic network architecture} approaches adapt the model structure during training to better capture the underlying data distribution. They succeed because dynamically evolving architectures can balance model capacity and data complexity, helping prevent both overfitting and underfitting while promoting efficient knowledge transfer. However, they sometime suffer from instability, as adding or removing layers can disrupt previously learned representations. The fundamental limitation lies in the trade-off between adaptive flexibility and training stability. The deeper design principle is that effective dynamic architectures should incorporate mechanisms that preserve learned knowledge during structural changes and regulate capacity growth in a stable manner.

\subsubsection{Multi-Task Objective}

{\em Regularizer} approaches address undesirable behaviors that arises when generative models are trained with limited data by introducing additional loss terms that constrain or guide the learning process. They succeed because these regularizers can effectively suppress issues such as overfitting, mode collapse, or distributional drift without adding new parameters or modifying the network architecture, making them computationally lightweight and easy to integrate. However, they often fail to generalize across different architectures or generative paradigms. Moreover, many of these regularizers are tailored to GAN-based frameworks and do not directly translate to diffusion or autoregressive models, limiting their broader applicability. This highlights a fundamental limitation: the effectiveness of regularization depends heavily on the model’s internal design and objective formulation. The deeper design principle is that successful regularizer design should be model-aware and grounded in the underlying training dynamics, ensuring adaptability across architectures while preserving stability and generalization in low-data generative modeling.

{\em Contrastive learning} approaches enhance generative modeling under limited data by introducing self-supervised pretext tasks that do not require labeled samples. They succeed because contrastive objectives encourage the model to learn discriminative yet semantically meaningful representations, providing an auxiliary supervisory signal that improves the quality and robustness of generated samples. By aligning representations of similar samples and separating dissimilar ones, these methods strengthen the discriminator’s feature space and improve the generator’s ability to capture fine-grained variations. However, current approaches are fundamentally tied to the adversarial framework, as the contrastive loss is typically applied within the discriminator. This restricts their applicability to GAN-based models and prevents straightforward extension to architectures such as VQ-VAE or diffusion models, which lack an explicit discriminator. The core limitation lies in the architectural dependency of contrastive learning, which confines its potential to a subset of generative paradigms. The deeper design principle is that effective integration of contrastive learning into generative modeling requires rethinking the placement of the contrastive objective, developing architecture-agnostic formulations that preserve its representational benefits while remaining compatible with non-adversarial generative frameworks.

{\em Masking} approaches improve generative modeling under limited data by intentionally increasing learning difficulty, which forces the model to capture more generalizable and semantically meaningful representations rather than memorizing training samples. They succeed because masking compels the model to infer missing information from contextual cues, strengthening its ability to model global dependencies and enhancing robustness to data scarcity. However, they often fail when the masking strategy is poorly designed. An inappropriate masking ratio can make the task too trivial, allowing shortcut learning, or too difficult, causing optimization instability and degraded generation quality. In addition, misguided choices about what and how to mask may bias the model toward uninformative regions or hide critical semantic content, limiting its capacity to learn coherent structure. The fundamental limitation is the sensitivity of masking-based learning to both the ratio and spatial distribution of masked regions. The deeper design principle is that effective masking should adapt to the semantic structure of the data and balance task difficulty with information retention so that the model learns representations that generalize beyond the observed samples.

{\em Knowledge distillation} enhances generative modeling with limited data by transferring knowledge from a large, well-trained teacher model to a smaller student model. It succeeds because the teacher provides rich supervisory signals that capture nuanced data statistics and semantic structure, allowing the student model to generalize better even under data scarcity. However, these methods often fail when the distribution of the target data differs significantly from that of the teacher’s training data, which leads to inaccurate or misleading feedback that can destabilize learning. The fundamental limitation lies in the dependency on the teacher model’s domain alignment. The deeper design principle is that effective knowledge distillation for generative modeling should employ adaptive teacher–student interaction, where the teacher selectively guides the student based on relevance to the target data.

{\em Prototype learning} enhances generative modeling under limited data by introducing a clustering-based auxiliary task that groups samples according to shared visual or semantic attributes. It succeeds because these prototypes provide additional structural supervision, encouraging the model to learn more coherent and discriminative representations that capture the underlying data manifold. However, the effectiveness of this approach is highly dependent on how prototypes are defined. Coarse definitions, such as categorizing samples simply as real versus fake, offer little additional information, while overly fine-grained or poorly separated prototypes can introduce noise and ambiguity, especially in datasets with diverse or overlapping categories. This reveals a fundamental limitation: prototype learning is sensitive to the granularity and semantic consistency of the prototype definitions, which may not always align with the true data distribution. The deeper design principle is that effective prototype learning should employ adaptive, semantically grounded prototype formation that balances abstraction and specificity.

\subsubsection{Exploiting Frequency Components}

Approaches based on frequency improve GM-DC by explicitly using frequency information, especially high frequency signals that capture fine details such as edges and textures. They succeed because they counter the common tendency of generative models to favor smooth low frequency content and to ignore detailed patterns. By encouraging awareness of both low and high frequency components, these methods improve image fidelity and reduce artifacts, producing more realistic synthesis. However, they can fail when the dataset is extremely small or highly varied, for example only dozens of images with diverse content. In such cases, the learned frequency cues may not generalize, and performance can degrade further when the data are imbalanced or long-tailed. The fundamental limitation is that current works usually rely on simple frequency extraction approaches such as DCT or Haar wavelets, while more advanced representations like multi-wavelet or wavelet packet transforms have been shown to capture frequency information with higher expressiveness and accuracy. This reliance on limited frequency encoders constrains the model’s ability to represent complex spatial variations and fine-grained details across diverse data. The deeper design principle is that future frequency-based methods should adopt richer, multi-resolution frequency representations that can capture both global structure and local details, enabling more accurate and semantically coherent image synthesis under limited data conditions.

\subsubsection{Meta-Learning}

{\em Optimization-based} methods aim to enhance the adaptability of GMDC by using meta-learning strategies that enable quick convergence to new domains with minimal supervision. They succeed because meta-learning provides well-conditioned initialization states for the generator, allowing efficient adaptation to unseen categories without extensive retraining. However, these methods often fail to synthesize realistic and detailed images, as the generated outputs tend to appear blurry or lack coherent structure, as the meta-objective prioritizes fast convergence over capturing high-frequency details and complex semantics. The fundamental limitation lies in their dependence on the trade-off between rapid adaptation and visual fidelity. The deeper design principle is that effective optimization-based transfer should integrate meta-learning with stable and perceptually aligned training objectives to ensure that the generator maintains both adaptability and realism when generalizing to unseen domains with limited data.

{\em Transformation-based} approaches aim to enhance GM-DC by learning transferable transformation patterns that capture how visual attributes vary within or across categories, and then applying these learned transformations to synthesize novel samples for unseen classes. They succeed because they model intra- and inter-category variations directly in the data space, allowing the generator to reuse structural and appearance-level transformations learned from seen classes. However, they sometimes fail when applied to unseen categories, as the learned transformations may not generalize due to the complex and inconsistent relationships between intra- and inter-category variations. This mismatch can lead to unrealistic or semantically incorrect generations, where the synthesized images lose the distinctive characteristics of the unseen category. A fundamental limitation lies in the assumption that the editing direction for a given attribute is universally shared across categories, which rarely holds true in practice. The deeper design principle is that effective transformation-based learning should incorporate category-aware and context-sensitive modeling of transformations, enabling adaptation to unseen domains while preserving both structural coherence and category identity.

{\em Fusion-based} approaches aim to improve GM-DC by learning to combine or interpolate feature representations from seen classes to generate samples for unseen ones. They succeed because they exploit shared structures and compositional patterns across categories, allowing the model to synthesize novel data by reusing and blending meaningful visual features learned from rich source domains. However, these methods often fail to produce diverse and high-fidelity images, as the fusion process, typically based on simple feature matching or attention-driven interpolation, can blur distinct attributes and introduce visual artifacts. This limitation arises from their reliance on high similarity among input features and the lack of semantic understanding during fusion, which leads to outputs that are overly smooth or too similar to the conditioning examples. The fundamental limitation lies in their shallow fusion strategy, which captures correlations but not causal or hierarchical relationships among features. The deeper design principle is that effective fusion-based generation requires semantically guided and hierarchically structured fusion mechanisms that preserve both shared and distinctive features, enabling richer diversity and sharper visual detail in low-data generative modeling.

\subsubsection{Modeling Internal Patch Distribution}

{\em Progressive Training} methods aim to learn the internal patch distribution of a single image by training a generative model in multiple stages, each responsible for capturing structure at different scales or noise levels. Approaches such as SinGAN achieve impressive results by leveraging self-similarity within an image to generate diverse yet coherent samples that preserve both global layout and fine textures, even from a single example. They succeed because this multi-scale learning effectively prevents overfitting and memorization, allowing realistic variations without requiring multiple training samples. However, these models often fail to capture high-level semantics or meaningful spatial relationships, resulting in incoherent arrangements or distorted objects when generating complex scenes. Their progressive layer-freezing strategy is time-consuming and easy to cause artifacts accumulation. The fundamental limitation lies in the lack of semantic understanding and restricted joint learning across scales. The deeper design principle is that future progressive training frameworks should integrate semantic and structural awareness with multi-scale learning, enabling generative models to achieve both texture fidelity and holistic scene coherence under data-constrained conditions.

{\em Non-progressive Training} methods train a generative model directly at a fixed scale, often by modifying the model architecture or leveraging a small set of correlated observations that capture natural variations in viewpoint, pose, and texture. These approaches succeed by exploiting the inherent diversity present within such correlated samples, effectively expanding the training distribution without the need for explicit data augmentation. This allows the model to achieve richer and more varied generations, even under one-shot or few-shot conditions. However, they often fail when the available samples exhibit significant spatial inconsistencies or complex non-rigid changes, which can lead to fragmented or incoherent object synthesis. These failures arise because the model lacks semantic understanding—it can capture local correlations but struggles to maintain global structure or coherence across variations. The fundamental limitation is that learning remains primarily appearance-driven, without explicit mechanisms to disentangle motion, shape, or identity. The deeper design principle is that effective non-progressive training should incorporate semantic and structure-aware priors that preserve spatial and conceptual consistency while exploiting natural intra-sample diversity, ensuring both fidelity and coherence in generative modeling under data-limited settings.

\subsection{Empirical Comparison}

In this section, we conduct experimental comparsions across tasks. We provide a comprehensive quantitative comparison of representative methods for a number of GM-DC scenarios. We discuss these scenarios because they attract substantial attention or they capture rapidly growing interest. The visual results were produced using either the official implementations or faithful reproductions that strictly follow the methodological details described in the original papers.

\begin{table}[ht]{
\begin{minipage}{0.48\linewidth}
\caption{FID comparisons of representative methods for task uGM-1. Training datasets are 100-shot Obama/Panda/Grumpy Cat images, and AnimalFace (160 cats and 389 dogs) images.}
\label{tab:ugm1}
\centering
\vspace{-0.2em}
\resizebox{\linewidth}{!}{
\begin{tabular}{l|ccccc}
\toprule
Method & Obama & Grumpy Cat & Panda & AFHQ-Cat & AFHQ-Dog \\
\hline
ADA            & 45.69 & 26.62 & 12.90 & 40.77 & 56.83 \\
LeCam          & 33.16 & 24.93 & 10.16 & 34.18 & 54.88 \\
GenCo          & 32.21 & 17.79 &  9.49 & 30.89 & 49.63 \\
InsGen         & 32.42 & 22.01 &  9.85 & 33.01 & 44.93 \\
Diffusion-GAN  & 28.55 & 21.87 &  8.69 & 33.18 & 68.15 \\
FakeCLR        & 26.95 & 19.56 &  8.42 & 26.34 & 42.02 \\
NICE           & 20.09 & 15.63 &  8.18 & 22.70 & 28.65 \\
DANI           & 10.08 & 14.92 &  3.04 & 17.72 & 16.81 \\
\hline
\end{tabular}
}
\end{minipage}
\hspace{0.2cm}
\begin{minipage}{0.48\linewidth}
\caption{Quantitative comparisons of representative methods for task uGM-2. The source model is StyleGAN-2 pretrained on FFHQ for both settings.}
\label{tab:ugm2}
\centering
\resizebox{0.97\linewidth}{!}{
\begin{tabular}{l|ccccc}
\toprule
\multirow{2}{*}{Method} & \multicolumn{2}{c}{FFHQ-Baby} & \multicolumn{2}{c}{AFHQ-Cat} \\
& FID ($\downarrow$) & Intra-LPIPS ($\uparrow$) & FID ($\downarrow$) & Intra-LPIPS ($\uparrow$) \\
\hline
TGAN    & 101.58 & 0.517 &  64.68 & 0.490 \\
FreezeD &  96.25 & 0.518 &  63.60 & 0.492 \\
EWC     &  79.93 & 0.521 &  74.61 & 0.587 \\
CDC     &  69.13 & 0.578 & 176.21 & \textbf{0.629} \\
DCL     &  56.48 & 0.580 & 156.82 & 0.616 \\
AdAM    &  48.83 & 0.590 &  58.07 & 0.557 \\
RICK    &  \textbf{39.39} & \textbf{0.608} & \textbf{53.27} & 0.569 \\
\hline
\end{tabular}
}
\end{minipage}
}
\end{table}

\begin{figure*}[htbp]
    \centering
    \includegraphics[width=0.9\textwidth]{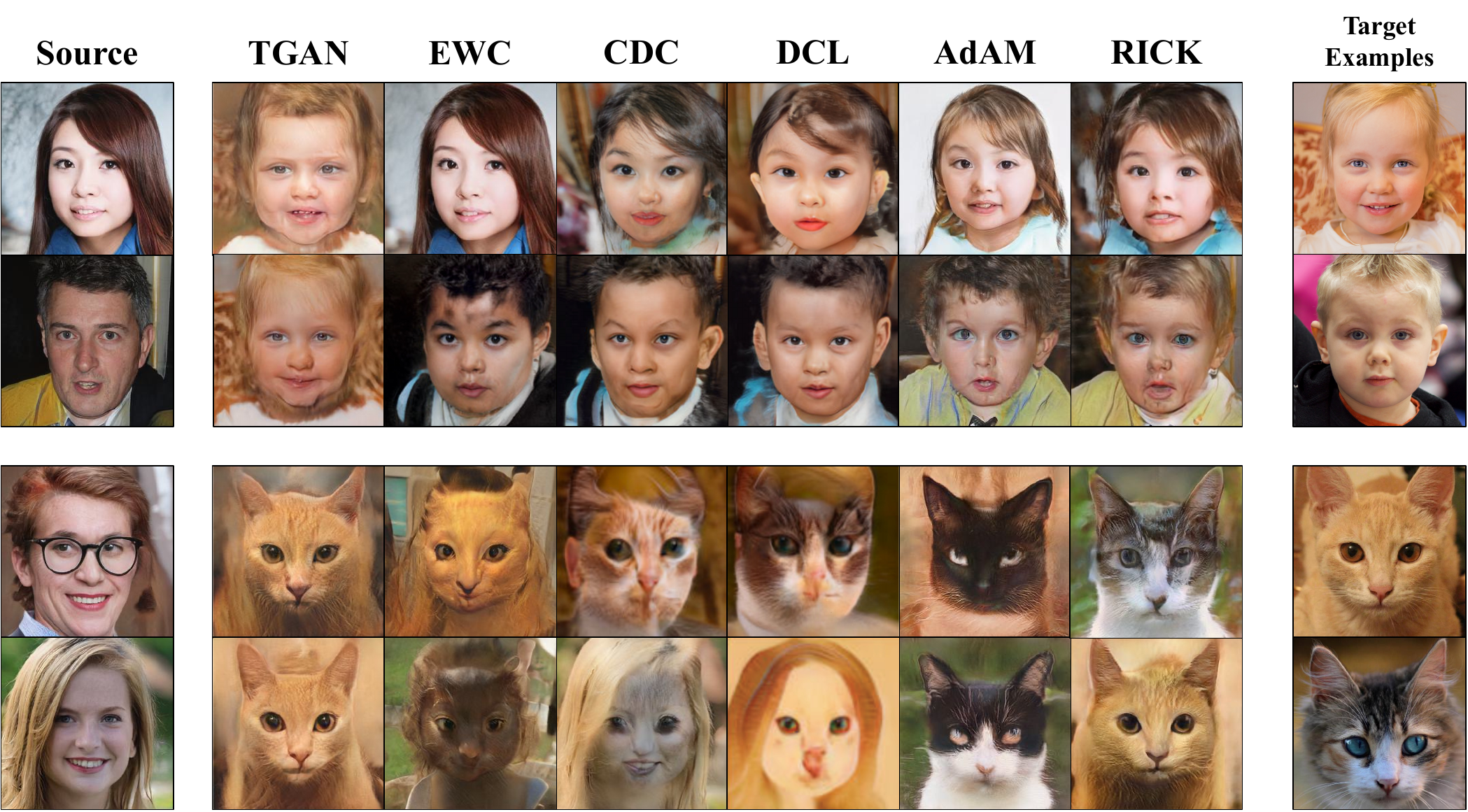}
    \caption{Qualitative comparisons of uGM-2 methods. The source model is StyleGAN-2 pretrained on FFHQ for both settings, and the target domains are 10-shot FFHQ-Baby and 10-shot AFHQ-Cat respectively.
    }
    \label{fig:qualitative_ugm2}
\end{figure*}

\textbf{uGM-1.}
Tab \ref{tab:ugm1} presents the FID comparison results for task uGM-1. Across all datasets, DANI achieves the best scores, substantially outperforming earlier methods such as ADA, LeCam, and GenCo. This demonstrates DANI’s superior ability to generate high-quality and diverse images through more effective normalization and adaptation strategies. In contrast, early approaches like ADA and LeCam produce higher FIDs due to limited feature fusion capability and training instability. Intermediate methods such as FakeCLR and NICE benefit from enhanced contrastive learning and normalization, achieving lower FID values but still falling short of DANI, particularly on challenging datasets like AFHQ-Dog.

\textbf{uGM-2.}
The results in Tab. \ref{tab:ugm2} and Fig. \ref{fig:qualitative_ugm2} show a clear progression from early transfer-based approaches to more adaptation-aware methods. RICK achieves the best overall performance on both target domains, with the lowest FID and highest Intra-LPIPS, indicating superior adaptation and diversity. Earlier approaches such as TGAN and FreezeD perform worse because they rely heavily on fixed discriminator transfer and lack mechanisms to preserve domain-specific knowledge during adaptation. EWC stabilizes learning by constraining parameter changes but can limit adaptation when domain gaps are large, as reflected in its uneven FID results. CDC and DCL improve diversity through correspondence and contrastive learning but are sensitive to similarity metrics and training stability. AdAM bridges these limitations through adaptive module selection, and RICK refines this by prioritizing useful kernels during adaptation, preventing negative transfer and yielding the best trade-off between fidelity and diversity.

\begin{table}[ht]{
\begin{minipage}{0.58\linewidth}
\caption{Quantitative comparisons of representative methods for task uGM-3. The source model is StyleGAN-2 pretrained on FFHQ for both settings.}
\label{tab:ugm3}
\centering
\resizebox{\linewidth}{!}{
    \begin{tabular}{l|cccc}
    \toprule
    \multirow{2}{*}{Method} & \multicolumn{2}{c}{Human $\rightarrow$ Werewolf} & \multicolumn{2}{c}{Photo $\rightarrow$ Sketch} \\
    & CLIP-I ($\uparrow$) & Intra-LPIPS ($\uparrow$) & CLIP-I ($\uparrow$) & Intra-LPIPS ($\uparrow$) \\
    \hline
    NADA   & 0.653 & 0.430 & 0.639 & 0.419 \\
    IPL    & 0.680 & 0.439 & 0.600 & 0.429 \\
    SVL    & 0.600 & 0.431 & 0.591 & \textbf{0.448} \\
    AIR    & \textbf{0.757} & \textbf{0.441} & \textbf{0.687} & 0.426 \\
    \hline
    \end{tabular}
}
\end{minipage}
\hspace{0.2cm}
\begin{minipage}{0.38\linewidth}
\caption{Quantitative comparisons of representative methods for task cGM-1. Experimental setting follows CbC to use a subset of the datasets with 20 classes.}
\label{tab:cgm1}
\centering
\resizebox{0.7\linewidth}{!}{
    \begin{tabular}{l|rr}
    \toprule
    Method & Food101 & AFHQ \\
    \hline
    ADA       & 111.65 & 90.11 \\
    DiffAug   & 28.70  & 25.09 \\
    CbC       & \textbf{20.12} & \textbf{16.25} \\
    \hline
    \end{tabular}
}
\end{minipage}
}
\end{table}

\begin{figure*}[htbp]
    \centering
    \includegraphics[width=0.9\textwidth]{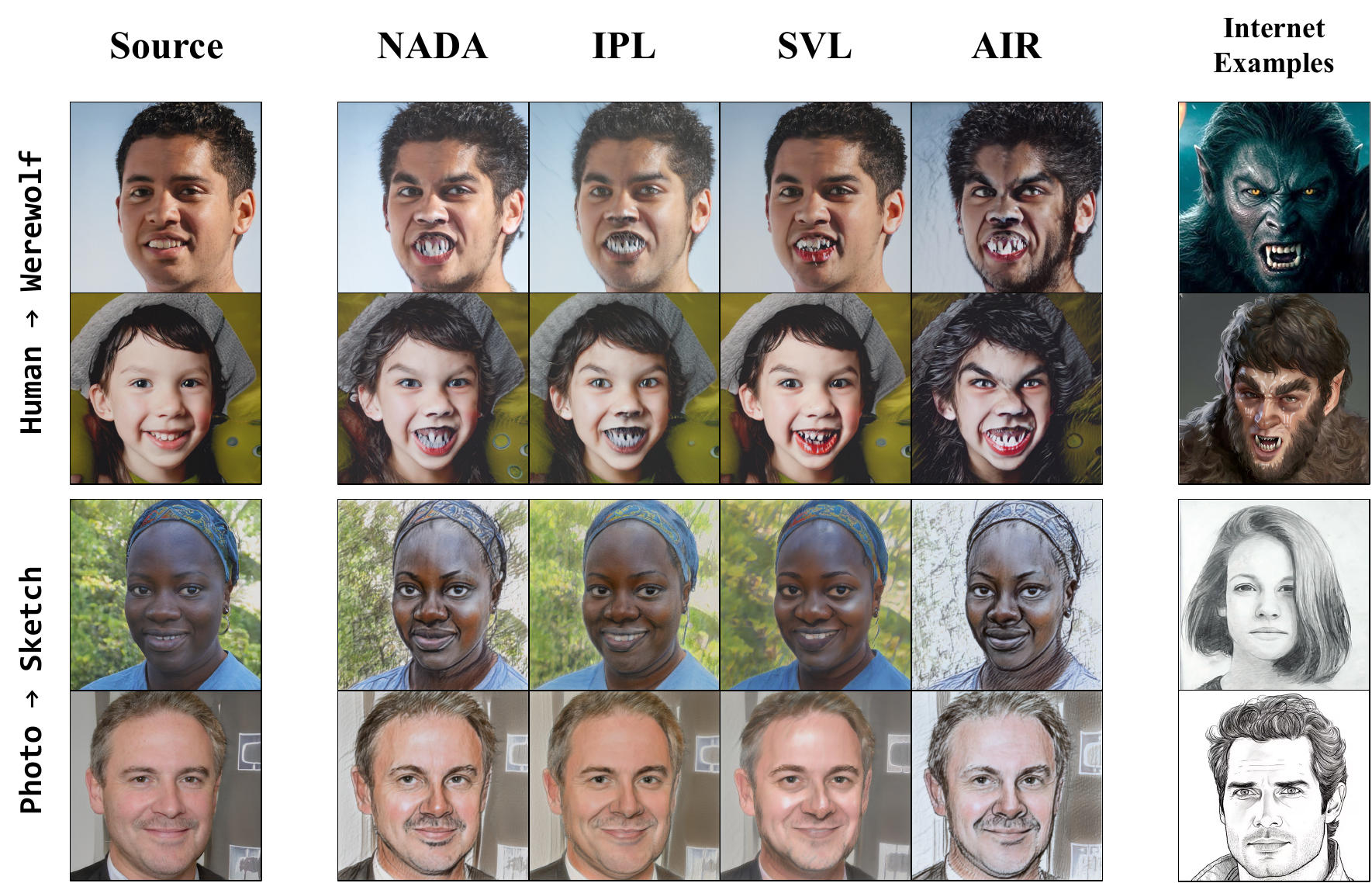}
    \caption{Qualitative comparisons of uGM-3 methods. The source model is StyleGAN-2 pretrained on FFHQ for both settings.
    }
    \label{fig:qualitative_ugm3}
\end{figure*}

\textbf{uGM-3.}
In the comparison among uGM-3 methods in Tab. \ref{tab:ugm3}, AIR achieves the highest CLIP-I and competitive Intra-LPIPS scores, demonstrating strong semantic alignment and image diversity. NADA provides an effective baseline using CLIP-guided text–image offsets but struggles with mode collapse due to reliance on single-prompt direction. IPL and SVL mitigate this issue by introducing sample-specific or distributional guidance, improving diversity (higher Intra-LPIPS), though their assumptions about perfect alignment in CLIP space limit performance on distant domains. AIR further advances this by explicitly addressing offset misalignment through iterative anchor sampling, achieving the most consistent results across both adaptation tasks.

\textbf{cGM-1.}
Tab \ref{tab:cgm1} reports the results for task cGM-1. CbC achieves the lowest FID on both Food101 and AFHQ datasets, outperforming ADA and DiffAug. This demonstrates that class-based contrastive strategies are particularly effective in improving generative quality and inter-class separability. While DiffAug improves stability compared to ADA, it still faces challenges with augmentation leakage, and ADA performs worst under constrained data due to overfitting tendencies.

\begin{table}[ht]{
\begin{minipage}{0.48\linewidth}
\caption{Quantitative comparisons of representative methods for task cGM-2.}
\label{tab:cgm2}
\centering
\vspace{-0.2em}
\resizebox{\linewidth}{!}{
    \begin{tabular}{l|cccc}
    \toprule
    \multirow{2}{*}{Method} & \multicolumn{2}{c}{Flowers} & \multicolumn{2}{c}{AFHQ} \\
    & FID ($\downarrow$) & LPIPS ($\uparrow$) & FID ($\downarrow$) & LPIPS ($\uparrow$) \\
    \hline
    FIGR         & 190.12 & 0.063 & 211.54 & 0.076 \\
    MatchingGAN  & 143.35 & 0.163 & 148.52 & 0.151 \\
    LoFGAN       & 78.83  & 0.387 & 113.01 & 0.489 \\
    WaveGAN      & 42.17  & 0.399 & 30.35  & 0.508 \\
    AGE          & 45.96  & 0.431 & 28.04  & \textbf{0.558} \\
    F2DGAN       & \textbf{38.26} & \textbf{0.433} & \textbf{25.24} & 0.546 \\
    \hline
    \end{tabular}
}
\end{minipage}
\hspace{0.2cm}
\begin{minipage}{0.48\linewidth}
\caption{FID comparisons of representative methods for task cGM-3. The experimental setting follows that in VPT.}
\label{tab:cgm3}
\centering
\resizebox{\linewidth}{!}{
    \begin{tabular}{l|ccc}
    \toprule
    Method & C101 & Flowers & DTD \\
    \hline
    MineGAN                 & 102.4 & 132.1 & 87.4 \\
    cGANTransfer            &  89.6 &  61.6 & 70.3 \\
    VPT (Non-Autoregressive) & \textbf{72.7} & 57.2 & \textbf{66.1} \\
    VPT (Autoregressive)    &  76.0 & \textbf{56.1} & 92.7 \\
    \hline
    \end{tabular}
}
\end{minipage}
}
\end{table}

\textbf{cGM-2.}
Tab. \ref{tab:cgm2} presents the quantitative comparison for task cGM-2. F2DGAN achieves the best overall results on both Flowers and AFHQ datasets, with the lowest FID and highest LPIPS, confirming the benefit of feature-to-distribution alignment for balanced fidelity and diversity. WaveGAN and AGE also perform competitively, with AGE showing the strongest diversity on AFHQ. Earlier methods such as FIGR, MatchingGAN, and LoFGAN lag behind due to their limited ability to capture complex intra-class variation and maintain global coherence.

\textbf{cGM-3.}
Tab. \ref{tab:cgm3} provides the FID results for task cGM-3. The VPT framework achieves the best performance, especially in its non-autoregressive form, which records the lowest FID scores on C101 and DTD and strong performance on Flowers. This validates the effectiveness of progressive visual prompting in cross-domain class-conditioned generation. cGANTransfer performs moderately well but lacks adaptive prompting mechanisms, while MineGAN shows the weakest performance due to limited fine-tuning efficiency under domain shifts.

\begin{table}[ht]{
\begin{minipage}{0.38\linewidth}
\caption{Quantitative comparisons of representative methods for task IGM evaluated on SinDDM dataset.}
\label{tab:igm}
\centering
\resizebox{\linewidth}{!}{
    \begin{tabular}{l|cc}
    \toprule
    Method & LPIPS ($\uparrow$) & SIFID ($\downarrow$) \\
    \hline
    SinGAN     & 0.187 & 0.155 \\
    ConSinGAN  & 0.155 & 0.095 \\
    GPNN       & 0.107 & \textbf{0.054} \\
    SinDDM     & \textbf{0.219} & 0.343 \\
    \hline
    \end{tabular}
}
\end{minipage}
\hspace{0.2cm}
\begin{minipage}{0.60\linewidth}
\caption{Quantitative comparisons of representative methods for task SGM evaluated on the DreamBooth dataset.}
\
\label{tab:sgm}
\centering
\vspace{-0.2em}
\resizebox{\linewidth}{!}{
    \begin{tabular}{l|cccc}
    \toprule
    Method & DINO ($\uparrow$) & CLIP-I ($\uparrow$) & CLIP-T ($\uparrow$) & Tuning-free \\
    \hline
    Real Images       & 0.774 & 0.885 & $-$  & $-$ \\
    DreamBooth        & \textbf{0.668} & \textbf{0.803} & \textbf{0.305} & \ding{55} \\
    CustomDiffusion   & 0.643 & 0.790 & \textbf{0.305} & \ding{55} \\
    Textual Inversion & 0.569 & 0.780 & 0.255 & \ding{55} \\
    BLIP-Diffusion    & 0.594 & 0.779 & 0.300 & \ding{51} \\
    ELITE             & \textbf{0.621} & 0.771 & 0.293 & \ding{51} \\
    MoMA              & 0.618 & \textbf{0.803} & \textbf{0.348} & \ding{51}
    \\
    \hline
    \end{tabular}
}
\end{minipage}
}
\end{table}

\begin{figure*}[htbp]
    \centering
    \includegraphics[width=0.9\textwidth]{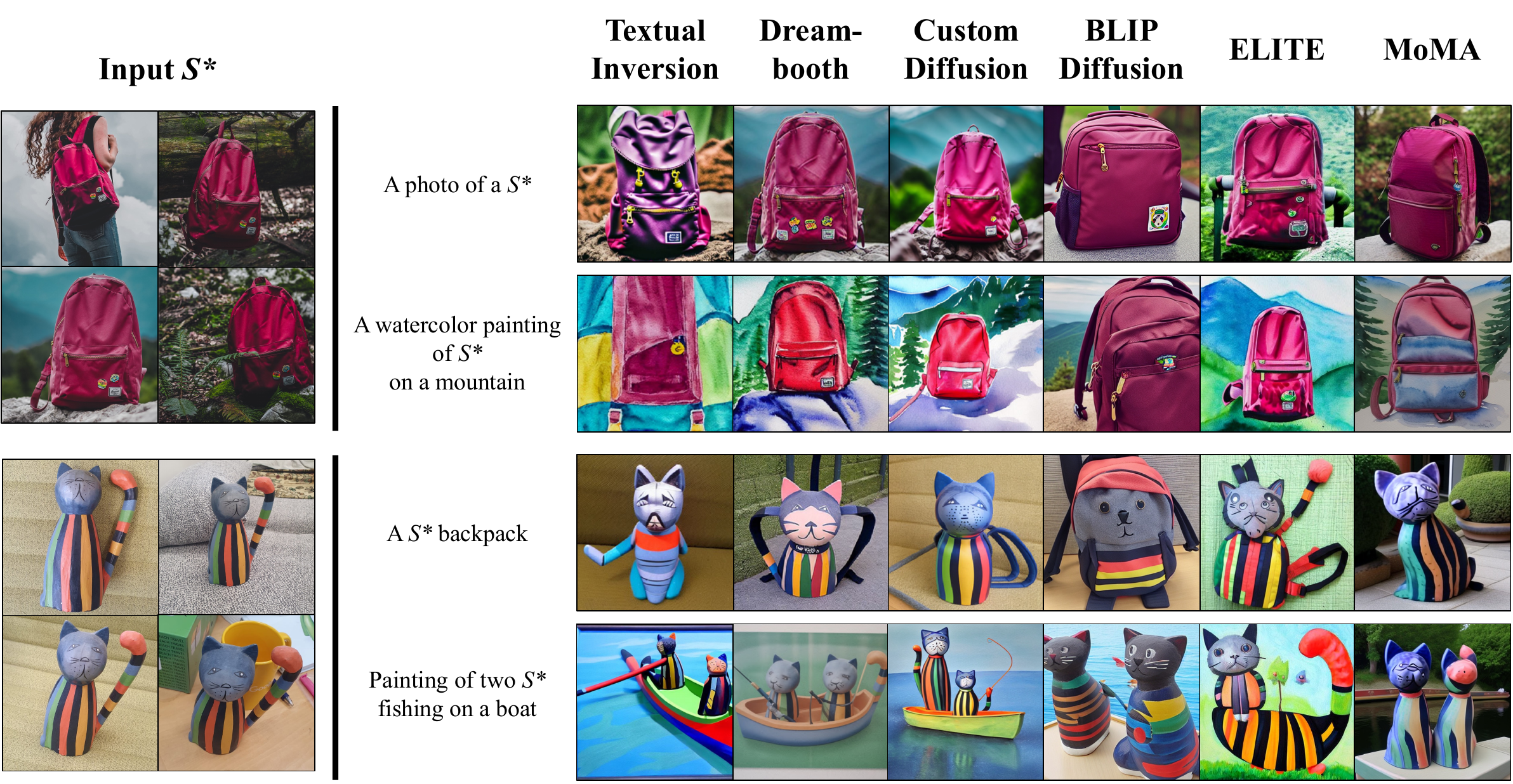}
    \caption{Qualitative comparisons of SGM methods.
    }
    \label{fig:qualitative_sgm}
\end{figure*}

\textbf{IGM.}
Tab. \ref{tab:igm} summarizes results for the image-specific generation task (IGM). SinDDM achieves the highest LPIPS score, indicating superior diversity, while GPNN attains the lowest SIFID, reflecting higher visual fidelity. The contrast between these methods highlights the trade-off between realism and diversity. SinGAN and ConSinGAN perform less effectively, producing less diverse and realistic outputs, suggesting weaker generalization compared to patch-based or diffusion-based approaches.

\textbf{SGM.}
Among subject-driven generation methods, as shwon in Tab. \ref{tab:sgm} and Fig. \ref{fig:qualitative_sgm}, DreamBooth achieves the highest fidelity across metrics but requires fine-tuning, making it computationally heavy and less flexible. Custom Diffusion maintains comparable performance with reduced fine-tuning cost, while Textual Inversion trades off fidelity for efficiency. In contrast, tuning-free methods such as BLIP-Diffusion, ELITE, and MoMA achieve competitive performance without optimization, with MoMA performing best in balancing fidelity, diversity, and efficiency through multimodal modeling. This illustrates a clear evolution from high-cost tuning-based personalization toward more efficient yet expressive tuning-free paradigms.

\section{Discussion}
\label{sec:discussion}
Here, we present an analysis the research landscape,  discuss the research gap and future directions in GM-DC.

\subsection{Analysis of the Research  Landscape}
\label{ssec:landscape_analysis}

In this work, we propose a {\bf taxonomy of eight different tasks for GM-DC} (
Fig.~\ref{fig:sankey},
Tab. \ref{tab:tasktaxonomy}) based on the problem setups of GM-DC publications.
Our investigation of the literature focusing on each task (Fig.~\ref{fig:works_statistics}) reveals that a significant portion of the works (up to 
84\%) 
concentrate on unconditional generation, either through training from scratch or adapting from a pre-trained model. Additionally, zero-shot unconditional generation is beginning to attract more attention.
Similarly, adaptation for in-domain classes has garnered considerable interest for conditional generation.
Meanwhile, conditional generation
for out-of-domain classes
via adaptation 
has not been 
explored adequately.
Furthermore, subject-driven generation, which enables more control over content generation, is an emerging task. 
We anticipate increasing interest on this task as recent text-to-image generative models become more accessible.

We further present a {\bf taxonomy of approaches for GM-DC} (Fig.~\ref{fig:sankey},
Tab.~\ref{tab:approaches}) as our another contribution. Our study reveals that transfer learning is a predominant solution for GM-DC, capable of tackling a large number of tasks (specifically, 5 out of 8 tasks, as indicated in Tab.~\ref{tab:approaches} and Fig.~\ref{fig:sankey}), while effectively handling all data constraints including limited data, few-shot, and zero-shot. 
Moreover, $\approx$54\% of the studies propose new methods based on transfer learning (Fig.~\ref{fig:works_statistics}). More than 
12\% of the studies propose methods
based on other approaches 
that are compatible to transfer learning, \eg data augmentation.
These methods could be used with transfer learning-based methods to improve performance.
The primary challenges in transfer learning are
selection and 
preservation of  source knowledge 
useful for generating 
high-quality and diverse target domain samples.
Adaptation-aware approach \citep{zhao2022adam, zhao2023rick}
could be a sound direction  in this aspect where they consider both source and target domains (the adaptation process) for knowledge preservation.
Language-guided approaches \citep{gal2022stylegannada, ruiz2023dreambooth, kumari2023mcc, liu2025air} are gaining increasing attention due to their ability to facilitate zero-shot generation through appropriate application of vision-language models during the transfer learning phase.
Visual prompt tuning \citep{sohn2023vpt} is a recent method, which guides the generation of target domain samples  by generating visual tokens.

Data augmentation \citep{karras2020ada, tran2021dag, wang2023diffusiongan} remains a potent technique in GM-DC where it boosts performance under limited data by increasing coverage of the data distribution through various transformations.
Multi-task objectives
\citep{yang2021insgen, tseng2021lecam, huang2022maskedgan} 
which incorporate additional learning objectives 
are usually complementary to data augmentation.
Various network architecture designs \citep{liu2021fastgan, li2022moca} that aim to prevent overfitting or preserve the feature maps are also shown to be 
significantly
effective for GM-DC.
Given that generative models tend to exhibit biases in capturing frequency components, enhancing the frequency awareness in these models is an emerging direction for GM-DC \citep{yang2022fregan}.
Meta-learning \citep{clouatre2019figr} enables generative models to learn 
inter-task 
knowledge from seen classes, and then handle new generation tasks from unseen classes usually without fine-tuning \citep{gu2021lofgan, hong2022deltagan}.
Internal patch-distribution modeling  \citep{shaham2019singan, nikankin2022sinfusion} effectively trains a generative model from scratch using a single reference image (scene) to produce novel scene compositions.

Regarding the types of generating models, 
our study shows that around 68\% of the GM-DC works
focus on GANs
(Fig.~\ref{fig:works_statistics}).
This preference can be attributed to the extensive research in GANs.
Recently, there has been a growing interest in DMs (30\%) 
and VAEs (2\%), particularly VQ-VAE, driven by the success of DMs \citep{ramesh2022dalle2, saharia2022imagen} and transformer-based token prediction methods in generative modeling \citep{chang2022maskgit, esser2021vqgan}.
We anticipate increasing attention directed toward DMs and VQ-VAEs.
Furthermore, our survey reveals an interesting trend: around 
71\% of the works focus on addressing the challenging task of few-shot learning, while 
26\% concentrate on limited data scenarios. 
While only 3\% of works address zero-shot learning, we expect growing interest due to recent advancements in vision-language models \citep{kwon2022oneclip, li2023scaling}.

\subsection{Trends of Works across Categories}

\begin{figure}[t]
  \centering
  \begin{minipage}[b]{0.48\textwidth}
    \centering
    \includegraphics[width=\textwidth]{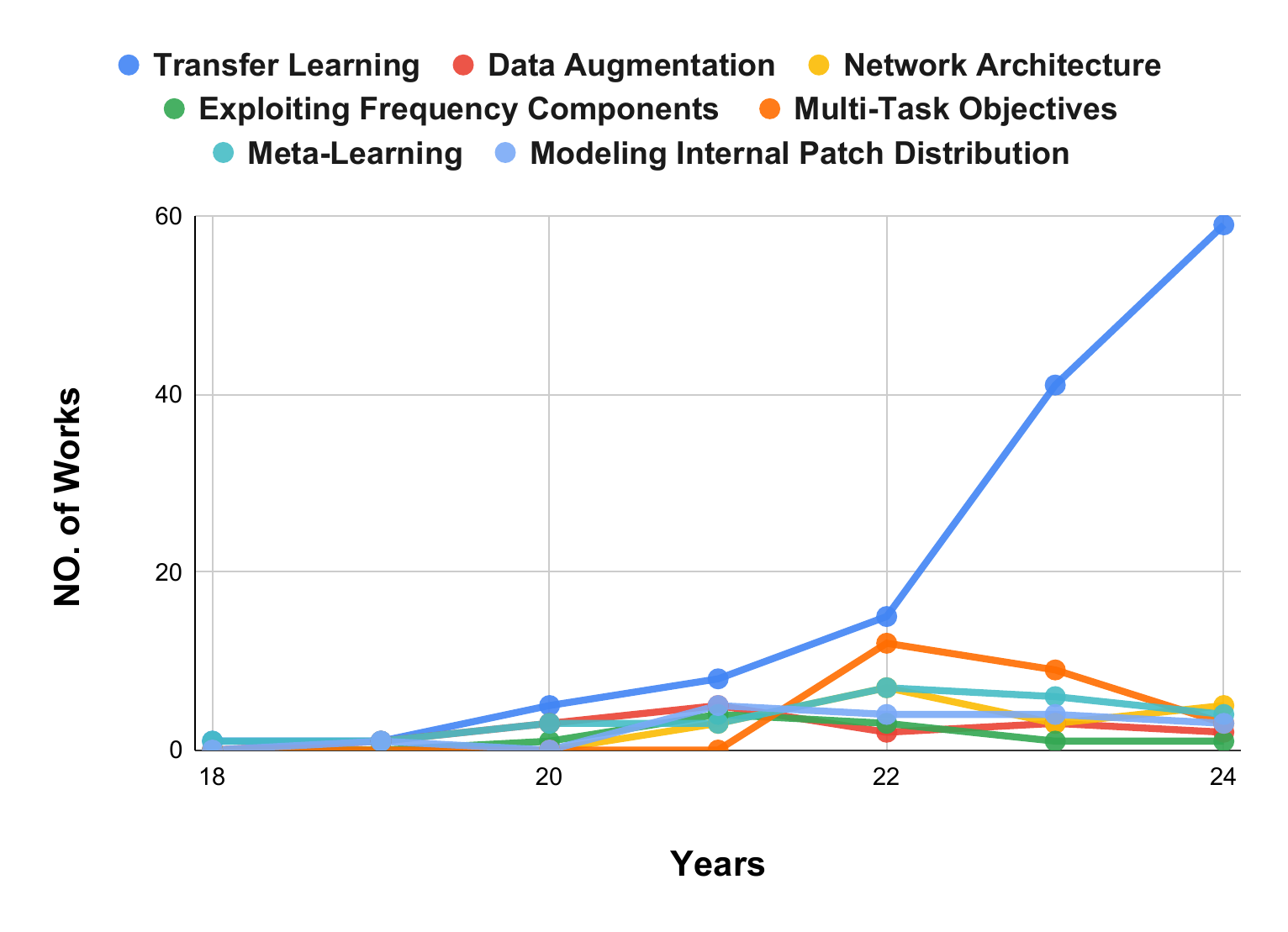}
    \caption{Trends of works in all categories.}
    \label{fig:trends_all}
  \end{minipage}
  \begin{minipage}[b]{0.48\textwidth}
    \centering
    \includegraphics[width=\textwidth]{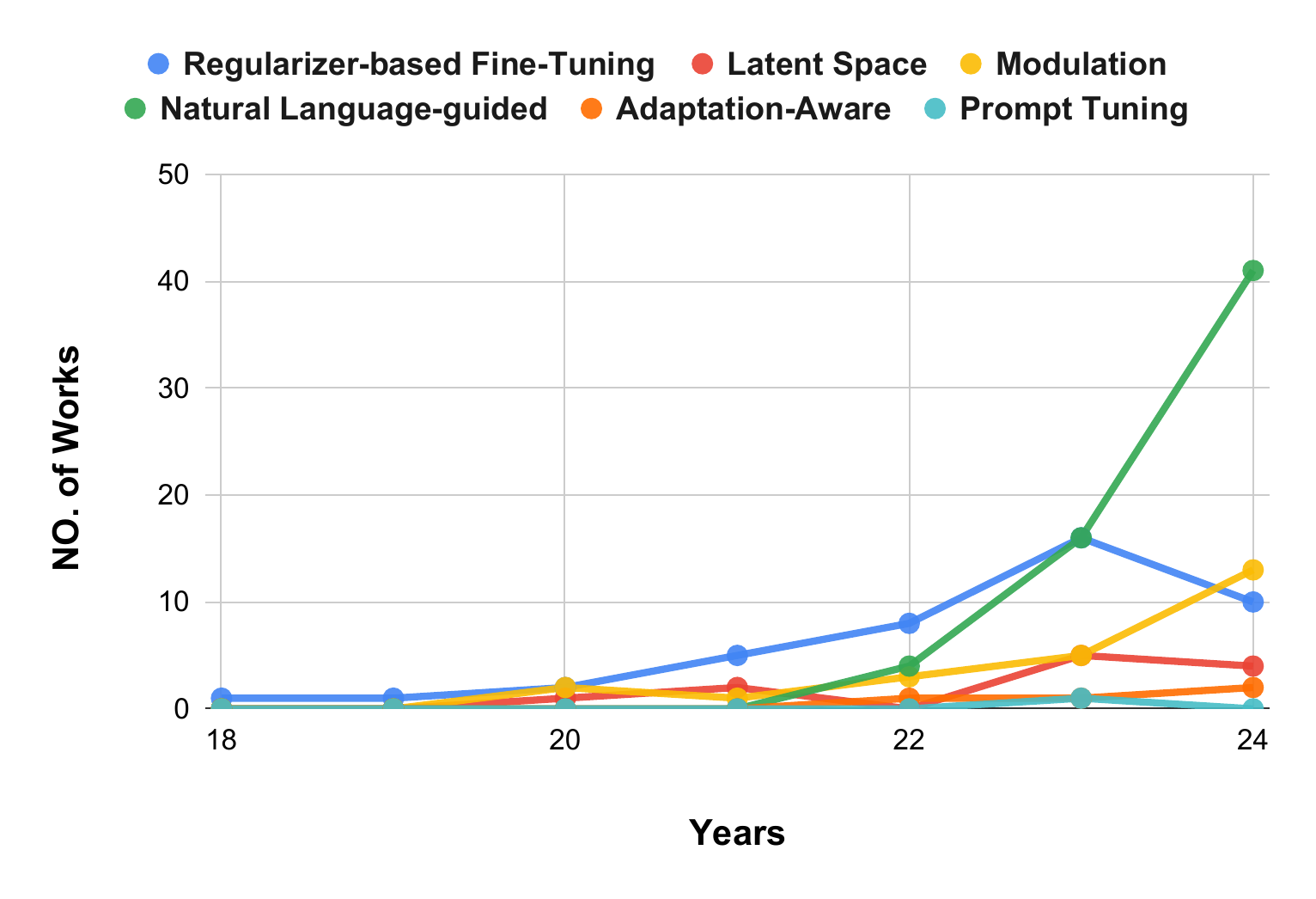}
    \caption{Trends of works in Transfer Learning.}
    \label{fig:trends_tl}
  \end{minipage}
\end{figure}

In this section, we discuss the trends of works across categories.  Overall, as illustrated in Fig. \ref{fig:trends_all}, Transfer Learning has consolidated as the predominant direction, while most other categories remain relatively minor. Specifically, Transfer Learning accounts for 29\% of works in 2021 and increasing to 77\% in 2024. By contrast, Data Augmentation, Network Architecture, Exploiting Frequency Components, Meta-Learning, and Modeling Internal Patch Distribution all maintain relatively low percentages of works, generally below 15\%. Only Multi-Task Objectives experienced a temporary peak, reaching 24\% in 2022, but subsequently declined.

Given its rapid growth and central role, we provide a more detailed analysis of Transfer Learning. As shown in Fig. \ref{fig:trends_tl}, within this category, we observe a gradual shift in emphasis over time. Earlier works primarily adopted Regularizer-based Fine-Tuning (40–63\% between 2020–2021). More recently, Natural Language-guided approaches have increased substantially, reaching 59\% in 2024. This development is closely related to the rise of multimodal foundation models such as CLIP and the introduction of new tasks like Subject-Driven Generation, which rely on language as a flexible control signal. Modulation methods also show an increasing trend, growing from 11\% to 19\%, supported by the adoption of parameter-efficient fine-tuning (PEFT) approaches such as LoRA. Prompt Tuning and Adaptation-Aware methods have also appeared in recent years, though their adoption remains limited.

\subsection{Research Gap and Future Directions}
\label{ssec:future_direction}

While GM-DC has witnessed steady progress, several important research gaps remain that limit its advancement and broader applicability. Addressing these gaps requires moving beyond incremental improvements on existing setups and instead rethinking the foundations of data-constrained generative modeling. In particular, future work should consider leveraging foundation models more effectively, grounding zero-shot generation for evolving concepts, enabling transfer to distant target domains, developing holistic evaluation, and adopting data-centric strategies. The following subsections detail these directions.

\subsubsection{Harnessing the power of foundation models}
As previously discussed, transfer learning is a prominent and highly effective solution for GM-DC.
Nevertheless, the majority of existing literature uses pre-trained StyleGAN2 (FFHQ) or BigGAN (ImageNet) networks as source models.
A potential future direction for GM-DC is to explore the capabilities of foundation models \citep{bommasani2021opportunities},  \ie large models trained using massive amounts of data.
In particular, text-image generation models including DALL$\cdot$E-2 \citep{ramesh2022dalle2} ($\approx$3.5B parameters), Imagen \citep{saharia2022imagen} ($\approx$4.6B parameters) and 
Stable Diffusion 3.5 \citep{esser2024scaling} ($\approx$8.1B parameters) 
encode knowledge regarding a wide range of concepts for high-quality, diverse image generation.
Leveraging such foundation models for GM-DC is relatively under-explored.

\subsubsection{Grounding zero-shot image generative capabilities}
Recent studies have demonstrated the feasibility of zero-shot image generation for 
well-known concepts, \eg ``Tolkien Elf'' \citep{gal2022stylegannada}. 
However, grounding zero-shot image generation models to generate evolving/ new semantic concepts remains a relatively unexplored and challenging area. 
For instance, how to generate an image depicting 
``Mass for the Beginning of the Petrine Ministry of the Bishop of Rome,'' an event that occurred in May 2025, 
that related images may not be captured by existing models. 
This requires strategies that allow continual learning, semantic concept editing, and the incorporation of temporal contexts.

\subsubsection{
Knowledge transfer 
for distant/ remote target domains}
Knowledge transfer has received significant attention in GM-DC research. Many works concentrate on utilizing pre-trained knowledge
of a source domain
to enhance learning in the target domain, as evident from the statistics in Fig. \ref{fig:sankey} and Fig. \ref{fig:works_statistics}. However, we remark that exploring knowledge transfer for modeling
target domains which are distant/ remote from the source domains 
still remains largely unexplored.
This problem is challenging, as demonstrated in our experiment to transfer knowledge from Human Faces $\rightarrow$ Flowers 
(Fig. \ref{fig:proximity-measurements-and-10-shot-flowers}), which clearly demonstrates the complexity of the task.
We urge more investigation in knowledge transfer for modeling distant/ remote target domains in GM-DC research.

\subsubsection{Holistic evaluation of GM-DC}
Evaluation of GM-DC presents multiple challenges including difficulties in estimating real data statistics under low-data regimes, lack of unified framework for human evaluation of GM-DC samples, and 
heavy reliance 
on
particular 
(pre-trained) feature extractors to quantify the capabilities of GM-DC. 
In particular, developing holistic evaluation frameworks integrating both objective measurements and subjective judgements tailored for different tasks is essential for understanding GM-DC capabilities.
Advancing holistic evaluation
is important for 
GM-DC methods to be applied in a 
variety of real-world scenarios.

\subsubsection{Data-centric approaches for GM-DC}
We remark that data-centric  approaches
\citep{whang2023datacollection}
for advancing  GM-DC  have been relatively overlooked in the literature.
Majority of GM-DC methods focus on advancing training procedures based on a given set of training samples, but little attention has been put on how GM-DC performance may be affected by characteristics of the given training samples.
Particularly, for 
GM-DC problems, where a domain is described using limited training samples, the characteristics of the  samples can have noticeable impact on performance of GM-DC methods, as hinted in our analysis (see Fig.
\ref{fig:challenges-data-selection}).
We suggest  
 greater emphasis on data collection, curation and pre-processing for GM-DC advancement.

\subsection{Beyond Image Generation}

Existing GM-DC works focus on image generation primarily.
There are a few works to study other data types.
\citet{zhu2023fewshot_3d} studies {\em 3D shape generation} under few-shot target data (10-shot) utilizing pre-trained 3D generative models and optimization adaptation to retain the probability distributions of pairwise adapted samples. 
CLIP-Sculptor \citep{sanghi2023clipsculptor, kim2023datid3d, kim2023podia} leverages CLIP guidance for zero-shot
{\em
3D generation}. 
\citet{wang2023cffont, yang2024fontdiffuser} studies few-shot {\em font generation} which aims to transfer the source domain style to the target domain. In particular, they introduce a content fusion module and a projected character loss to improve the quality of skeleton transfer for few-shot font generation.
\citet{careil2023few} explores the problem of few-shot {\em semantic image generation} where the objective is to generate realistic images based on semantic segmentation maps. Their approach employs transfer learning on both GANs and DMs for few-shot semantic image synthesis. \citet{couairon2023zero} further extend it in a zero-shot manner on Stable Diffusion \citep{rombach2022latentdiffusion}.

\section{Conclusion}
\label{sec:conclusion}

Generative Modeling under Data Constraints (GM-DC) is an important research area. 
This survey delves into this field by meticulously examining
research papers in this area, 
encompassing 
different types of
generative models including VAEs, GANs, and Diffusion Models. 
Drawing from this analysis, we identify  several challenges encountered in GM-DC, including those related to training, data selection, and model evaluation. 
Moreover, we propose two taxonomies to categorize works related to  GM-DC: a task taxonomy that
identifies the variety  of 
generation 
tasks, and an approach taxonomy that categorizes 
the  extensive list of solutions for these tasks. 
We present a Sankey diagram to illuminate the interactions between different GM-DC tasks, approaches, and methods.
Additionally, we present
an organized review 
of existing GM-DC works and discuss research gaps and  future research directions. 
Our aspiration is that this survey not only could offer valuable insights to researchers but also help spark further advancements in GM-DC.

\subsubsection*{Ethics Statement}
Generative models could be mis-used to disseminate mis- and disinformation 
due to their ability to generate realistic content.
In particular, advanced generative models could be mis-used by malicious users to fabricate deepfake images,  
portraying individuals engaging in actions they never actually performed.
Advances in GM-DC could exacerbate the situation as it becomes possible to 
generate realistic content with less data.
We advocate for ethical and responsible usage of GM-DC methods and studying of  mitigation techniques \citep{mirsky2021creation, chandrasegaran2021closer, Chandrasegaran_2022_ECCV, zhao2023recipe, wen2023tree, doloriel2024frequency}.

\subsubsection*{Acknowledgments}
This research is supported by the National Research Foundation, Singapore under its AI Singapore Programmes (AISG Award No.: AISG2-TC-2022-007); The Agency for Science, Technology and Research (A*STAR) under its MTC Programmatic Funds (Grant No. M23L7b0021). This research is supported by the National Research Foundation, Singapore and Infocomm Media Development Authority under its Trust Tech Funding Initiative. Any opinions, findings and conclusions or recommendations expressed in this material are those of the author(s) and do not reflect the views of National Research Foundation, Singapore and Infocomm Media
Development Authority.

\bibliography{main}

\begin{thebibliography}{364}
\providecommand{\natexlab}[1]{#1}
\providecommand{\url}[1]{\texttt{#1}}
\expandafter\ifx\csname urlstyle\endcsname\relax
  \providecommand{\doi}[1]{doi: #1}\else
  \providecommand{\doi}{doi: \begingroup \urlstyle{rm}\Url}\fi

\bibitem[Abdollahzadeh et~al.(2021)Abdollahzadeh, Malekzadeh, and Cheung]{milad2021revisit}
Milad Abdollahzadeh, Touba Malekzadeh, and Ngai-Man~Man Cheung.
\newblock Revisit multimodal meta-learning through the lens of multi-task learning.
\newblock In \emph{Advances in Neural Information Processing Systems}, 2021.

\bibitem[Ackley et~al.(1985)Ackley, Hinton, and Sejnowski]{ackley1985learning}
David~H Ackley, Geoffrey~E Hinton, and Terrence~J Sejnowski.
\newblock A learning algorithm for boltzmann machines.
\newblock \emph{Cognitive science}, 9\penalty0 (1):\penalty0 147--169, 1985.

\bibitem[Aghabozorgi et~al.(2023)Aghabozorgi, Peng, and Li]{aghabozorgi2023adaptiveimle}
Mehran Aghabozorgi, Shichong Peng, and Ke~Li.
\newblock Adaptive imle for few-shot pretraining-free generative modelling.
\newblock In \emph{Proceedings of the International Conference on Machine Learning}, 2023.

\bibitem[Aksac et~al.(2019)Aksac, Demetrick, Ozyer, and Alhajj]{aksac2019brecahad}
Alper Aksac, Douglas~J Demetrick, Tansel Ozyer, and Reda Alhajj.
\newblock Brecahad: A dataset for breast cancer histopathological annotation and diagnosis.
\newblock \emph{BMC research notes}, 12\penalty0 (1):\penalty0 1--3, 2019.

\bibitem[Alanov et~al.(2022)Alanov, Titov, and Vetrov]{alanov2022hyperdomainnet}
Aibek Alanov, Vadim Titov, and Dmitry~P Vetrov.
\newblock Hyperdomainnet: Universal domain adaptation for generative adversarial networks.
\newblock In \emph{Advances in Neural Information Processing Systems}, 2022.

\bibitem[Alanov et~al.(2023)Alanov, Titov, Nakhodnov, and Vetrov]{alanov2023styledomain}
Aibek Alanov, Vadim Titov, Maksim Nakhodnov, and Dmitry Vetrov.
\newblock Styledomain: Efficient and lightweight parameterizations of stylegan for one-shot and few-shot domain adaptation.
\newblock In \emph{Proceedings of the IEEE/CVF International Conference on Computer Vision}, pp.\  2184--2194, 2023.

\bibitem[Aldhubri et~al.(2024)Aldhubri, Lu, and Fu]{aldhubri2024sagan}
Ali Aldhubri, Jianfeng Lu, and Guanyiman Fu.
\newblock Sagan: Skip attention generative adversarial networks for few-shot image generation.
\newblock \emph{Digital Signal Processing}, 149:\penalty0 104466, 2024.

\bibitem[Ali et~al.(2025)Ali, Rossi, and Bertozzi]{ali2025cfts}
Munsif Ali, Leonardo Rossi, and Massimo Bertozzi.
\newblock Cfts-gan: Continual few-shot teacher student for generative adversarial networks.
\newblock In \emph{International Conference on Pattern Recognition}, pp.\  249--262. Springer, 2025.

\bibitem[Anees et~al.(2024)Anees, Baykal, Kizil, Ceylan, Erdem, and Erdem]{anees2024hypergan}
Abdul~Basit Anees, Ahmet~Canberk Baykal, Muhammed~Burak Kizil, Duygu Ceylan, Erkut Erdem, and Aykut Erdem.
\newblock Hypergan-clip: A unified framework for domain adaptation, image synthesis and manipulation.
\newblock In \emph{SIGGRAPH Asia 2024 Conference Papers}, pp.\  1--12, 2024.

\bibitem[Antoniou et~al.(2017)Antoniou, Storkey, and Edwards]{antoniou2017dagan}
Antreas Antoniou, Amos Storkey, and Harrison Edwards.
\newblock Data augmentation generative adversarial networks.
\newblock \emph{arXiv preprint arXiv:1711.04340}, 2017.

\bibitem[Arar et~al.(2024)Arar, Voynov, Hertz, Avrahami, Fruchter, Pritch, Cohen-Or, and Shamir]{arar2024palp}
Moab Arar, Andrey Voynov, Amir Hertz, Omri Avrahami, Shlomi Fruchter, Yael Pritch, Daniel Cohen-Or, and Ariel Shamir.
\newblock Palp: prompt aligned personalization of text-to-image models.
\newblock In \emph{SIGGRAPH Asia 2024 Conference Papers}, pp.\  1--11, 2024.

\bibitem[Arjovsky et~al.(2017)Arjovsky, Chintala, and Bottou]{arjovsky2017wasserstein}
Martin Arjovsky, Soumith Chintala, and L{\'e}on Bottou.
\newblock Wasserstein generative adversarial networks.
\newblock In \emph{Proceedings of the International Conference on Machine Learning}, 2017.

\bibitem[Asokan \& Seelamantula(2023)Asokan and Seelamantula]{asokan2023spider}
Siddarth Asokan and Chandra~Sekhar Seelamantula.
\newblock Spider gan: Leveraging friendly neighbors to accelerate gan training.
\newblock In \emph{Proceedings of the IEEE/CVF Conference on Computer Vision and Pattern Recognition}, pp.\  3883--3893, 2023.

\bibitem[Bakare(2025)]{lanre-2025}
Lanre Bakare.
\newblock {Super Trouper meets supercomputer: AI helping Abba star to write musical}.
\newblock 2025.
\newblock URL \url{https://www.theguardian.com/music/2025/jun/04/abba-bjorn-ulvaeus-ai-musical-london}.

\bibitem[Bar-Tal et~al.(2023)Bar-Tal, Yariv, Lipman, and Dekel]{bar2023multidiffusion}
Omer Bar-Tal, Lior Yariv, Yaron Lipman, and Tali Dekel.
\newblock Multidiffusion: Fusing diffusion paths for controlled image generation.
\newblock In \emph{Proceedings of the International Conference on Machine Learning}, 2023.

\bibitem[Bartunov \& Vetrov(2018)Bartunov and Vetrov]{bartunov2018few}
Sergey Bartunov and Dmitry Vetrov.
\newblock Few-shot generative modeling with generative matching networks.
\newblock In \emph{International Conference on Artificial Intelligence and Statistics}, 2018.

\bibitem[Bie et~al.(2024)Bie, Yang, Zhou, Ghanem, Zhang, Yao, Wu, Holmes, Golnari, Clifton, et~al.]{bie2024renaissance}
Fengxiang Bie, Yibo Yang, Zhongzhu Zhou, Adam Ghanem, Minjia Zhang, Zhewei Yao, Xiaoxia Wu, Connor Holmes, Pareesa Golnari, David~A Clifton, et~al.
\newblock Renaissance: A survey into ai text-to-image generation in the era of large model.
\newblock \emph{IEEE transactions on pattern analysis and machine intelligence}, 2024.

\bibitem[Bińkowski et~al.(2018)Bińkowski, Sutherland, Arbel, and Gretton]{binkowski2018demystifying}
Mikołaj Bińkowski, Dougal~J. Sutherland, Michael Arbel, and Arthur Gretton.
\newblock Demystifying mmd gans.
\newblock In \emph{International Conference on Learning Representations}, 2018.

\bibitem[Bommasani et~al.(2021)Bommasani, Hudson, Adeli, Altman, Arora, von Arx, Bernstein, Bohg, Bosselut, Brunskill, et~al.]{bommasani2021opportunities}
Rishi Bommasani, Drew~A Hudson, Ehsan Adeli, Russ Altman, Simran Arora, Sydney von Arx, Michael~S Bernstein, Jeannette Bohg, Antoine Bosselut, Emma Brunskill, et~al.
\newblock On the opportunities and risks of foundation models.
\newblock \emph{arXiv preprint arXiv:2108.07258}, 2021.

\bibitem[Brock et~al.(2019)Brock, Donahue, and Simonyan]{brock2019biggan}
Andrew Brock, Jeff Donahue, and Karen Simonyan.
\newblock Large scale gan training for high fidelity natural image synthesis.
\newblock In \emph{International Conference on Learning Representations}, 2019.

\bibitem[Cai et~al.(2024)Cai, Wei, Ji, Bai, Han, and Zuo]{cai2024decoupled}
Yufei Cai, Yuxiang Wei, Zhilong Ji, Jinfeng Bai, Hu~Han, and Wangmeng Zuo.
\newblock Decoupled textual embeddings for customized image generation.
\newblock In \emph{Proceedings of the AAAI Conference on Artificial Intelligence}, volume~38, pp.\  909--917, 2024.

\bibitem[Cao \& Gong(2025)Cao and Gong]{cao2025CRDI}
Yu~Cao and Shaogang Gong.
\newblock Few-shot image generation by conditional relaxing diffusion inversion.
\newblock In \emph{European Conference on Computer Vision}, pp.\  20--37. Springer, 2025.

\bibitem[Careil et~al.(2023)Careil, Verbeek, and Lathuili{\`e}re]{careil2023few}
Marl{\`e}ne Careil, Jakob Verbeek, and St{\'e}phane Lathuili{\`e}re.
\newblock Few-shot semantic image synthesis with class affinity transfer.
\newblock In \emph{Proceedings of the IEEE/CVF Conference on Computer Vision and Pattern Recognition}, 2023.

\bibitem[Caron et~al.(2021)Caron, Touvron, Misra, J{\'e}gou, Mairal, Bojanowski, and Joulin]{caron2021dino}
Mathilde Caron, Hugo Touvron, Ishan Misra, Herv{\'e} J{\'e}gou, Julien Mairal, Piotr Bojanowski, and Armand Joulin.
\newblock Emerging properties in self-supervised vision transformers.
\newblock In \emph{Proceedings of the IEEE/CVF international conference on computer vision}, pp.\  9650--9660, 2021.

\bibitem[Casanova et~al.(2021)Casanova, Careil, Verbeek, Drozdzal, and Romero~Soriano]{casanova2021icgan}
Arantxa Casanova, Marlene Careil, Jakob Verbeek, Michal Drozdzal, and Adriana Romero~Soriano.
\newblock Instance-conditioned gan.
\newblock In \emph{Advances in Neural Information Processing Systems}, 2021.

\bibitem[Chai et~al.(2022)Chai, Gharbi, Shechtman, Isola, and Zhang]{chai2022anyresolution}
Lucy Chai, Michael Gharbi, Eli Shechtman, Phillip Isola, and Richard Zhang.
\newblock Any-resolution training for high-resolution image synthesis.
\newblock In \emph{Proceedings of the European Conference on Computer Vision}, 2022.

\bibitem[Chan et~al.(2024)Chan, Zhao, Jia, Yang, and Wang]{chan2024sag}
Kelvin C.~K. Chan, Yang Zhao, Xuhui Jia, Ming-Hsuan Yang, and Huisheng Wang.
\newblock Improving subject-driven image synthesis with subject-agnostic guidance.
\newblock In \emph{IEEE/CVF Conference on Computer Vision and Pattern Recognition}, 2024.

\bibitem[Chandrasegaran et~al.(2022{\natexlab{a}})Chandrasegaran, Tran, Zhao, and Cheung]{chandrasegaran2022revisiting}
K.~Chandrasegaran, N.~T. Tran, Y.~Zhao, and N.~M. Cheung.
\newblock Revisiting label smoothing and knowledge distillation compatibility: What was missing?
\newblock In \emph{Proceedings of the International Conference on Machine Learning}, 2022{\natexlab{a}}.

\bibitem[Chandrasegaran et~al.(2021)Chandrasegaran, Tran, and Cheung]{chandrasegaran2021closer}
Keshigeyan Chandrasegaran, Ngoc-Trung Tran, and Ngai-Man Cheung.
\newblock A closer look at fourier spectrum discrepancies for cnn-generated images detection.
\newblock In \emph{Proceedings of the IEEE/CVF Conference on Computer Vision and Pattern Recognition}, 2021.

\bibitem[Chandrasegaran et~al.(2022{\natexlab{b}})Chandrasegaran, Tran, Binder, and Cheung]{Chandrasegaran_2022_ECCV}
Keshigeyan Chandrasegaran, Ngoc-Trung Tran, Alexander Binder, and Ngai-Man Cheung.
\newblock Discovering transferable forensic features for cnn-generated images detection.
\newblock In \emph{Proceedings of the European Conference on Computer Vision (ECCV)}, Oct 2022{\natexlab{b}}.

\bibitem[Chandrasegaran et~al.(2025)Chandrasegaran, Poli, Fu, Kim, Hadzic, Li, Gupta, Massaroli, Mirhoseini, Niebles, Ermon, and Li]{chandrasegaran2024grafting}
Keshigeyan Chandrasegaran, Michael Poli, Daniel~Y. Fu, Dongjun Kim, Lea~M. Hadzic, Manling Li, Agrim Gupta, Stefano Massaroli, Azalia Mirhoseini, Juan~Carlos Niebles, Stefano Ermon, and Fei-Fei Li.
\newblock Exploring diffusion transformer designs via grafting.
\newblock \emph{arXiv preprint arXiv:2506.05340}, 2025.

\bibitem[Chang et~al.(2022)Chang, Zhang, Jiang, Liu, and Freeman]{chang2022maskgit}
Huiwen Chang, Han Zhang, Lu~Jiang, Ce~Liu, and William~T Freeman.
\newblock Maskgit: Masked generative image transformer.
\newblock In \emph{Proceedings of the IEEE/CVF Conference on Computer Vision and Pattern Recognition}, 2022.

\bibitem[Chauhan \& Shah(2021)Chauhan and Shah]{chauhan2021topic}
Uttam Chauhan and Apurva Shah.
\newblock Topic modeling using latent dirichlet allocation: A survey.
\newblock \emph{ACM Computing Surveys (CSUR)}, 54\penalty0 (7):\penalty0 1--35, 2021.

\bibitem[Chen et~al.(2021{\natexlab{a}})Chen, Cheng, Gan, Liu, and Wang]{chen2021advaug}
Tianlong Chen, Yu~Cheng, Zhe Gan, Jingjing Liu, and Zhangyang Wang.
\newblock Data-efficient gan training beyond (just) augmentations: A lottery ticket perspective.
\newblock In \emph{Advances in Neural Information Processing Systems}, 2021{\natexlab{a}}.

\bibitem[Chen et~al.(2020)Chen, Kornblith, Norouzi, and Hinton]{chen2020simple}
Ting Chen, Simon Kornblith, Mohammad Norouzi, and Geoffrey Hinton.
\newblock A simple framework for contrastive learning of visual representations.
\newblock In \emph{International conference on machine learning}, pp.\  1597--1607. PmLR, 2020.

\bibitem[Chen et~al.(2021{\natexlab{b}})Chen, Zhao, Yang, Li, Kang, and Lu]{chen2021sa}
Xi~Chen, Hongdong Zhao, Dongxu Yang, Yueyuan Li, Qing Kang, and Haiyan Lu.
\newblock Sa-singan: self-attention for single-image generation adversarial networks.
\newblock \emph{Machine Vision and Applications}, 32:\penalty0 1--14, 2021{\natexlab{b}}.

\bibitem[Chen et~al.(2024{\natexlab{a}})Chen, Huang, Liu, Shen, Zhao, and Zhao]{chen2024anydoor}
Xi~Chen, Lianghua Huang, Yu~Liu, Yujun Shen, Deli Zhao, and Hengshuang Zhao.
\newblock Anydoor: Zero-shot object-level image customization.
\newblock In \emph{Proceedings of the IEEE/CVF Conference on Computer Vision and Pattern Recognition}, pp.\  6593--6602, 2024{\natexlab{a}}.

\bibitem[Chen et~al.(2024{\natexlab{b}})Chen, Mihajlovic, Wang, Prokudin, and Tang]{chen2024morphable}
Xiyi Chen, Marko Mihajlovic, Shaofei Wang, Sergey Prokudin, and Siyu Tang.
\newblock Morphable diffusion: 3d-consistent diffusion for single-image avatar creation.
\newblock In \emph{IEEE Conference on Computer Vision and Pattern Recognition (CVPR)}, 2024{\natexlab{b}}.

\bibitem[Chen et~al.(2023)Chen, Yan, Wang, and Zheng]{chen2023iot}
Yi~Chen, Yunfeng Yan, Xianbo Wang, and Yi~Zheng.
\newblock Iot-enabled few-shot image generation for power scene defect detection based on self-attention and global--local fusion.
\newblock \emph{Sensors}, 23\penalty0 (14):\penalty0 6531, 2023.

\bibitem[Cheng et~al.(2024)Cheng, Liu, Liu, Xu, and Liu]{cheng2024frequency}
Kan Cheng, Haidong Liu, Jiayu Liu, Bo~Xu, and Xinyue Liu.
\newblock Frequency-auxiliary one-shot domain adaptation of generative adversarial networks.
\newblock \emph{Electronics}, 13\penalty0 (13):\penalty0 2643, 2024.

\bibitem[Choi et~al.(2020)Choi, Uh, Yoo, and Ha]{choi2020starganv2}
Yunjey Choi, Youngjung Uh, Jaejun Yoo, and Jung-Woo Ha.
\newblock Stargan v2: Diverse image synthesis for multiple domains.
\newblock In \emph{Proceedings of the IEEE Conference on Computer Vision and Pattern Recognition}, 2020.

\bibitem[Chong \& Forsyth(2022)Chong and Forsyth]{chong2022jojogan}
Min~Jin Chong and David Forsyth.
\newblock Jojogan: One shot face stylization.
\newblock In \emph{Proceedings of the European Conference on Computer Vision}. Springer, 2022.

\bibitem[Chong et~al.(2024)Chong, Singh, Li, Lu, and Forsyth]{chong2024p2d}
Min~Jin Chong, Krishna~Kumar Singh, Yijun Li, Jingwan Lu, and David Forsyth.
\newblock P2d: Plug and play discriminator for accelerating gan frameworks.
\newblock In \emph{Proceedings of the IEEE/CVF Winter Conference on Applications of Computer Vision}, pp.\  5422--5431, 2024.

\bibitem[Clou{\^a}tre \& Demers(2019)Clou{\^a}tre and Demers]{clouatre2019figr}
Louis Clou{\^a}tre and Marc Demers.
\newblock Figr: Few-shot image generation with reptile.
\newblock \emph{arXiv preprint arXiv:1901.02199}, 2019.

\bibitem[Cong et~al.(2020)Cong, Zhao, Li, Wang, and Carin]{cong2020ganmemory}
Yulai Cong, Miaoyun Zhao, Jianqiao Li, Sijia Wang, and Lawrence Carin.
\newblock Gan memory with no forgetting.
\newblock In \emph{Advances in Neural Information Processing Systems}, 2020.

\bibitem[Couairon et~al.(2023)Couairon, Careil, Cord, Lathuiliere, and Verbeek]{couairon2023zero}
Guillaume Couairon, Marlene Careil, Matthieu Cord, St{\'e}phane Lathuiliere, and Jakob Verbeek.
\newblock Zero-shot spatial layout conditioning for text-to-image diffusion models.
\newblock In \emph{Proceedings of the IEEE/CVF International Conference on Computer Vision}, pp.\  2174--2183, 2023.

\bibitem[Cui et~al.(2022)Cui, Huang, Luo, Zhang, Zhan, and Lu]{cui2022genco}
Kaiwen Cui, Jiaxing Huang, Zhipeng Luo, Gongjie Zhang, Fangneng Zhan, and Shijian Lu.
\newblock Genco: Generative co-training for generative adversarial networks with limited data.
\newblock In \emph{Proceedings of the AAAI Conference on Artificial Intelligence}, 2022.

\bibitem[Cui et~al.(2023)Cui, Yu, Zhan, Liao, Lu, and Xing]{cui2023kddlgan}
Kaiwen Cui, Yingchen Yu, Fangneng Zhan, Shengcai Liao, Shijian Lu, and Eric Xing.
\newblock Kd-dlgan: Data limited image generation via knowledge distillation.
\newblock In \emph{Proceedings of the IEEE/CVF Conference on Computer Vision and Pattern Recognition}, 2023.

\bibitem[Cui et~al.(2024)Cui, Guo, An, Deng, Zhao, Wei, and Feng]{cui2024idadapter}
Siying Cui, Jia Guo, Xiang An, Jiankang Deng, Yongle Zhao, Xinyu Wei, and Ziyong Feng.
\newblock Idadapter: Learning mixed features for tuning-free personalization of text-to-image models.
\newblock In \emph{Proceedings of the IEEE/CVF Conference on Computer Vision and Pattern Recognition}, pp.\  950--959, 2024.

\bibitem[Dai et~al.(2022)Dai, Hang, and Guo]{dai2021implicit}
Mengyu Dai, Haibin Hang, and Xiaoyang Guo.
\newblock Adaptive feature interpolation for low-shot image generation.
\newblock In \emph{Proceedings of the European Conference on Computer Vision}, 2022.

\bibitem[De~Souza et~al.(2023)De~Souza, Marques, Batagelo, and Gois]{de2023review}
Vinicius Luis~Trevisan De~Souza, Bruno Augusto~Dorta Marques, Harlen~Costa Batagelo, and Joao~Paulo Gois.
\newblock A review on generative adversarial networks for image generation.
\newblock \emph{Computers \& Graphics}, 114:\penalty0 13--25, 2023.

\bibitem[Deng et~al.(2009)Deng, Dong, Socher, Li, Li, and Fei-Fei]{deng2009imagenet}
Jia Deng, Wei Dong, Richard Socher, Li-Jia Li, Kai Li, and Li~Fei-Fei.
\newblock Imagenet: A large-scale hierarchical image database.
\newblock In \emph{Proceedings of the IEEE/CVF Conference on Computer Vision and Pattern Recognition}, 2009.

\bibitem[Devlin et~al.(2019)Devlin, Chang, Lee, and Toutanova]{devlin2019bert}
Jacob Devlin, Ming-Wei Chang, Kenton Lee, and Kristina Toutanova.
\newblock Bert: Pre-training of deep bidirectional transformers for language understanding.
\newblock In \emph{Proceedings of North American Chapter of the Association for Computational Linguistics}, 2019.

\bibitem[Dhariwal \& Nichol(2021)Dhariwal and Nichol]{dhariwal2021diffusionvsGAN}
Prafulla Dhariwal and Alexander Nichol.
\newblock Diffusion models beat gans on image synthesis.
\newblock In \emph{Advances in Neural Information Processing Systems}, 2021.

\bibitem[Ding et~al.(2022)Ding, Han, Wang, Wu, Jin, Tu, and Huang]{ding2022age}
Guanqi Ding, Xinzhe Han, Shuhui Wang, Shuzhe Wu, Xin Jin, Dandan Tu, and Qingming Huang.
\newblock Attribute group editing for reliable few-shot image generation.
\newblock In \emph{Proceedings of the IEEE/CVF Conference on Computer Vision and Pattern Recognition}, 2022.

\bibitem[Ding et~al.(2023)Ding, Han, Wang, Jin, Tu, and Huang]{ding2023sage}
Guanqi Ding, Xinzhe Han, Shuhui Wang, Xin Jin, Dandan Tu, and Qingming Huang.
\newblock Stable attribute group editing for reliable few-shot image generation.
\newblock \emph{arXiv preprint arXiv:2302.00179}, 2023.

\bibitem[Doloriel \& Cheung(2024)Doloriel and Cheung]{doloriel2024frequency}
Chandler~Timm Doloriel and Ngai-Man Cheung.
\newblock Frequency masking for universal deepfake detection.
\newblock In \emph{ICASSP 2024 - 2024 IEEE International Conference on Acoustics, Speech and Signal Processing (ICASSP)}, pp.\  13466--13470, 2024.
\newblock \doi{10.1109/ICASSP48485.2024.10446290}.

\bibitem[Duan et~al.(2023{\natexlab{a}})Duan, Hong, Niu, and Zhang]{duan2023few}
Yuxuan Duan, Yan Hong, Li~Niu, and Liqing Zhang.
\newblock Few-shot defect image generation via defect-aware feature manipulation.
\newblock In \emph{Proceedings of the AAAI conference on artificial intelligence}, volume~37, pp.\  571--578, 2023{\natexlab{a}}.

\bibitem[Duan et~al.(2023{\natexlab{b}})Duan, Niu, Hong, and Zhang]{duan2023weditgan}
Yuxuan Duan, Li~Niu, Yan Hong, and Liqing Zhang.
\newblock Weditgan: Few-shot image generation via latent space relocation.
\newblock \emph{arXiv preprint arXiv:2305.06671}, 2023{\natexlab{b}}.

\bibitem[Duan et~al.(2024)Duan, Hong, Zhang, Lan, Zhu, Wang, Zhang, Niu, and Zhang]{duan2024domaingallery}
Yuxuan Duan, Yan Hong, Bo~Zhang, Jun Lan, Huijia Zhu, Weiqiang Wang, Jianfu Zhang, Li~Niu, and Liqing Zhang.
\newblock Domaingallery: Few-shot domain-driven image generation by attribute-centric finetuning.
\newblock \emph{arXiv preprint arXiv:2411.04571}, 2024.

\bibitem[Durall et~al.(2020)Durall, Keuper, and Keuper]{durall2020watchupconvolution}
Ricard Durall, Margret Keuper, and Janis Keuper.
\newblock Watch your up-convolution: Cnn based generative deep neural networks are failing to reproduce spectral distributions.
\newblock In \emph{Proceedings of the IEEE/CVF Conference on Computer Vision and Pattern Recognition}, 2020.

\bibitem[Esser et~al.(2021)Esser, Rombach, and Ommer]{esser2021vqgan}
Patrick Esser, Robin Rombach, and Bjorn Ommer.
\newblock Taming transformers for high-resolution image synthesis.
\newblock In \emph{Proceedings of the IEEE/CVF Conference on Computer Vision and Pattern Recognition}, 2021.

\bibitem[Esser et~al.(2024)Esser, Kulal, Blattmann, Entezari, M{\"u}ller, Saini, Levi, Lorenz, Sauer, Boesel, et~al.]{esser2024scaling}
Patrick Esser, Sumith Kulal, Andreas Blattmann, Rahim Entezari, Jonas M{\"u}ller, Harry Saini, Yam Levi, Dominik Lorenz, Axel Sauer, Frederic Boesel, et~al.
\newblock Scaling rectified flow transformers for high-resolution image synthesis.
\newblock In \emph{Forty-first international conference on machine learning}, 2024.

\bibitem[Everaert et~al.(2023)Everaert, Bocchio, Arpa, S{\"u}sstrunk, and Achanta]{everaert2023dis}
Martin~Nicolas Everaert, Marco Bocchio, Sami Arpa, Sabine S{\"u}sstrunk, and Radhakrishna Achanta.
\newblock Diffusion in style.
\newblock In \emph{Proceedings of the IEEE/CVF International Conference on Computer Vision}, pp.\  2251--2261, 2023.

\bibitem[Fang et~al.(2022)Fang, Sun, and Schwing]{fang2022diggan}
Tiantian Fang, Ruoyu Sun, and Alex Schwing.
\newblock Diggan: Discriminator gradient gap regularization for gan training with limited data.
\newblock In \emph{Advances in Neural Information Processing Systems}, 2022.

\bibitem[Finn et~al.(2017)Finn, Abbeel, and Levine]{finn2017maml}
Chelsea Finn, Pieter Abbeel, and Sergey Levine.
\newblock Model-agnostic meta-learning for fast adaptation of deep network.
\newblock In \emph{Proceedings of the International Conference on Machine Learning}, 2017.

\bibitem[Frankle \& Carbin(2019)Frankle and Carbin]{frankle2019lotteryticket}
Jonathan Frankle and Michael Carbin.
\newblock The lottery ticket hypothesis: Finding sparse, trainable neural networks.
\newblock In \emph{International Conference on Learning Representations}, 2019.

\bibitem[Fu et~al.(2023)Fu, Tamir, Sundaram, Chai, Zhang, Dekel, and Isola]{fu2023dreamsim}
Stephanie Fu, Netanel Tamir, Shobhita Sundaram, Lucy Chai, Richard Zhang, Tali Dekel, and Phillip Isola.
\newblock Dreamsim: Learning new dimensions of human visual similarity using synthetic data.
\newblock \emph{arXiv:2306.09344}, 2023.

\bibitem[Gal et~al.(2022{\natexlab{a}})Gal, Alaluf, Atzmon, Patashnik, Bermano, Chechik, and Cohen-Or]{gal2022textualinversion}
Rinon Gal, Yuval Alaluf, Yuval Atzmon, Or~Patashnik, Amit~H Bermano, Gal Chechik, and Daniel Cohen-Or.
\newblock An image is worth one word: Personalizing text-to-image generation using textual inversion.
\newblock \emph{arXiv preprint arXiv:2208.01618}, 2022{\natexlab{a}}.

\bibitem[Gal et~al.(2022{\natexlab{b}})Gal, Patashnik, Maron, Bermano, Chechik, and Cohen-Or]{gal2022stylegannada}
Rinon Gal, Or~Patashnik, Haggai Maron, Amit~H Bermano, Gal Chechik, and Daniel Cohen-Or.
\newblock Stylegan-nada: Clip-guided domain adaptation of image generators.
\newblock \emph{ACM Transactions on Graphics}, 41\penalty0 (4):\penalty0 1--13, 2022{\natexlab{b}}.

\bibitem[Gal et~al.(2023)Gal, Arar, Atzmon, Bermano, Chechik, and Cohen-Or]{gal2023e4t}
Rinon Gal, Moab Arar, Yuval Atzmon, Amit~H Bermano, Gal Chechik, and Daniel Cohen-Or.
\newblock Encoder-based domain tuning for fast personalization of text-to-image models.
\newblock \emph{ACM Transactions on Graphics (TOG)}, 42\penalty0 (4):\penalty0 1--13, 2023.

\bibitem[Gat et~al.(2024)Gat, Remez, Shaul, Kreuk, Chen, Synnaeve, Adi, and Lipman]{gat2024discrete}
Itai Gat, Tal Remez, Neta Shaul, Felix Kreuk, Ricky~TQ Chen, Gabriel Synnaeve, Yossi Adi, and Yaron Lipman.
\newblock Discrete flow matching.
\newblock \emph{Advances in Neural Information Processing Systems}, 37:\penalty0 133345--133385, 2024.

\bibitem[Gharoun et~al.(2024)Gharoun, Momenifar, Chen, and Gandomi]{gharoun2024meta}
Hassan Gharoun, Fereshteh Momenifar, Fang Chen, and Amir~H Gandomi.
\newblock Meta-learning approaches for few-shot learning: A survey of recent advances.
\newblock \emph{ACM Computing Surveys}, 56\penalty0 (12):\penalty0 1--41, 2024.

\bibitem[Giannone \& Winther(2022)Giannone and Winther]{pmlr-v162-giannone22a}
Giorgio Giannone and Ole Winther.
\newblock Scha-vae: Hierarchical context aggregation for few-shot generation.
\newblock In \emph{Proceedings of the International Conference on Machine Learning}, 2022.

\bibitem[Giannone et~al.(2022)Giannone, Nielsen, and Winther]{giannone2022fsdm}
Giorgio Giannone, Didrik Nielsen, and Ole Winther.
\newblock Few-shot diffusion models.
\newblock \emph{arXiv preprint arXiv:2205.15463}, 2022.

\bibitem[Gonzales \& Wintz(1987)Gonzales and Wintz]{gonzales1987digital}
Rafael~C Gonzales and Paul Wintz.
\newblock \emph{Digital image processing}.
\newblock Addison-Wesley Longman Publishing Co., Inc., 1987.

\bibitem[Goodfellow et~al.(2014)Goodfellow, Pouget-Abadie, Mirza, Xu, Warde-Farley, Ozair, Courville, and Bengio]{goodfellow2014GANs}
Ian Goodfellow, Jean Pouget-Abadie, Mehdi Mirza, Bing Xu, David Warde-Farley, Sherjil Ozair, Aaron Courville, and Yoshua Bengio.
\newblock Generative adversarial nets.
\newblock In \emph{Advances in Neural Information Processing Systems}, 2014.

\bibitem[Gou et~al.(2023)Gou, Li, Lv, Zhang, Xing, and He]{gou2023csr}
Yao Gou, Min Li, Yilong Lv, Yusen Zhang, Yuhang Xing, and Yujie He.
\newblock Rethinking cross-domain semantic relation for few-shot image generation.
\newblock \emph{Applied Intelligence}, pp.\  1--14, 2023.

\bibitem[Gou et~al.(2024)Gou, Li, Zhang, He, and He]{gou2024few}
Yao Gou, Min Li, Yusen Zhang, Zhuzhen He, and Yujie He.
\newblock Few-shot image generation with reverse contrastive learning.
\newblock \emph{Neural Networks}, 169:\penalty0 154--164, 2024.

\bibitem[Gouk et~al.(2021)Gouk, Frank, Pfahringer, and Cree]{gouk2021regularisation}
Henry Gouk, Eibe Frank, Bernhard Pfahringer, and Michael~J Cree.
\newblock Regularisation of neural networks by enforcing lipschitz continuity.
\newblock \emph{Machine Learning}, 110:\penalty0 393--416, 2021.

\bibitem[Grigoryev et~al.(2022)Grigoryev, Voynov, and Babenko]{grigoryev2022when}
Timofey Grigoryev, Andrey Voynov, and Artem Babenko.
\newblock When, why, and which pretrained gans are useful?
\newblock In \emph{International Conference on Learning Representations}, 2022.

\bibitem[Gu et~al.(2023)Gu, Wang, Wu, Shi, Chen, Fan, Xiao, Zhao, Chang, Wu, et~al.]{gu2023mixofshow}
Yuchao Gu, Xintao Wang, Jay~Zhangjie Wu, Yujun Shi, Yunpeng Chen, Zihan Fan, Wuyou Xiao, Rui Zhao, Shuning Chang, Weijia Wu, et~al.
\newblock Mix-of-show: Decentralized low-rank adaptation for multi-concept customization of diffusion models.
\newblock \emph{Advances in Neural Information Processing Systems}, 36:\penalty0 15890--15902, 2023.

\bibitem[Gu et~al.(2021)Gu, Li, Huo, Wang, and Gao]{gu2021lofgan}
Zheng Gu, Wenbin Li, Jing Huo, Lei Wang, and Yang Gao.
\newblock Lofgan: Fusing local representations for few-shot image generation.
\newblock In \emph{Proceedings of the IEEE/CVF International Conference on Computer Vision}, 2021.

\bibitem[Guan et~al.(2025)Guan, Ge, Tai, Yang, Li, and You]{guan2025hybridbooth}
Shanyan Guan, Yanhao Ge, Ying Tai, Jian Yang, Wei Li, and Mingyu You.
\newblock Hybridbooth: Hybrid prompt inversion for efficient subject-driven generation.
\newblock In \emph{European Conference on Computer Vision}, pp.\  403--419. Springer, 2025.

\bibitem[Gulrajani et~al.(2017)Gulrajani, Ahmed, Arjovsky, Dumoulin, and Courville]{gulrajani2017improved}
Ishaan Gulrajani, Faruk Ahmed, Martin Arjovsky, Vincent Dumoulin, and Aaron~C Courville.
\newblock Improved training of wasserstein gans.
\newblock In \emph{Advances in Neural Information Processing Systems}, 2017.

\bibitem[Guo et~al.(2023)Guo, Wang, Wu, Zhang, Wang, Xu, Shi, Huang, and Song]{guo2023ipl}
Jiayi Guo, Chaofei Wang, You Wu, Eric Zhang, Kai Wang, Xingqian Xu, Humphrey Shi, Gao Huang, and Shiji Song.
\newblock Zero-shot generative model adaptation via image-specific prompt learning.
\newblock In \emph{Proceedings of the IEEE/CVF Conference on Computer Vision and Pattern Recognition}, 2023.

\bibitem[Guo et~al.(2025)Guo, Zhang, Tong, Zhao, Gao, Li, and Heng]{guo2025got}
Ziyu Guo, Renrui Zhang, Chengzhuo Tong, Zhizheng Zhao, Peng Gao, Hongsheng Li, and Pheng-Ann Heng.
\newblock Can we generate images with cot? let's verify and reinforce image generation step by step.
\newblock In \emph{IEEE/CVF Conference on Computer Vision and Pattern Recognition}, 2025.

\bibitem[Gupta et~al.(2024)Gupta, Hayat, Dhall, and Do]{gupta2024conditional}
Parul Gupta, Munawar Hayat, Abhinav Dhall, and Thanh-Toan Do.
\newblock Conditional distribution modelling for few-shot image synthesis with diffusion models.
\newblock In \emph{Proceedings of the Asian Conference on Computer Vision}, pp.\  818--834, 2024.

\bibitem[Han et~al.(2024{\natexlab{a}})Han, Kokkinos, and Torr]{han2024vfusion3d}
Junlin Han, Filippos Kokkinos, and Philip Torr.
\newblock Vfusion3d: Learning scalable 3d generative models from video diffusion models.
\newblock \emph{European Conference on Computer Vision (ECCV)}, 2024{\natexlab{a}}.

\bibitem[Han et~al.(2024{\natexlab{b}})Han, Bao, and Bai]{han2024latentsdm}
Xianjun Han, Taoli Bao, and Can Bai.
\newblock A faster single-image denoising diffusion model: Emphasizing the role of the latent image code.
\newblock \emph{Available at SSRN 5000004}, 2024{\natexlab{b}}.

\bibitem[He et~al.(2016)He, Zhang, Ren, and Sun]{he2016resnet}
Kaiming He, Xiangyu Zhang, Shaoqing Ren, and Jian Sun.
\newblock Deep residual learning for image recognition.
\newblock In \emph{Proceedings of the IEEE/CVF Conference on Computer Vision and Pattern Recognition}, 2016.

\bibitem[He et~al.(2020)He, Fan, Wu, Xie, and Girshick]{he2020momentum}
Kaiming He, Haoqi Fan, Yuxin Wu, Saining Xie, and Ross Girshick.
\newblock Momentum contrast for unsupervised visual representation learning.
\newblock In \emph{Proceedings of the IEEE/CVF Conference on Computer Vision and Pattern Recognition}, 2020.

\bibitem[He et~al.(2024)He, Yang, Liu, and Lin]{he2024few}
Xiaosheng He, Fan Yang, Fayao Liu, and Guosheng Lin.
\newblock Few-shot image generation via style adaptation and content preservation.
\newblock \emph{IEEE Transactions on Neural Networks and Learning Systems}, 2024.

\bibitem[He \& Fu(2021)He and Fu]{he2021recurrent}
Xiaoyu He and Zhenyong Fu.
\newblock Recurrent singan: Towards scale-agnostic single image gans.
\newblock In \emph{Proceedings of the 2021 5th international conference on electronic information technology and computer engineering}, pp.\  361--366, 2021.

\bibitem[Heusel et~al.(2017)Heusel, Ramsauer, Unterthiner, Nessler, and Hochreiter]{heusel2017twotimescale}
Martin Heusel, Hubert Ramsauer, Thomas Unterthiner, Bernhard Nessler, and Sepp Hochreiter.
\newblock Gans trained by a two time-scale update rule converge to a local nash equilibrium.
\newblock In \emph{Advances in Neural Information Processing Systems}, 2017.

\bibitem[Hinton et~al.(2015)Hinton, Vinyals, and Dean]{hinton-distill}
Geoffrey Hinton, Oriol Vinyals, and Jeffrey Dean.
\newblock Distilling the knowledge in a neural network.
\newblock In \emph{NeurIPS Deep Learning and Representation Learning Workshop}, 2015.

\bibitem[Hinz et~al.(2021)Hinz, Fisher, Wang, and Wermter]{hinz2021consingan}
Tobias Hinz, Matthew Fisher, Oliver Wang, and Stefan Wermter.
\newblock Improved techniques for training single-image gans.
\newblock In \emph{Proceedings of the IEEE/CVF Winter Conference on Applications of Computer Vision}, 2021.

\bibitem[Hiruta et~al.(2022)Hiruta, Saito, Hatakeyama, Hashimoto, and Kurihara]{hiruta2022cfgan}
Komei Hiruta, Ryusuke Saito, Taro Hatakeyama, Atsushi Hashimoto, and Satoshi Kurihara.
\newblock Conditional gan for small datasets.
\newblock In \emph{2022 IEEE International Symposium on Multimedia (ISM)}, pp.\  278--281. IEEE, 2022.

\bibitem[Ho et~al.(2019)Ho, Chen, Srinivas, Duan, and Abbeel]{ho2019flow++}
Jonathan Ho, Xi~Chen, Aravind Srinivas, Yan Duan, and Pieter Abbeel.
\newblock Flow++: Improving flow-based generative models with variational dequantization and architecture design.
\newblock In \emph{Proceedings of the International Conference on Machine Learning}, 2019.

\bibitem[Ho et~al.(2020)Ho, Jain, and Abbeel]{ho2020denoising}
Jonathan Ho, Ajay Jain, and Pieter Abbeel.
\newblock Denoising diffusion probabilistic models.
\newblock In \emph{Advances in Neural Information Processing Systems}, 2020.

\bibitem[Hong et~al.(2020{\natexlab{a}})Hong, Niu, Zhang, and Zhang]{hong2020matchinggan}
Yan Hong, Li~Niu, Jianfu Zhang, and Liqing Zhang.
\newblock Matchinggan: Matching-based few-shot image generation.
\newblock In \emph{IEEE International Conference on Multimedia and Expo}, 2020{\natexlab{a}}.

\bibitem[Hong et~al.(2020{\natexlab{b}})Hong, Niu, Zhang, Zhao, Fu, and Zhang]{hong2020f2gan}
Yan Hong, Li~Niu, Jianfu Zhang, Weijie Zhao, Chen Fu, and Liqing Zhang.
\newblock F2gan: Fusing-and-filling gan for few-shot image generation.
\newblock In \emph{Proceedings of the 28th ACM international conference on multimedia}, pp.\  2535--2543, 2020{\natexlab{b}}.

\bibitem[Hong et~al.(2022{\natexlab{a}})Hong, Niu, Zhang, and Zhang]{hong2022deltagan}
Yan Hong, Li~Niu, Jianfu Zhang, and Liqing Zhang.
\newblock Deltagan: Towards diverse few-shot image generation with sample-specific delta.
\newblock In \emph{Proceedings of the European Conference on Computer Vision}, 2022{\natexlab{a}}.

\bibitem[Hong et~al.(2022{\natexlab{b}})Hong, Niu, Zhang, and Zhang]{hong2022disco}
Yan Hong, Li~Niu, Jianfu Zhang, and Liqing Zhang.
\newblock Few-shot image generation using discrete content representation.
\newblock In \emph{Proceedings of the ACM International Conference on Multimedia}, 2022{\natexlab{b}}.

\bibitem[Hong et~al.(2025)Hong, Duan, Zhang, Chen, Lan, Zhu, Wang, and Zhang]{hong2025ComFusion}
Yan Hong, Yuxuan Duan, Bo~Zhang, Haoxing Chen, Jun Lan, Huijia Zhu, Weiqiang Wang, and Jianfu Zhang.
\newblock Comfusion: Enhancing personalized generation by instance-scene compositing and fusion.
\newblock In \emph{European Conference on Computer Vision}, pp.\  1--18. Springer, 2025.

\bibitem[Hospedales et~al.(2021)Hospedales, Antoniou, Micaelli, and Storkey]{hospedales2021meta}
Timothy Hospedales, Antreas Antoniou, Paul Micaelli, and Amos Storkey.
\newblock Meta-learning in neural networks: A survey.
\newblock \emph{IEEE transactions on pattern analysis and machine intelligence}, 44\penalty0 (9):\penalty0 5149--5169, 2021.

\bibitem[Hou(2023)]{hou2023regularizing}
Liang Hou.
\newblock Regularizing label-augmented generative adversarial networks under limited data.
\newblock \emph{IEEE Access}, 11:\penalty0 28966--28976, 2023.

\bibitem[Hou et~al.(2021)Hou, Shen, Cao, and Cheng]{hou2021labelaugmentation}
Liang Hou, Huawei Shen, Qi~Cao, and Xueqi Cheng.
\newblock Self-supervised gans with label augmentation.
\newblock In \emph{Advances in Neural Information Processing Systems}, 2021.

\bibitem[Hou et~al.(2024)Hou, Cao, Yuan, Zhao, Ma, Pan, Wan, Wang, Shen, and Cheng]{hou2024augselfgan}
Liang Hou, Qi~Cao, Yige Yuan, Songtao Zhao, Chongyang Ma, Siyuan Pan, Pengfei Wan, Zhongyuan Wang, Huawei Shen, and Xueqi Cheng.
\newblock Augmentation-aware self-supervision for data-efficient gan training.
\newblock \emph{Advances in Neural Information Processing Systems}, 36, 2024.

\bibitem[Hou et~al.(2022{\natexlab{a}})Hou, Liu, Wan, and You]{hou2022exploitingkd}
Xingzhong Hou, Boxiao Liu, Fang Wan, and Haihang You.
\newblock Exploiting knowledge distillation for few-shot image generation, 2022{\natexlab{a}}.
\newblock URL \url{https://openreview.net/forum?id=vsEi1UMa7TC}.

\bibitem[Hou et~al.(2022{\natexlab{b}})Hou, Liu, Zhang, Shi, Jiang, and You]{hou_dynamic}
Xingzhong Hou, Boxiao Liu, Shuai Zhang, Lulin Shi, Zite Jiang, and Haihang You.
\newblock Dynamic weighted semantic correspondence for few-shot image generative adaptation.
\newblock In \emph{Proceedings of the ACM International Conference on Multimedia}, 2022{\natexlab{b}}.

\bibitem[Hu et~al.(2024{\natexlab{a}})Hu, Liu, and Wu]{hu2024DOGAN}
Cong Hu, Si-hao Liu, and Xiao-jun Wu.
\newblock Lightweight dual-path octave generative adversarial networks for few-shot image generation.
\newblock \emph{Multimedia Systems}, 30\penalty0 (5):\penalty0 278, 2024{\natexlab{a}}.

\bibitem[Hu et~al.(2018)Hu, Shen, and Sun]{hu2018squeeze}
Jie Hu, Li~Shen, and Gang Sun.
\newblock Squeeze-and-excitation networks.
\newblock In \emph{Proceedings of the IEEE/CVF Conference on Computer Vision and Pattern Recognition}, 2018.

\bibitem[Hu et~al.(2023{\natexlab{a}})Hu, Zhang, Liu, Yi, Kou, Zhu, Chen, Wang, Wang, and Ma]{hu2023phasic}
Teng Hu, Jiangning Zhang, Liang Liu, Ran Yi, Siqi Kou, Haokun Zhu, Xu~Chen, Yabiao Wang, Chengjie Wang, and Lizhuang Ma.
\newblock Phasic content fusing diffusion model with directional distribution consistency for few-shot model adaption.
\newblock In \emph{Proceedings of the IEEE/CVF International Conference on Computer Vision}, pp.\  2406--2415, 2023{\natexlab{a}}.

\bibitem[Hu et~al.(2024{\natexlab{b}})Hu, Zhang, Yi, Du, Chen, Liu, Wang, and Wang]{hu2024anomalydiffusion}
Teng Hu, Jiangning Zhang, Ran Yi, Yuzhen Du, Xu~Chen, Liang Liu, Yabiao Wang, and Chengjie Wang.
\newblock Anomalydiffusion: Few-shot anomaly image generation with diffusion model.
\newblock In \emph{Proceedings of the AAAI Conference on Artificial Intelligence}, volume~38, pp.\  8526--8534, 2024{\natexlab{b}}.

\bibitem[Hu et~al.(2023{\natexlab{b}})Hu, Jin, Liang, Liu, Luo, Li, and Peng]{hu2023diffusion}
Xinrong Hu, Yuxin Jin, Jinxing Liang, Junping Liu, Ruiqi Luo, Min Li, and Tao Peng.
\newblock Diffusion model for image generation-a survey.
\newblock In \emph{2023 2nd International Conference on Artificial Intelligence, Human-Computer Interaction and Robotics (AIHCIR)}, pp.\  416--424. IEEE, 2023{\natexlab{b}}.

\bibitem[Huang et~al.(2021)Huang, Wang, Rabusseau, and Makhzani]{huang2021few}
Andy Huang, Kuan-Chieh Wang, Guillaume Rabusseau, and Alireza Makhzani.
\newblock Few shot image generation via implicit autoencoding of support sets.
\newblock In \emph{Fifth Workshop on Meta-Learning at the Conference on Neural Information Processing Systems}, 2021.

\bibitem[Huang et~al.(2022)Huang, Cui, Guan, Xiao, Zhan, Lu, Liao, and Xing]{huang2022maskedgan}
Jiaxing Huang, Kaiwen Cui, Dayan Guan, Aoran Xiao, Fangneng Zhan, Shijian Lu, Shengcai Liao, and Eric Xing.
\newblock Masked generative adversarial networks are data-efficient generation learners.
\newblock In \emph{Advances in Neural Information Processing Systems}, 2022.

\bibitem[Huang et~al.(2024{\natexlab{a}})Huang, Mao, Liu, He, and Zhang]{huang2024realcustom}
Mengqi Huang, Zhendong Mao, Mingcong Liu, Qian He, and Yongdong Zhang.
\newblock Realcustom: Narrowing real text word for real-time open-domain text-to-image customization.
\newblock In \emph{Proceedings of the IEEE/CVF Conference on Computer Vision and Pattern Recognition}, pp.\  7476--7485, 2024{\natexlab{a}}.

\bibitem[Huang et~al.(2024{\natexlab{b}})Huang, Gokaslan, Kuleshov, and Tompkin]{huang2024r3gan}
Nick Huang, Aaron Gokaslan, Volodymyr Kuleshov, and James Tompkin.
\newblock The {GAN} is dead; long live the {GAN}! a modern {GAN} baseline.
\newblock In \emph{The Thirty-eighth Annual Conference on Neural Information Processing Systems}, 2024{\natexlab{b}}.

\bibitem[Hyung et~al.(2024)Hyung, Shin, and Choo]{hyung2024magicapture}
Junha Hyung, Jaeyo Shin, and Jaegul Choo.
\newblock Magicapture: High-resolution multi-concept portrait customization.
\newblock In \emph{Proceedings of the AAAI Conference on Artificial Intelligence}, volume~38, pp.\  2445--2453, 2024.

\bibitem[Israr et~al.(2024)Israr, Saeed, and Zhao]{israr2024few}
Syed~Muhammad Israr, Rehan Saeed, and Feng Zhao.
\newblock Few-shot adaptation of gans using self-supervised consistency regularization.
\newblock \emph{Knowledge-Based Systems}, 302:\penalty0 112256, 2024.

\bibitem[Jabbar et~al.(2021)Jabbar, Li, and Omar]{jabbar2021survey}
Abdul Jabbar, Xi~Li, and Bourahla Omar.
\newblock A survey on generative adversarial networks: Variants, applications, and training.
\newblock \emph{ACM Computing Surveys (CSUR)}, 54\penalty0 (8):\penalty0 1--49, 2021.

\bibitem[Jamali(2025)]{jamali-2025}
Lily Jamali.
\newblock {AI chatbot to be embedded in Google search}.
\newblock 2025.
\newblock URL \url{https://www.bbc.com/news/articles/cpw77qwd117o}.

\bibitem[Jeon et~al.(2023)Jeon, Liu, Lee, Hong, Fu, and Byun]{jeon2023svdm}
Seogkyu Jeon, Bei Liu, Pilhyeon Lee, Kibeom Hong, Jianlong Fu, and Hyeran Byun.
\newblock Improving diversity in zero-shot gan adaptation with semantic variations.
\newblock In \emph{Proceedings of the IEEE/CVF International Conference on Computer Vision}, pp.\  7258--7267, 2023.

\bibitem[Jeong et~al.(2023)Jeong, Kwon, and Ye]{jeong2023cii}
Hyeonho Jeong, Gihyun Kwon, and Jong~Chul Ye.
\newblock Zero-shot generation of coherent storybook from plain text story using diffusion models.
\newblock \emph{arXiv preprint arXiv:2302.03900}, 2023.

\bibitem[Jiang et~al.(2021)Jiang, Dai, Wu, and Loy]{jiang2021deceive}
Liming Jiang, Bo~Dai, Wayne Wu, and Chen~Change Loy.
\newblock Deceive d: Adaptive pseudo augmentation for gan training with limited data.
\newblock In \emph{Advances in Neural Information Processing Systems}, 2021.

\bibitem[Jiang et~al.(2023{\natexlab{a}})Jiang, Liu, Li, and Xu]{jiang2023personalized}
Shuyi Jiang, Daochang Liu, Dingquan Li, and Chang Xu.
\newblock Personalized image generation for color vision deficiency population.
\newblock In \emph{Proceedings of the International Conference on Computer Vision (ICCV)}, 2023{\natexlab{a}}.

\bibitem[Jiang et~al.(2023{\natexlab{b}})Jiang, Yan, Zhang, Liu, and Sun]{jiang2023tcgan}
Yunliang Jiang, Lili Yan, Xiongtao Zhang, Yong Liu, and Danfeng Sun.
\newblock Tcgan: Semantic-aware and structure-preserved gans with individual vision transformer for fast arbitrary one-shot image generation.
\newblock \emph{arXiv preprint arXiv:2302.08047}, 2023{\natexlab{b}}.

\bibitem[Jin et~al.(2025)Jin, Shen, Fu, and Yang]{jin2025CGR}
Jian Jin, Yang Shen, Zhenyong Fu, and Jian Yang.
\newblock Customized generation reimagined: Fidelity and editability harmonized.
\newblock In \emph{European Conference on Computer Vision}, pp.\  410--426. Springer, 2025.

\bibitem[Kang et~al.(2023)Kang, Zhu, Zhang, Park, Shechtman, Paris, and Park]{kang2023scaling}
Minguk Kang, Jun-Yan Zhu, Richard Zhang, Jaesik Park, Eli Shechtman, Sylvain Paris, and Taesung Park.
\newblock Scaling up gans for text-to-image synthesis.
\newblock In \emph{Proceedings of the IEEE/CVF conference on computer vision and pattern recognition}, pp.\  10124--10134, 2023.

\bibitem[Karras et~al.(2019)Karras, Laine, and Aila]{karras2019style}
Tero Karras, Samuli Laine, and Timo Aila.
\newblock A style-based generator architecture for generative adversarial networks.
\newblock In \emph{Proceedings of the IEEE/CVF Conference on Computer Vision and Pattern Recognition}, 2019.

\bibitem[Karras et~al.(2020{\natexlab{a}})Karras, Aittala, Hellsten, Laine, Lehtinen, and Aila]{karras2020ada}
Tero Karras, Miika Aittala, Janne Hellsten, Samuli Laine, Jaakko Lehtinen, and Timo Aila.
\newblock Training generative adversarial networks with limited data.
\newblock In \emph{Advances in Neural Information Processing Systems}, 2020{\natexlab{a}}.

\bibitem[Karras et~al.(2020{\natexlab{b}})Karras, Laine, Aittala, Hellsten, Lehtinen, and Aila]{karras2020analyzing}
Tero Karras, Samuli Laine, Miika Aittala, Janne Hellsten, Jaakko Lehtinen, and Timo Aila.
\newblock Analyzing and improving the image quality of stylegan.
\newblock In \emph{Proceedings of the IEEE/CVF Conference on Computer Vision and Pattern Recognition}, 2020{\natexlab{b}}.

\bibitem[Karras et~al.(2022)Karras, Aittala, Aila, and Laine]{karras2022elucidating}
Tero Karras, Miika Aittala, Timo Aila, and Samuli Laine.
\newblock Elucidating the design space of diffusion-based generative models.
\newblock In \emph{Advances in Neural Information Processing Systems}, 2022.

\bibitem[Kato et~al.(2023{\natexlab{a}})Kato, Mikawa, and Fujisawa]{kato2023faster}
Yuichi Kato, Masahiko Mikawa, and Makoto Fujisawa.
\newblock Faster few-shot face image generation with features of specific group using pivotal tuning inversion and pca.
\newblock In \emph{2023 International Conference on Artificial Intelligence in Information and Communication (ICAIIC)}, pp.\  419--424. IEEE, 2023{\natexlab{a}}.

\bibitem[Kato et~al.(2023{\natexlab{b}})Kato, Mikawa, and Fujisawa]{yuichi2023_fewshot}
Yuichi Kato, Masahiko Mikawa, and Makoto Fujisawa.
\newblock Faster few-shot face image generation with features of specific group using pivotal tuning inversion and pca.
\newblock In \emph{International Conference on Artificial Intelligence in Information and Communication}, 2023{\natexlab{b}}.

\bibitem[Khayatkhoei \& Elgammal(2022)Khayatkhoei and Elgammal]{khayatkhoei2022ganfrequencybias}
Mahyar Khayatkhoei and Ahmed Elgammal.
\newblock Spatial frequency bias in convolutional generative adversarial networks.
\newblock In \emph{Proceedings of the AAAI Conference on Artificial Intelligence}, 2022.

\bibitem[Kim et~al.(2025)Kim, Song, Castells, and Choi]{kim2025bksdm}
Bo-Kyeong Kim, Hyoung-Kyu Song, Thibault Castells, and Shinkook Choi.
\newblock Bk-sdm: A lightweight, fast, and cheap version of stable diffusion.
\newblock In \emph{European Conference on Computer Vision}, pp.\  381--399. Springer, 2025.

\bibitem[Kim \& Chun(2023)Kim and Chun]{kim2023datid3d}
Gwanghyun Kim and Se~Young Chun.
\newblock Datid-3d: Diversity-preserved domain adaptation using text-to-image diffusion for 3d generative model.
\newblock In \emph{Proceedings of the IEEE/CVF Conference on Computer Vision and Pattern Recognition (CVPR)}, 2023.

\bibitem[Kim et~al.(2023)Kim, Jang, and Chun]{kim2023podia}
Gwanghyun Kim, Ji~Ha Jang, and Se~Young Chun.
\newblock Podia-3d: Domain adaptation of 3d generative model across large domain gap using pose-preserved text-to-image diffusion.
\newblock In \emph{Proceedings of the IEEE/CVF international conference on computer vision}, pp.\  22603--22612, 2023.

\bibitem[Kim et~al.(2022{\natexlab{a}})Kim, Choi, and Uh]{kim2022feature}
Junho Kim, Yunjey Choi, and Youngjung Uh.
\newblock Feature statistics mixing regularization for generative adversarial networks.
\newblock In \emph{Proceedings of the IEEE/CVF conference on computer vision and pattern recognition}, pp.\  11294--11303, 2022{\natexlab{a}}.

\bibitem[Kim et~al.(2022{\natexlab{b}})Kim, Kang, Kim, Baek, and Cho]{kim2022dynagan}
Seongtae Kim, Kyoungkook Kang, Geonung Kim, Seung-Hwan Baek, and Sunghyun Cho.
\newblock Dynagan: Dynamic few-shot adaptation of gans to multiple domains.
\newblock In \emph{ACM Transactions on Graphics (SIGGRAPH Asia)}, 2022{\natexlab{b}}.

\bibitem[Kingma \& Welling(2014)Kingma and Welling]{kingma2013VAE}
Diederik~P Kingma and Max Welling.
\newblock Auto-encoding variational bayes.
\newblock In \emph{International Conference on Learning Representations}, 2014.

\bibitem[Kingma et~al.(2019)Kingma, Welling, et~al.]{kingma2019introduction}
Diederik~P Kingma, Max Welling, et~al.
\newblock An introduction to variational autoencoders.
\newblock \emph{Foundations and Trends{\textregistered} in Machine Learning}, 12\penalty0 (4):\penalty0 307--392, 2019.

\bibitem[Kligvasser et~al.(2022)Kligvasser, Shaham, Alkobi, and Michaeli]{kligvasser2022blendgan}
Idan Kligvasser, Tamar~Rott Shaham, Noa Alkobi, and Tomer Michaeli.
\newblock Blendgan: Learning and blending the internal distributions of single images by spatial image-identity conditioning.
\newblock \emph{arXiv preprint arXiv:2212.01589}, 2022.

\bibitem[Kong et~al.(2022)Kong, Kim, Han, and Kwak]{kong2022mdl}
Chaerin Kong, Jeesoo Kim, Donghoon Han, and Nojun Kwak.
\newblock Few-shot image generation with mixup-based distance learning.
\newblock In \emph{Proceedings of the European Conference on Computer Vision}, 2022.

\bibitem[Kong et~al.(2025)Kong, Zhang, Yang, Wang, Zhang, Wu, Chen, Liu, and Luo]{kong2025OMG}
Zhe Kong, Yong Zhang, Tianyu Yang, Tao Wang, Kaihao Zhang, Bizhu Wu, Guanying Chen, Wei Liu, and Wenhan Luo.
\newblock Omg: Occlusion-friendly personalized multi-concept generation in diffusion models.
\newblock In \emph{European Conference on Computer Vision}, pp.\  253--270. Springer, 2025.

\bibitem[Krizhevsky et~al.(2009)Krizhevsky, Hinton, et~al.]{krizhevsky2009cifar100}
Alex Krizhevsky, Geoffrey Hinton, et~al.
\newblock Learning multiple layers of features from tiny images, 2009.

\bibitem[Kulikov et~al.(2023)Kulikov, Yadin, Kleiner, and Michaeli]{kulikov2023sinddm}
Vladimir Kulikov, Shahar Yadin, Matan Kleiner, and Tomer Michaeli.
\newblock Sinddm: A single image denoising diffusion model.
\newblock In \emph{Proceedings of the International Conference on Machine Learning}, 2023.

\bibitem[Kumar \& Sivakumar(2023)Kumar and Sivakumar]{kumar2023deffGAN}
Rajiv Kumar and G~Sivakumar.
\newblock Deff-gan: Diverse attribute transfer for few-shot image synthesis.
\newblock \emph{arXiv preprint arXiv:2302.14533}, 2023.

\bibitem[Kumari et~al.(2022)Kumari, Zhang, Shechtman, and Zhu]{kumari2022ensembling}
Nupur Kumari, Richard Zhang, Eli Shechtman, and Jun-Yan Zhu.
\newblock Ensembling off-the-shelf models for gan training.
\newblock In \emph{Proceedings of the IEEE/CVF Conference on Computer Vision and Pattern Recognition}, 2022.

\bibitem[Kumari et~al.(2023{\natexlab{a}})Kumari, Zhang, Wang, Shechtman, Zhang, and Zhu]{kumari2023ablating}
Nupur Kumari, Bingliang Zhang, Sheng-Yu Wang, Eli Shechtman, Richard Zhang, and Jun-Yan Zhu.
\newblock Ablating concepts in text-to-image diffusion models.
\newblock In \emph{Proceedings of the IEEE/CVF International Conference on Computer Vision}, pp.\  22691--22702, 2023{\natexlab{a}}.

\bibitem[Kumari et~al.(2023{\natexlab{b}})Kumari, Zhang, Zhang, Shechtman, and Zhu]{kumari2023mcc}
Nupur Kumari, Bingliang Zhang, Richard Zhang, Eli Shechtman, and Jun-Yan Zhu.
\newblock Multi-concept customization of text-to-image diffusion.
\newblock In \emph{Proceedings of the IEEE/CVF Conference on Computer Vision and Pattern Recognition}, 2023{\natexlab{b}}.

\bibitem[Kwon \& Ye(2023)Kwon and Ye]{kwon2022oneclip}
Gihyun Kwon and Jong~Chul Ye.
\newblock One-shot adaptation of gan in just one clip.
\newblock \emph{IEEE Transactions on Pattern Analysis and Machine Intelligence}, 2023.

\bibitem[Kynk{\"a}{\"a}nniemi et~al.(2023)Kynk{\"a}{\"a}nniemi, Karras, Aittala, Aila, and Lehtinen]{kynkaanniemi2023the}
Tuomas Kynk{\"a}{\"a}nniemi, Tero Karras, Miika Aittala, Timo Aila, and Jaakko Lehtinen.
\newblock The role of imagenet classes in fr\'echet inception distance.
\newblock In \emph{International Conference on Learning Representations}, 2023.

\bibitem[Lamba \& Sophia(2025)Lamba and Sophia]{lamba-2025}
Kritika Lamba and Deborah Sophia.
\newblock {AI-generated music accounts for 18\% of all tracks uploaded to Deezer}.
\newblock 2025.
\newblock URL \url{https://www.reuters.com/technology/artificial-intelligence/ai-generated-music-accounts-18-all-tracks-uploaded-deezer-2025-04-16}.

\bibitem[Lee et~al.(2021)Lee, Kang, Jeong, Sul, and Zhang]{Lee2021C3}
Hyuk-Gi Lee, Gi-Cheon Kang, Changhoon Jeong, Han-Wool Sul, and Byoung-Tak Zhang.
\newblock $c^3$: Contrastive learning for cross-domain correspondence in few-shot image generation.
\newblock In \emph{Controllable Generative Modeling in Language and Vision Workshop at NeurIPS}, 2021.

\bibitem[Li et~al.(2024{\natexlab{a}})Li, Zhang, Nie, Han, Hu, Qiu, and Guo]{li2024styostylizefaceoneshot}
Bonan Li, Zicheng Zhang, Xuecheng Nie, Congying Han, Yinhan Hu, Xinmin Qiu, and Tiande Guo.
\newblock Styo: Stylize your face in only one-shot, 2024{\natexlab{a}}.
\newblock URL \url{https://arxiv.org/abs/2303.03231}.

\bibitem[Li et~al.(2023{\natexlab{a}})Li, Li, and Hoi]{li2023blipdiffusion}
Dongxu Li, Junnan Li, and Steven~CH Hoi.
\newblock Blip-diffusion: Pre-trained subject representation for controllable text-to-image generation and editing.
\newblock \emph{arXiv preprint arXiv:2305.14720}, 2023{\natexlab{a}}.

\bibitem[Li et~al.(2024{\natexlab{b}})Li, Liu, Lin, Zhang, Zhao, Zheng, Yang, Jiang, Wu, Cai, et~al.]{li2024unihda}
Hengjia Li, Yang Liu, Yuqi Lin, Zhanwei Zhang, Yibo Zhao, Tu~Zheng, Zheng Yang, Yuchun Jiang, Boxi Wu, Deng Cai, et~al.
\newblock Unihda: A unified and versatile framework for multi-modal hybrid domain adaptation.
\newblock \emph{arXiv preprint arXiv:2401.12596}, 2024{\natexlab{b}}.

\bibitem[Li et~al.(2024{\natexlab{c}})Li, Liu, Xia, Lin, Wang, Zheng, Yang, Zhong, Ren, and He]{li2023hda}
Hengjia Li, Yang Liu, Linxuan Xia, Yuqi Lin, Wenxiao Wang, Tu~Zheng, Zheng Yang, Xiaohui Zhong, Xiaobo Ren, and Xiaofei He.
\newblock Few-shot hybrid domain adaptation of image generator.
\newblock In \emph{The Twelfth International Conference on Learning Representations}, 2024{\natexlab{c}}.

\bibitem[Li et~al.(2025{\natexlab{a}})Li, Zhang, Zhu, and Ren]{li2025comprehensive}
Jun Li, Chenyang Zhang, Wei Zhu, and Yawei Ren.
\newblock A comprehensive survey of image generation models based on deep learning.
\newblock \emph{Annals of Data Science}, 12\penalty0 (1):\penalty0 141--170, 2025{\natexlab{a}}.

\bibitem[Li \& Malik(2018)Li and Malik]{li2018imle}
Ke~Li and Jitendra Malik.
\newblock Implicit maximum likelihood estimation.
\newblock \emph{arXiv preprint arXiv:1809.09087}, 2018.

\bibitem[Li et~al.(2022{\natexlab{a}})Li, Zhang, and Wang]{li2022hae}
Lingxiao Li, Yi~Zhang, and Shuhui Wang.
\newblock The euclidean space is evil: Hyperbolic attribute editing for few-shot image generation.
\newblock \emph{arXiv preprint arXiv:2211.12347}, 2022{\natexlab{a}}.

\bibitem[Li et~al.(2025{\natexlab{b}})Li, Nie, Chen, Jiang, Wu, Lin, Liu, Peng, Wang, and Zheng]{li2025TFIC}
Pengzhi Li, Qiang Nie, Ying Chen, Xi~Jiang, Kai Wu, Yuhuan Lin, Yong Liu, Jinlong Peng, Chengjie Wang, and Feng Zheng.
\newblock Tuning-free image customization with image and text guidance.
\newblock In \emph{European Conference on Computer Vision}, pp.\  233--250. Springer, 2025{\natexlab{b}}.

\bibitem[Li et~al.(2024{\natexlab{d}})Li, Pu, Zhao, Yang, Gu, Li, and Xu]{li2024dual}
Siqi Li, Yuanyuan Pu, Zhengpeng Zhao, Qiuxia Yang, Jinjing Gu, Yupan Li, and Dan Xu.
\newblock Dual-path hypernetworks of style and text for one-shot domain adaptation.
\newblock \emph{Applied Intelligence}, 54\penalty0 (3):\penalty0 2614--2630, 2024{\natexlab{d}}.

\bibitem[Li et~al.(2022{\natexlab{b}})Li, Li, Rockwell, Farimani, and Lee]{li2022moca}
Tianqin Li, Zijie Li, Harold Rockwell, Amir Farimani, and Tai~Sing Lee.
\newblock Prototype memory and attention mechanisms for few-shot image generation.
\newblock In \emph{Proceedings of the Eleventh International Conference on Learning Representations}, 2022{\natexlab{b}}.

\bibitem[Li et~al.(2023{\natexlab{b}})Li, Xu, Wu, Wang, Lu, Song, and Li]{li2023ammgan}
Wenkuan Li, Wenyi Xu, Xubin Wu, Qianshan Wang, Qiang Lu, Tianxia Song, and Haifang Li.
\newblock Ammgan: Adaptive multi-scale modulation generative adversarial network for few-shot image generation.
\newblock \emph{Applied Intelligence}, pp.\  1--19, 2023{\natexlab{b}}.

\bibitem[Li et~al.(2023{\natexlab{c}})Li, Fan, Hu, Feichtenhofer, and He]{li2023scaling}
Yanghao Li, Haoqi Fan, Ronghang Hu, Christoph Feichtenhofer, and Kaiming He.
\newblock Scaling language-image pre-training via masking.
\newblock In \emph{Proceedings of the IEEE/CVF Conference on Computer Vision and Pattern Recognition}, 2023{\natexlab{c}}.

\bibitem[Li et~al.(2020)Li, Zhang, Lu, and Shechtman]{li2020ewc}
Yijun Li, Richard Zhang, Jingwan Lu, and Eli Shechtman.
\newblock Few-shot image generation with elastic weight consolidation.
\newblock In \emph{Advances in Neural Information Processing Systems}, 2020.

\bibitem[Li et~al.(2022{\natexlab{c}})Li, Wang, Zheng, Zhang, and Li]{li2022fakeclr}
Ziqiang Li, Chaoyue Wang, Heliang Zheng, Jing Zhang, and Bin Li.
\newblock Fakeclr: Exploring contrastive learning for solving latent discontinuity in data-efficient gans.
\newblock In \emph{Proceedings of the European Conference on Computer Vision}, 2022{\natexlab{c}}.

\bibitem[Li et~al.(2022{\natexlab{d}})Li, Xia, Zhang, Wang, and Li]{li2022degan}
Ziqiang Li, Beihao Xia, Jing Zhang, Chaoyue Wang, and Bin Li.
\newblock A comprehensive survey on data-efficient gans in image generation.
\newblock \emph{arXiv preprint arXiv:2204.08329}, 2022{\natexlab{d}}.

\bibitem[Li et~al.(2024{\natexlab{e}})Li, Wang, Rui, Xue, Leng, Fu, and Li]{li2024pyp}
Ziqiang Li, Chaoyue Wang, Xue Rui, Chao Xue, Jiaxu Leng, Zhangjie Fu, and Bin Li.
\newblock Peer is your pillar: A data-unbalanced conditional gans for few-shot image generation.
\newblock \emph{IEEE Transactions on Circuits and Systems for Video Technology}, 2024{\natexlab{e}}.

\bibitem[Liang et~al.(2020)Liang, Liu, and Liu]{liang2020dawson}
Weixin Liang, Zixuan Liu, and Can Liu.
\newblock Dawson: A domain adaptive few shot generation framework.
\newblock \emph{arXiv preprint arXiv:2001.00576}, 2020.

\bibitem[Liu et~al.(2021)Liu, Zhu, Song, and Elgammal]{liu2021fastgan}
Bingchen Liu, Yizhe Zhu, Kunpeng Song, and Ahmed Elgammal.
\newblock Towards faster and stabilized gan training for high-fidelity few-shot image synthesis.
\newblock In \emph{International Conference on Learning Representations}, 2021.

\bibitem[Liu et~al.(2025{\natexlab{a}})Liu, Abdollahzadeh, and Cheung]{liu2025air}
Guimeng Liu, Milad Abdollahzadeh, and Ngai-Man Cheung.
\newblock Air: Zero-shot generative model adaptation with iterative refinement.
\newblock \emph{arXiv preprint arXiv:2506.10895}, 2025{\natexlab{a}}.

\bibitem[Liu et~al.(2025{\natexlab{b}})Liu, Hu, Song, Chen, and Wu]{liu2025fewconv}
Si-Hao Liu, Cong Hu, Xiao-Ning Song, Jia-Sheng Chen, and Xiao-Jun Wu.
\newblock Fewconv: Efficient variant convolution for few-shot image generation.
\newblock In \emph{International Conference on Pattern Recognition}, pp.\  424--440. Springer, 2025{\natexlab{b}}.

\bibitem[Liu et~al.(2022)Liu, Mao, Wu, Feichtenhofer, Darrell, and Xie]{liu2022convnet}
Zhuang Liu, Hanzi Mao, Chao-Yuan Wu, Christoph Feichtenhofer, Trevor Darrell, and Saining Xie.
\newblock A convnet for the 2020s.
\newblock In \emph{Proceedings of the IEEE/CVF Conference on Computer Vision and Pattern Recognition}, 2022.

\bibitem[Lu et~al.(2023)Lu, Tunanyan, Wang, Navasardyan, Wang, and Shi]{lu2023specialistdiffusion}
Haoming Lu, Hazarapet Tunanyan, Kai Wang, Shant Navasardyan, Zhangyang Wang, and Humphrey Shi.
\newblock Specialist diffusion: Plug-and-play sample-efficient fine-tuning of text-to-image diffusion models to learn any unseen style.
\newblock In \emph{Proceedings of the IEEE/CVF Conference on Computer Vision and Pattern Recognition}, 2023.

\bibitem[Madry et~al.(2018)Madry, Makelov, Schmidt, Tsipras, and Vladu]{madry2018towards}
Aleksander Madry, Aleksandar Makelov, Ludwig Schmidt, Dimitris Tsipras, and Adrian Vladu.
\newblock Towards deep learning models resistant to adversarial attacks.
\newblock In \emph{International Conference on Learning Representations}, 2018.

\bibitem[Mangla et~al.(2022)Mangla, Kumari, Singh, Krishnamurthy, and Balasubramanian]{mangla2022data}
Puneet Mangla, Nupur Kumari, Mayank Singh, Balaji Krishnamurthy, and Vineeth~N Balasubramanian.
\newblock Data instance prior (disp) in generative adversarial networks.
\newblock In \emph{Proceedings of the IEEE/CVF Winter Conference on Applications of Computer Vision}, pp.\  451--461, 2022.

\bibitem[Mao et~al.(2019)Mao, Lee, Tseng, Ma, and Yang]{mao2019mode}
Qi~Mao, Hsin-Ying Lee, Hung-Yu Tseng, Siwei Ma, and Ming-Hsuan Yang.
\newblock Mode seeking generative adversarial networks for diverse image synthesis.
\newblock In \emph{Proceedings of the IEEE/CVF conference on computer vision and pattern recognition}, pp.\  1429--1437, 2019.

\bibitem[Metz(2025{\natexlab{a}})]{metz-image-2025}
Cade Metz.
\newblock {OpenAI Unveils New Image Generator for ChatGPT}.
\newblock 2025{\natexlab{a}}.
\newblock URL \url{https://www.nytimes.com/2025/03/25/technology/chatgpt-image-generator.html?searchResultPosition=1}.

\bibitem[Metz(2025{\natexlab{b}})]{metz-text-2025}
Cade Metz.
\newblock {OpenAI Unveils A.I. Technology for ‘Natural Conversation’}.
\newblock 2025{\natexlab{b}}.
\newblock URL \url{https://www.nytimes.com/2025/02/27/technology/openai-artificial-intelligence-technology.html?searchResultPosition=16}.

\bibitem[Mirsky \& Lee(2021)Mirsky and Lee]{mirsky2021creation}
Yisroel Mirsky and Wenke Lee.
\newblock The creation and detection of deepfakes: A survey.
\newblock \emph{ACM Computing Surveys (CSUR)}, 54\penalty0 (1):\penalty0 1--41, 2021.

\bibitem[Mo et~al.(2020)Mo, Cho, and Shin]{mo2020freezed}
Sangwoo Mo, Minsu Cho, and Jinwoo Shin.
\newblock Freeze the discriminator: a simple baseline for fine-tuning gans.
\newblock \emph{CVPR AI for Content Creation Workshop}, 2020.

\bibitem[Mondal et~al.(2023)Mondal, Tiwary, Singla, and AP]{mondal2023lcl}
Arnab~Kumar Mondal, Piyush Tiwary, Parag Singla, and Prathosh AP.
\newblock Few-shot cross-domain image generation via inference-time latent-code learning.
\newblock In \emph{The Eleventh International Conference on Learning Representations}, 2023.

\bibitem[Mondal et~al.(2024)Mondal, Tiwary, Singla, and A.P.]{Mondal2024SoLAD}
Arnab~Kumar Mondal, Piyush Tiwary, Parag Singla, and Prathosh A.P.
\newblock Solad: Sampling over latent adapter for few shot generation.
\newblock \emph{IEEE Signal Processing Letters}, 2024.

\bibitem[Moon et~al.(2023)Moon, Kim, and Heo]{moon2023prosc}
Jongbo Moon, Hyunjun Kim, and Jae-Pil Heo.
\newblock Progressive few-shot adaptation of generative model with align-free spatial correlation.
\newblock In \emph{Proceedings of the AAAI Conference on Artificial Intelligence}, 2023.

\bibitem[Moon et~al.(2022)Moon, Choi, Lee, Ha, and Lee]{moon2022finetuning}
Taehong Moon, Moonseok Choi, Gayoung Lee, Jung-Woo Ha, and Juho Lee.
\newblock Fine-tuning diffusion models with limited data.
\newblock In \emph{NeurIPS 2022 Workshop on Score-Based Methods}, 2022.

\bibitem[Motamed et~al.(2025)Motamed, Paudel, and Van~Gool]{motamed2025Lego}
Saman Motamed, Danda~Pani Paudel, and Luc Van~Gool.
\newblock Lego: Learning to disentangle and invert personalized concepts beyond object appearance in text-to-image diffusion models.
\newblock In \emph{European Conference on Computer Vision}, pp.\  116--133. Springer, 2025.

\bibitem[Nam et~al.(2024)Nam, Kim, Lee, Jin, Kim, and Chang]{nam2024dreammatcher}
Jisu Nam, Heesu Kim, DongJae Lee, Siyoon Jin, Seungryong Kim, and Seunggyu Chang.
\newblock Dreammatcher: appearance matching self-attention for semantically-consistent text-to-image personalization.
\newblock In \emph{Proceedings of the IEEE/CVF Conference on Computer Vision and Pattern Recognition}, pp.\  8100--8110, 2024.

\bibitem[Nellis(2024)]{nellis-2024}
Stephen Nellis.
\newblock {Adobe to bring full AI image generation to Photoshop this year}.
\newblock 2024.
\newblock URL \url{https://www.reuters.com/technology/adobe-bring-full-ai-image-generation-photoshop-this-year-2024-04-23/}.

\bibitem[Nguyen et~al.(2023)Nguyen, Chandrasegaran, Abdollahzadeh, and Cheung]{nguyen2023rethinking}
Ngoc-Bao Nguyen, Keshigeyan Chandrasegaran, Milad Abdollahzadeh, and Ngai-Man Cheung.
\newblock Re-thinking model inversion attacks against deep neural networks.
\newblock In \emph{Proceedings of the IEEE/CVF Conference on Computer Vision and Pattern Recognition}, 2023.

\bibitem[Ni \& Koniusz(2023)Ni and Koniusz]{ni2023nice}
Yao Ni and Piotr Koniusz.
\newblock {NICE}: Noise-modulated consistency regularization for data-efficient {GAN}s.
\newblock In \emph{Thirty-seventh Conference on Neural Information Processing Systems}, 2023.

\bibitem[Ni \& Koniusz(2024)Ni and Koniusz]{ni2024chain}
Yao Ni and Piotr Koniusz.
\newblock Chain: Enhancing generalization in data-efficient gans via lipschitz continuity constrained normalization.
\newblock In \emph{Proceedings of the IEEE/CVF Conference on Computer Vision and Pattern Recognition}, pp.\  6763--6774, 2024.

\bibitem[Nichol et~al.(2018)Nichol, Achiam, and Schulman]{nichol2018reptile}
Alex Nichol, Joshua Achiam, and John Schulman.
\newblock On first-order meta-learning algorithms.
\newblock \emph{arXiv preprint arXiv:1803.02999}, 2018.

\bibitem[Nichol \& Dhariwal(2021)Nichol and Dhariwal]{nichol2021improveddenoisingDIM}
Alexander~Quinn Nichol and Prafulla Dhariwal.
\newblock Improved denoising diffusion probabilistic models.
\newblock In \emph{Proceedings of the International Conference on Machine Learning}, 2021.

\bibitem[Nie et~al.(2025)Nie, Zhu, You, Zhang, Ou, Hu, Zhou, Lin, Wen, and Li]{nie2025large}
Shen Nie, Fengqi Zhu, Zebin You, Xiaolu Zhang, Jingyang Ou, Jun Hu, Jun Zhou, Yankai Lin, Ji-Rong Wen, and Chongxuan Li.
\newblock Large language diffusion models.
\newblock \emph{arXiv preprint arXiv:2502.09992}, 2025.

\bibitem[Nikankin et~al.(2023)Nikankin, Haim, and Irani]{nikankin2022sinfusion}
Yaniv Nikankin, Niv Haim, and Michal Irani.
\newblock Sinfusion: Training diffusion models on a single image or video.
\newblock In \emph{International Conference on Machine Learning}, 2023.

\bibitem[Nilsback \& Zisserman(2008)Nilsback and Zisserman]{nilsback2008oxford_flower}
Maria-Elena Nilsback and Andrew Zisserman.
\newblock Automated flower classification over a large number of classes.
\newblock In \emph{Indian Conference on Computer Vision, Graphics \& Image Processing}, 2008.

\bibitem[Nitzan et~al.(2023)Nitzan, Gharbi, Zhang, Park, Zhu, Cohen-Or, and Shechtman]{nitzan2023domain}
Yotam Nitzan, Michael Gharbi, Richard Zhang, Taesung Park, Jun-Yan Zhu, Daniel Cohen-Or, and Eli Shechtman.
\newblock Domain expansion of image generators.
\newblock In \emph{Proceedings of the IEEE/CVF Conference on Computer Vision and Pattern Recognition (CVPR)}, 2023.

\bibitem[Noguchi \& Harada(2019)Noguchi and Harada]{noguchi2019bsa}
Atsuhiro Noguchi and Tatsuya Harada.
\newblock Image generation from small datasets via batch statistics adaptation.
\newblock In \emph{Proceedings of the IEEE/CVF International Conference on Computer Vision}, 2019.

\bibitem[Ojha et~al.(2021)Ojha, Li, Lu, Efros, Lee, Shechtman, and Zhang]{ojha2021cdc}
Utkarsh Ojha, Yijun Li, Jingwan Lu, Alexei~A Efros, Yong~Jae Lee, Eli Shechtman, and Richard Zhang.
\newblock Few-shot image generation via cross-domain correspondence.
\newblock In \emph{Proceedings of the IEEE/CVF Conference on Computer Vision and Pattern Recognition}, 2021.

\bibitem[Oppenheim et~al.(1997)Oppenheim, Willsky, Nawab, and Ding]{oppenheim1997signals}
Alan~V Oppenheim, Alan~S Willsky, Syed~Hamid Nawab, and Jian-Jiun Ding.
\newblock \emph{Signals and systems}, volume~2.
\newblock Prentice hall Upper Saddle River, NJ, 1997.

\bibitem[Pan \& Yang(2009)Pan and Yang]{pan2009yang-qiang-transfer}
Sinno~Jialin Pan and Qiang Yang.
\newblock A survey on transfer learning.
\newblock \emph{IEEE Transactions on knowledge and data engineering}, 22\penalty0 (10):\penalty0 1345--1359, 2009.

\bibitem[Pang et~al.(2024)Pang, Yin, Xie, Wang, Li, and Mao]{pang2023crossinitialization}
Lianyu Pang, Jian Yin, Haoran Xie, Qiping Wang, Qing Li, and Xudong Mao.
\newblock Cross initialization for face personalization of text-to-image models.
\newblock In \emph{IEEE/CVF Conference on Computer Vision and Pattern Recognition}, 2024.

\bibitem[Park et~al.(2024)Park, Jeong, Lee, Han, and Paik]{park2024promptsdm}
Jiwon Park, Dasol Jeong, Hyebean Lee, Seunghee Han, and Joonki Paik.
\newblock Prompt-based learning for image variation using single image multi-scale diffusion models.
\newblock \emph{IEEE Access}, 12:\penalty0 158810--158823, 2024.

\bibitem[Parmar et~al.(2022)Parmar, Zhang, and Zhu]{Parmar_2022_CVPR}
Gaurav Parmar, Richard Zhang, and Jun-Yan Zhu.
\newblock On aliased resizing and surprising subtleties in gan evaluation.
\newblock In \emph{Proceedings of the IEEE/CVF Conference on Computer Vision and Pattern Recognition (CVPR)}, 2022.

\bibitem[Pascual et~al.(2024)Pascual, Maiza, Sesma-Sara, Paternain, and Galar]{pascual2024enhancing}
Rub{\'e}n Pascual, Adri{\'a}n Maiza, Mikel Sesma-Sara, Daniel Paternain, and Mikel Galar.
\newblock Enhancing dreambooth with lora for generating unlimited characters with stable diffusion.
\newblock In \emph{2024 International Joint Conference on Neural Networks (IJCNN)}, pp.\  1--8. IEEE, 2024.

\bibitem[Peebles \& Xie(2023)Peebles and Xie]{peebles2023scalable}
William Peebles and Saining Xie.
\newblock Scalable diffusion models with transformers.
\newblock In \emph{Proceedings of the IEEE/CVF international conference on computer vision}, pp.\  4195--4205, 2023.

\bibitem[Peng et~al.(2024)Peng, Zhu, Jiang, Tai, Luo, Zhang, Lin, Jin, Wang, and Ji]{peng2024portraitbooth}
Xu~Peng, Junwei Zhu, Boyuan Jiang, Ying Tai, Donghao Luo, Jiangning Zhang, Wei Lin, Taisong Jin, Chengjie Wang, and Rongrong Ji.
\newblock Portraitbooth: A versatile portrait model for fast identity-preserved personalization.
\newblock In \emph{Proceedings of the IEEE/CVF Conference on Computer Vision and Pattern Recognition}, pp.\  27080--27090, 2024.

\bibitem[Phaphuangwittayakul et~al.(2021)Phaphuangwittayakul, Guo, and Ying]{phaphuangwittayakul2021faml}
Aniwat Phaphuangwittayakul, Yi~Guo, and Fangli Ying.
\newblock Fast adaptive meta-learning for few-shot image generation.
\newblock \emph{IEEE Transactions on Multimedia}, 24:\penalty0 2205--2217, 2021.

\bibitem[Phaphuangwittayakul et~al.(2022)Phaphuangwittayakul, Ying, Guo, Zhou, and Chakpitak]{phaphuangwittayakul2022cmlgan}
Aniwat Phaphuangwittayakul, Fangli Ying, Yi~Guo, Liting Zhou, and Nopasit Chakpitak.
\newblock Few-shot image generation based on contrastive meta-learning generative adversarial network.
\newblock \emph{The Visual Computer}, pp.\  1--14, 2022.

\bibitem[Phung et~al.(2005)Phung, Duong, Venkatesh, and Bui]{phung2005topic}
Dinh~Q Phung, Thi~V Duong, Svetha Venkatesh, and Hung~H Bui.
\newblock Topic transition detection using hierarchical hidden markov and semi-markov models.
\newblock In \emph{Proceedings of the 13th annual ACM international conference on Multimedia}, pp.\  11--20, 2005.

\bibitem[Po et~al.(2024)Po, Yang, Aberman, and Wetzstein]{po2024orthogonal}
Ryan Po, Guandao Yang, Kfir Aberman, and Gordon Wetzstein.
\newblock Orthogonal adaptation for modular customization of diffusion models.
\newblock In \emph{Proceedings of the IEEE/CVF Conference on Computer Vision and Pattern Recognition}, pp.\  7964--7973, 2024.

\bibitem[Porwik \& Lisowska(2004)Porwik and Lisowska]{porwik2004haar}
Piotr Porwik and Agnieszka Lisowska.
\newblock The haar-wavelet transform in digital image processing: its status and achievements.
\newblock \emph{Machine graphics and vision}, 13\penalty0 (1/2):\penalty0 79--98, 2004.

\bibitem[Pouyanfar et~al.(2018)Pouyanfar, Sadiq, Yan, Tian, Tao, Reyes, Shyu, Chen, and Iyengar]{pouyanfar2018survey}
Samira Pouyanfar, Saad Sadiq, Yilin Yan, Haiman Tian, Yudong Tao, Maria~Presa Reyes, Mei-Ling Shyu, Shu-Ching Chen, and Sundaraja~S Iyengar.
\newblock A survey on deep learning: Algorithms, techniques, and applications.
\newblock \emph{ACM Computing Surveys (CSUR)}, 51\penalty0 (5):\penalty0 1--36, 2018.

\bibitem[Qiao et~al.(2024)Qiao, Shang, Liu, Sun, Ji, and Chen]{qiao2024facechain}
Pengchong Qiao, Lei Shang, Chang Liu, Baigui Sun, Xiangyang Ji, and Jie Chen.
\newblock Facechain-sude: Building derived class to inherit category attributes for one-shot subject-driven generation.
\newblock In \emph{Proceedings of the IEEE/CVF Conference on Computer Vision and Pattern Recognition}, pp.\  7215--7224, 2024.

\bibitem[Radford et~al.(2021)Radford, Kim, Hallacy, Ramesh, Goh, Agarwal, Sastry, Askell, Mishkin, Clark, et~al.]{radford2021CLIP}
Alec Radford, Jong~Wook Kim, Chris Hallacy, Aditya Ramesh, Gabriel Goh, Sandhini Agarwal, Girish Sastry, Amanda Askell, Pamela Mishkin, Jack Clark, et~al.
\newblock Learning transferable visual models from natural language supervision.
\newblock In \emph{Proceedings of the International Conference on Machine Learning}, 2021.

\bibitem[Rahaman et~al.(2019)Rahaman, Baratin, Arpit, Draxler, Lin, Hamprecht, Bengio, and Courville]{rahaman2019nnfrequncybias}
Nasim Rahaman, Aristide Baratin, Devansh Arpit, Felix Draxler, Min Lin, Fred Hamprecht, Yoshua Bengio, and Aaron Courville.
\newblock On the spectral bias of neural networks.
\newblock In \emph{Proceedings of the International Conference on Machine Learning}, 2019.

\bibitem[Ram et~al.(2025)Ram, Neiman, Feng, Stuart, Tran, and Chilimbi]{ram2025dreamblend}
Shwetha Ram, Tal Neiman, Qianli Feng, Andrew Stuart, Son Tran, and Trishul Chilimbi.
\newblock Dreamblend: Advancing personalized fine-tuning of text-to-image diffusion models.
\newblock In \emph{WACV}, 2025.

\bibitem[Ramesh et~al.(2021)Ramesh, Pavlov, Goh, Gray, Voss, Radford, Chen, and Sutskever]{ramesh2021dalle}
Aditya Ramesh, Mikhail Pavlov, Gabriel Goh, Scott Gray, Chelsea Voss, Alec Radford, Mark Chen, and Ilya Sutskever.
\newblock Zero-shot text-to-image generation.
\newblock In \emph{Proceedings of the International Conference on Machine Learning}, 2021.

\bibitem[Ramesh et~al.(2022)Ramesh, Dhariwal, Nichol, Chu, and Chen]{ramesh2022dalle2}
Aditya Ramesh, Prafulla Dhariwal, Alex Nichol, Casey Chu, and Mark Chen.
\newblock Hierarchical text-conditional image generation with clip latents.
\newblock \emph{arXiv preprint arXiv:2204.06125}, 2022.

\bibitem[Reynolds et~al.(2009)]{reynolds2009gaussian}
Douglas~A Reynolds et~al.
\newblock Gaussian mixture models.
\newblock \emph{Encyclopedia of biometrics}, 741\penalty0 (659-663), 2009.

\bibitem[Robb et~al.(2020)Robb, Chu, Kumar, and Huang]{robb2020svd}
Esther Robb, Wen-Sheng Chu, Abhishek Kumar, and Jia-Bin Huang.
\newblock Few-shot adaptation of generative adversarial networks.
\newblock \emph{arXiv preprint arXiv:2010.11943}, 2020.

\bibitem[Rombach et~al.(2022)Rombach, Blattmann, Lorenz, Esser, and Ommer]{rombach2022latentdiffusion}
Robin Rombach, Andreas Blattmann, Dominik Lorenz, Patrick Esser, and Bj{\"o}rn Ommer.
\newblock High-resolution image synthesis with latent diffusion models.
\newblock In \emph{Proceedings of the IEEE/CVF Conference on Computer Vision and Pattern Recognition}, 2022.

\bibitem[Ronneberger et~al.(2015)Ronneberger, Fischer, and Brox]{ronneberger2015unet}
Olaf Ronneberger, Philipp Fischer, and Thomas Brox.
\newblock U-net: Convolutional networks for biomedical image segmentation.
\newblock In \emph{International Conference on Medical Image Computing and Computer-Assisted Intervention}, 2015.

\bibitem[Ruiz et~al.(2023)Ruiz, Li, Jampani, Pritch, Rubinstein, and Aberman]{ruiz2023dreambooth}
Nataniel Ruiz, Yuanzhen Li, Varun Jampani, Yael Pritch, Michael Rubinstein, and Kfir Aberman.
\newblock Dreambooth: Fine tuning text-to-image diffusion models for subject-driven generation.
\newblock In \emph{Proceedings of the IEEE/CVF Conference on Computer Vision and Pattern Recognition}, 2023.

\bibitem[Ruiz et~al.(2024)Ruiz, Li, Jampani, Wei, Hou, Pritch, Wadhwa, Rubinstein, and Aberman]{ruiz2024hyperdreambooth}
Nataniel Ruiz, Yuanzhen Li, Varun Jampani, Wei Wei, Tingbo Hou, Yael Pritch, Neal Wadhwa, Michael Rubinstein, and Kfir Aberman.
\newblock Hyperdreambooth: Hypernetworks for fast personalization of text-to-image models.
\newblock In \emph{Proceedings of the IEEE/CVF Conference on Computer Vision and Pattern Recognition}, pp.\  6527--6536, 2024.

\bibitem[Saharia et~al.(2022)Saharia, Chan, Saxena, Li, Whang, Denton, Ghasemipour, Gontijo~Lopes, Karagol~Ayan, Salimans, et~al.]{saharia2022imagen}
Chitwan Saharia, William Chan, Saurabh Saxena, Lala Li, Jay Whang, Emily~L Denton, Kamyar Ghasemipour, Raphael Gontijo~Lopes, Burcu Karagol~Ayan, Tim Salimans, et~al.
\newblock Photorealistic text-to-image diffusion models with deep language understanding.
\newblock In \emph{Advances in Neural Information Processing Systems}, 2022.

\bibitem[Salimans et~al.(2016)Salimans, Goodfellow, Zaremba, Cheung, Radford, and Chen]{salimans2016improved}
Tim Salimans, Ian Goodfellow, Wojciech Zaremba, Vicki Cheung, Alec Radford, and Xi~Chen.
\newblock Improved techniques for training gans.
\newblock In \emph{Advances in Neural Information Processing Systems}, 2016.

\bibitem[Sanghi et~al.(2023)Sanghi, Fu, Liu, Willis, Shayani, Khasahmadi, Sridhar, and Ritchie]{sanghi2023clipsculptor}
Aditya Sanghi, Rao Fu, Vivian Liu, Karl~DD Willis, Hooman Shayani, Amir~H Khasahmadi, Srinath Sridhar, and Daniel Ritchie.
\newblock Clip-sculptor: Zero-shot generation of high-fidelity and diverse shapes from natural language.
\newblock In \emph{Proceedings of the IEEE/CVF Conference on Computer Vision and Pattern Recognition}, 2023.

\bibitem[Santos \& Papa(2022)Santos and Papa]{santos2022avoiding}
Claudio Filipi Gon{\c{c}}alves~Dos Santos and Jo{\~a}o~Paulo Papa.
\newblock Avoiding overfitting: A survey on regularization methods for convolutional neural networks.
\newblock \emph{ACM Computing Surveys}, 54\penalty0 (10s):\penalty0 1--25, 2022.

\bibitem[Sauer et~al.(2021)Sauer, Chitta, M{\"u}ller, and Geiger]{sauer2021projectedgan}
Axel Sauer, Kashyap Chitta, Jens M{\"u}ller, and Andreas Geiger.
\newblock Projected gans converge faster.
\newblock In \emph{Advances in Neural Information Processing Systems}, 2021.

\bibitem[Saxena et~al.(2023)Saxena, Cao, Xu, and Kulshrestha]{saxena2023regan}
Divya Saxena, Jiannong Cao, Jiahao Xu, and Tarun Kulshrestha.
\newblock Re-gan: Data-efficient gans training via architectural reconfiguration.
\newblock In \emph{Proceedings of the IEEE/CVF Conference on Computer Vision and Pattern Recognition}, 2023.

\bibitem[Saxena et~al.(2024)Saxena, Cao, Xu, and Kulshrestha]{saxena2024rggan}
Divya Saxena, Jiannong Cao, Jiahao Xu, and Tarun Kulshrestha.
\newblock Rg-gan: Dynamic regenerative pruning for data-efficient generative adversarial networks.
\newblock In \emph{Proceedings of the AAAI Conference on Artificial Intelligence}, volume~38, pp.\  4704--4712, 2024.

\bibitem[Schiff et~al.(2024)Schiff, Kao, Gokaslan, Dao, Gu, and Kuleshov]{schiff2024caduceus}
Yair Schiff, Chia-Hsiang Kao, Aaron Gokaslan, Tri Dao, Albert Gu, and Volodymyr Kuleshov.
\newblock Caduceus: Bi-directional equivariant long-range dna sequence modeling.
\newblock \emph{arXiv preprint arXiv:2403.03234}, 2024.

\bibitem[Schuhmann et~al.(2021)Schuhmann, Vencu, Beaumont, Kaczmarczyk, Mullis, Katta, Coombes, Jitsev, and Komatsuzaki]{schuhmann2021laion400m}
Christoph Schuhmann, Richard Vencu, Romain Beaumont, Robert Kaczmarczyk, Clayton Mullis, Aarush Katta, Theo Coombes, Jenia Jitsev, and Aran Komatsuzaki.
\newblock Laion-400m: Open dataset of clip-filtered 400 million image-text pairs, 2021.

\bibitem[Schwarz et~al.(2021)Schwarz, Liao, and Geiger]{schwarz2021frequency}
Katja Schwarz, Yiyi Liao, and Andreas Geiger.
\newblock On the frequency bias of generative models.
\newblock In \emph{Advances in Neural Information Processing Systems}, 2021.

\bibitem[Seo et~al.(2023)Seo, Kang, and Park]{seo2023lfsgan}
Juwon Seo, Ji-Su Kang, and Gyeong-Moon Park.
\newblock Lfs-gan: Lifelong few-shot image generation.
\newblock In \emph{Proceedings of the IEEE/CVF International Conference on Computer Vision}, pp.\  11356--11366, 2023.

\bibitem[Shah et~al.(2025)Shah, Ruiz, Cole, Lu, Lazebnik, Li, and Jampani]{shah2025ziplora}
Viraj Shah, Nataniel Ruiz, Forrester Cole, Erika Lu, Svetlana Lazebnik, Yuanzhen Li, and Varun Jampani.
\newblock Ziplora: Any subject in any style by effectively merging loras.
\newblock In \emph{European Conference on Computer Vision}, pp.\  422--438. Springer, 2025.

\bibitem[Shaham et~al.(2019)Shaham, Dekel, and Michaeli]{shaham2019singan}
Tamar~Rott Shaham, Tali Dekel, and Tomer Michaeli.
\newblock Singan: Learning a generative model from a single natural image.
\newblock In \emph{Proceedings of the IEEE/CVF International Conference on Computer Vision}, 2019.

\bibitem[Shahbazi et~al.(2021)Shahbazi, Huang, Paudel, Chhatkuli, and Van~Gool]{shahbazi2021efficient}
Mohamad Shahbazi, Zhiwu Huang, Danda~Pani Paudel, Ajad Chhatkuli, and Luc Van~Gool.
\newblock Efficient conditional gan transfer with knowledge propagation across classes.
\newblock In \emph{Proceedings of the IEEE/CVF Conference on Computer Vision and Pattern Recognition}, 2021.

\bibitem[Shahbazi et~al.(2022)Shahbazi, Danelljan, Paudel, and Gool]{shahbazi2022collapse}
Mohamad Shahbazi, Martin Danelljan, Danda~Pani Paudel, and Luc~Van Gool.
\newblock Collapse by conditioning: Training class-conditional gans with limited data.
\newblock In \emph{International Conference on Learning Representations}, 2022.

\bibitem[Shearing(2025)]{shearing-2025}
Hazel Shearing.
\newblock {Teachers can use AI to save time on marking, new guidance says}.
\newblock 2025.
\newblock URL \url{https://www.bbc.com/news/articles/c1kvyj7dkp0o}.

\bibitem[Shi et~al.(2023{\natexlab{a}})Shi, Liu, Zhou, and Zhou]{shi2023autoinfogan}
Jiachen Shi, Wenzhen Liu, Guoqiang Zhou, and Yuming Zhou.
\newblock Autoinfo gan: Toward a better image synthesis gan framework for high-fidelity few-shot datasets via nas and contrastive learning.
\newblock \emph{Knowledge-Based Systems}, 276:\penalty0 110757, 2023{\natexlab{a}}.

\bibitem[Shi et~al.(2023{\natexlab{b}})Shi, Xiong, Lin, and Jung]{shi2023instantbooth}
Jing Shi, Wei Xiong, Zhe Lin, and Hyun~Joon Jung.
\newblock Instantbooth: Personalized text-to-image generation without test-time finetuning.
\newblock \emph{arXiv preprint arXiv:2304.03411}, 2023{\natexlab{b}}.

\bibitem[Sinha et~al.(2021{\natexlab{a}})Sinha, Ayush, Song, Uzkent, Jin, and Ermon]{sinha2021negative}
Abhishek Sinha, Kumar Ayush, Jiaming Song, Burak Uzkent, Hongxia Jin, and Stefano Ermon.
\newblock Negative data augmentation.
\newblock \emph{arXiv preprint arXiv:2102.05113}, 2021{\natexlab{a}}.

\bibitem[Sinha et~al.(2021{\natexlab{b}})Sinha, Song, Meng, and Ermon]{sinha2021d2c}
Abhishek Sinha, Jiaming Song, Chenlin Meng, and Stefano Ermon.
\newblock D2c: Diffusion-decoding models for few-shot conditional generation.
\newblock In \emph{Advances in Neural Information Processing Systems}, 2021{\natexlab{b}}.

\bibitem[Snell et~al.(2017)Snell, Swersky, and Zemel]{snell2017prototypical}
Jake Snell, Kevin Swersky, and Richard Zemel.
\newblock Prototypical networks for few-shot learning.
\newblock In \emph{Advances in Neural Information Processing Systems}, 2017.

\bibitem[Sohn et~al.(2023)Sohn, Chang, Lezama, Polania, Zhang, Hao, Essa, and Jiang]{sohn2023vpt}
Kihyuk Sohn, Huiwen Chang, Jos{\'e} Lezama, Luisa Polania, Han Zhang, Yuan Hao, Irfan Essa, and Lu~Jiang.
\newblock Visual prompt tuning for generative transfer learning.
\newblock In \emph{Proceedings of the IEEE/CVF Conference on Computer Vision and Pattern Recognition}, 2023.

\bibitem[Song et~al.(2020)Song, Meng, and Ermon]{song2020denoising}
Jiaming Song, Chenlin Meng, and Stefano Ermon.
\newblock Denoising diffusion implicit models.
\newblock \emph{arXiv:2010.02502}, 2020.

\bibitem[Song et~al.(2024{\natexlab{a}})Song, Han, Liu, Metaxas, and Elgammal]{song2024stylegan}
Kunpeng Song, Ligong Han, Bingchen Liu, Dimitris Metaxas, and Ahmed Elgammal.
\newblock Stylegan-fusion: Diffusion guided domain adaptation of image generators.
\newblock In \emph{Proceedings of the IEEE/CVF Winter Conference on Applications of Computer Vision}, pp.\  5453--5463, 2024{\natexlab{a}}.

\bibitem[Song et~al.(2025)Song, Zhu, Liu, Yan, Elgammal, and Yang]{song2025MOMA}
Kunpeng Song, Yizhe Zhu, Bingchen Liu, Qing Yan, Ahmed Elgammal, and Xiao Yang.
\newblock Moma: Multimodal llm adapter for fast personalized image generation.
\newblock In \emph{European Conference on Computer Vision}, pp.\  117--132. Springer, 2025.

\bibitem[Song et~al.(2024{\natexlab{b}})Song, Yang, Yang, and Lin]{song2024towards}
Nan Song, Xiaofeng Yang, Ze~Yang, and Guosheng Lin.
\newblock Towards lifelong few-shot customization of text-to-image diffusion.
\newblock \emph{arXiv preprint arXiv:2411.05544}, 2024{\natexlab{b}}.

\bibitem[Song et~al.(2023)Song, Wang, Cai, Mondal, and Sahoo]{song2023comprehensive}
Yisheng Song, Ting Wang, Puyu Cai, Subrota~K Mondal, and Jyoti~Prakash Sahoo.
\newblock A comprehensive survey of few-shot learning: Evolution, applications, challenges, and opportunities.
\newblock \emph{ACM Computing Surveys}, 55\penalty0 (13s):\penalty0 1--40, 2023.

\bibitem[Sun et~al.(2024)Sun, Liang, Dong, Li, Ding, and Cong]{sun2024create}
Gan Sun, Wenqi Liang, Jiahua Dong, Jun Li, Zhengming Ding, and Yang Cong.
\newblock Create your world: Lifelong text-to-image diffusion.
\newblock \emph{IEEE Transactions on Pattern Analysis and Machine Intelligence}, 2024.

\bibitem[Sun et~al.(2019)Sun, Liu, Chua, and Schiele]{sun2019meta}
Qianru Sun, Yaoyao Liu, Tat-Seng Chua, and Bernt Schiele.
\newblock Meta-transfer learning for few-shot learning.
\newblock In \emph{Proceedings of the IEEE/CVF conference on computer vision and pattern recognition}, pp.\  403--412, 2019.

\bibitem[Sung et~al.(2018)Sung, Yang, Zhang, Xiang, Torr, and Hospedales]{sung2018relationnet}
Flood Sung, Yongxin Yang, Li~Zhang, Tao Xiang, Philip~HS Torr, and Timothy~M Hospedales.
\newblock Learning to compare: Relation network for few-shot learning.
\newblock In \emph{Proceedings of the IEEE/CVF Conference on Computer Vision and Pattern Recognition}, 2018.

\bibitem[Sushko et~al.(2021)Sushko, Gall, and Khoreva]{sushko2021oneshotgan}
Vadim Sushko, Jurgen Gall, and Anna Khoreva.
\newblock One-shot gan: Learning to generate samples from single images and videos.
\newblock In \emph{Proceedings of the IEEE/CVF Conference on Computer Vision and Pattern Recognition}, 2021.

\bibitem[Sushko et~al.(2023)Sushko, Wang, and Gall]{sushko2023smoothsim}
Vadim Sushko, Ruyu Wang, and Juergen Gall.
\newblock Smoothness similarity regularization for few-shot gan adaptation.
\newblock In \emph{Proceedings of the IEEE/CVF International Conference on Computer Vision}, pp.\  7073--7082, 2023.

\bibitem[Szegedy et~al.(2016)Szegedy, Vanhoucke, Ioffe, Shlens, and Wojna]{szegedy2016labelsmoothing}
Christian Szegedy, Vincent Vanhoucke, Sergey Ioffe, Jon Shlens, and Zbigniew Wojna.
\newblock Rethinking the inception architecture for computer vision.
\newblock In \emph{Proceedings of the IEEE/CVF Conference on Computer Vision and Pattern Recognition}, 2016.

\bibitem[Tan et~al.(2018)Tan, Sun, Kong, Zhang, Yang, and Liu]{tan2018survey}
Chuanqi Tan, Fuchun Sun, Tao Kong, Wenchang Zhang, Chao Yang, and Chunfang Liu.
\newblock A survey on deep transfer learning.
\newblock In \emph{Artificial Neural Networks and Machine Learning--ICANN 2018: 27th International Conference on Artificial Neural Networks, Rhodes, Greece, October 4-7, 2018, Proceedings, Part III 27}, pp.\  270--279. Springer, 2018.

\bibitem[Tancik et~al.(2020)Tancik, Srinivasan, Mildenhall, Fridovich-Keil, Raghavan, Singhal, Ramamoorthi, Barron, and Ng]{tancik2020fourier}
Matthew Tancik, Pratul Srinivasan, Ben Mildenhall, Sara Fridovich-Keil, Nithin Raghavan, Utkarsh Singhal, Ravi Ramamoorthi, Jonathan Barron, and Ren Ng.
\newblock Fourier features let networks learn high frequency functions in low dimensional domains.
\newblock In \emph{Advances in Neural Information Processing Systems}, 2020.

\bibitem[Teo et~al.(2024{\natexlab{a}})Teo, Abdollahzadeh, and Cheung]{teo2024fairtl}
Christopher T.~H. Teo, Milad Abdollahzadeh, and Ngai-Man Cheung.
\newblock Fairtl: A transfer learning approach for bias mitigation in deep generative models.
\newblock \emph{IEEE Journal of Selected Topics in Signal Processing}, 18\penalty0 (2):\penalty0 155--167, 2024{\natexlab{a}}.
\newblock \doi{10.1109/JSTSP.2024.3363419}.

\bibitem[Teo et~al.(2023)Teo, Abdollahzadeh, and Cheung]{teo2023fairtl}
Christopher~TH Teo, Milad Abdollahzadeh, and Ngai-Man Cheung.
\newblock Fair generative models via transfer learning.
\newblock In \emph{Proceedings of the AAAI Conference on Artificial Intelligence}, 2023.

\bibitem[Teo et~al.(2024{\natexlab{b}})Teo, Abdollahzadeh, Ma, and Cheung]{teo2024fairqueue}
Christopher~TH Teo, Milad Abdollahzadeh, Xinda Ma, and Ngai-man Cheung.
\newblock Fairqueue: Rethinking prompt learning for fair text-to-image generation.
\newblock In \emph{The Thirty-eighth Annual Conference on Neural Information Processing Systems}, 2024{\natexlab{b}}.

\bibitem[Thopalli et~al.(2023)Thopalli, Subramanyam, Turaga, and Thiagarajan]{thopalli2023targetaware}
Kowshik Thopalli, Rakshith Subramanyam, Pavan Turaga, and Jayaraman~J Thiagarajan.
\newblock Target-aware generative augmentations for single-shot adaptation.
\newblock \emph{arXiv preprint arXiv:2305.13284}, 2023.

\bibitem[Tong \& Hu(2025)Tong and Hu]{tong-2025}
Anna Tong and Krystal Hu.
\newblock {AI startups revolutionize coding industry, leading to sky-high valuations}.
\newblock 2025.
\newblock URL \url{https://www.reuters.com/business/ai-vibe-coding-startups-burst-onto-scene-with-sky-high-valuations-2025-06-03/}.

\bibitem[Tran et~al.(2021)Tran, Tran, Nguyen, Nguyen, and Cheung]{tran2021dag}
Ngoc-Trung Tran, Viet-Hung Tran, Ngoc-Bao Nguyen, Trung-Kien Nguyen, and Ngai-Man Cheung.
\newblock On data augmentation for gan training.
\newblock \emph{IEEE Transactions on Image Processing}, 30:\penalty0 1882--1897, 2021.

\bibitem[Tseng et~al.(2021)Tseng, Jiang, Liu, Yang, and Yang]{tseng2021lecam}
Hung-Yu Tseng, Lu~Jiang, Ce~Liu, Ming-Hsuan Yang, and Weilong Yang.
\newblock Regularizing generative adversarial networks under limited data.
\newblock In \emph{Proceedings of the IEEE/CVF Conference on Computer Vision and Pattern Recognition}, 2021.

\bibitem[Vahdat \& Kautz(2020)Vahdat and Kautz]{vahdat2020nvae}
Arash Vahdat and Jan Kautz.
\newblock Nvae: A deep hierarchical variational autoencoder.
\newblock In \emph{Advances in Neural Information Processing Systems}, 2020.

\bibitem[Van Den~Oord et~al.(2017)Van Den~Oord, Vinyals, et~al.]{van2017neural}
Aaron Van Den~Oord, Oriol Vinyals, et~al.
\newblock Neural discrete representation learning.
\newblock In \emph{Advances in Neural Information Processing Systems}, 2017.

\bibitem[van~der Maaten \& Hinton(2008)van~der Maaten and Hinton]{JMLR:v9:vandermaaten08a_tsne}
Laurens van~der Maaten and Geoffrey Hinton.
\newblock Visualizing data using t-sne.
\newblock \emph{Journal of Machine Learning Research}, 9\penalty0 (86):\penalty0 2579--2605, 2008.

\bibitem[Van~Loan(1976)]{van1976generalizing}
Charles~F Van~Loan.
\newblock Generalizing the singular value decomposition.
\newblock \emph{SIAM Journal on Numerical Analysis}, 13\penalty0 (1):\penalty0 76--83, 1976.

\bibitem[Varshney et~al.(2021)Varshney, Verma, Srijith, Carin, and Rai]{varshney2021camgan}
Sakshi Varshney, Vinay~Kumar Verma, PK~Srijith, Lawrence Carin, and Piyush Rai.
\newblock Cam-gan: Continual adaptation modules for generative adversarial networks.
\newblock In \emph{Advances in Neural Information Processing Systems}, 2021.

\bibitem[Vashist et~al.(2024)Vashist, Peng, and Li]{vashist2024rejection}
Chirag Vashist, Shichong Peng, and Ke~Li.
\newblock Rejection sampling imle: Designing priors for better few-shot image synthesis.
\newblock In \emph{European Conference on Computer Vision}, pp.\  441--456. Springer, 2024.

\bibitem[Villani et~al.(2009)]{villani2009optimal}
C{\'e}dric Villani et~al.
\newblock \emph{Optimal transport: old and new}, volume 338.
\newblock Springer, 2009.

\bibitem[Vinyals et~al.(2016)Vinyals, Blundell, Lillicrap, Wierstra, et~al.]{vinyals2016matching}
Oriol Vinyals, Charles Blundell, Timothy Lillicrap, Daan Wierstra, et~al.
\newblock Matching networks for one shot learning.
\newblock In \emph{Advances in Neural Information Processing Systems}, 2016.

\bibitem[Wang et~al.(2023{\natexlab{a}})Wang, Zhou, Ge, Jiang, Bao, and Xu]{wang2023cffont}
Chi Wang, Min Zhou, Tiezheng Ge, Yuning Jiang, Hujun Bao, and Weiwei Xu.
\newblock Cf-font: Content fusion for few-shot font generation.
\newblock In \emph{Proceedings of the IEEE/CVF Conference on Computer Vision and Pattern Recognition}, 2023{\natexlab{a}}.

\bibitem[Wang et~al.(2019)Wang, Liew, Zou, Zhou, and Feng]{wang2019panet}
Kaixin Wang, Jun~Hao Liew, Yingtian Zou, Daquan Zhou, and Jiashi Feng.
\newblock Panet: Few-shot image semantic segmentation with prototype alignment.
\newblock In \emph{Proceedings of the IEEE/CVF International Conference on Computer Vision}, 2019.

\bibitem[Wang et~al.(2022{\natexlab{a}})Wang, Bao, Zhou, Chen, Chen, Yuan, and Li]{wang2022sindiffusion}
Weilun Wang, Jianmin Bao, Wengang Zhou, Dongdong Chen, Dong Chen, Lu~Yuan, and Houqiang Li.
\newblock Sindiffusion: Learning a diffusion model from a single natural image.
\newblock \emph{arXiv preprint arXiv:2211.12445}, 2022{\natexlab{a}}.

\bibitem[Wang \& Tang(2008)Wang and Tang]{wang2008faceSketches}
Xiaogang Wang and Xiaoou Tang.
\newblock Face photo-sketch synthesis and recognition.
\newblock \emph{IEEE Transactions on Pattern Analysis and Machine Intelligence}, 31\penalty0 (11):\penalty0 1955--1967, 2008.

\bibitem[Wang et~al.(2023{\natexlab{b}})Wang, Lin, Liu, and Xu]{wang2023efficient}
Xiyu Wang, Baijiong Lin, Daochang Liu, and Chang Xu.
\newblock Efficient transfer learning in diffusion models via adversarial noise.
\newblock \emph{arXiv preprint arXiv:2308.11948}, 2023{\natexlab{b}}.

\bibitem[Wang et~al.(2024)Wang, Lin, Liu, Chen, and Xu]{wang2024bridging}
Xiyu Wang, Baijiong Lin, Daochang Liu, Ying-Cong Chen, and Chang Xu.
\newblock Bridging data gaps in diffusion models with adversarial noise-based transfer learning.
\newblock In \emph{Forty-first International Conference on Machine Learning}, 2024.

\bibitem[Wang et~al.(2022{\natexlab{b}})Wang, Jiang, Zhao, Liu, and Wang]{wang2022ccasingan}
Xueqin Wang, Wenzong Jiang, Lifei Zhao, Baodi Liu, and Yanjiang Wang.
\newblock Ccasingan: Cascaded channel attention guided single-image gans.
\newblock In \emph{2022 16th IEEE International Conference on Signal Processing (ICSP)}, volume~1, pp.\  61--65. IEEE, 2022{\natexlab{b}}.

\bibitem[Wang et~al.(2020{\natexlab{a}})Wang, Yao, Kwok, and Ni]{wang2020generalizing}
Yaqing Wang, Quanming Yao, James~T Kwok, and Lionel~M Ni.
\newblock Generalizing from a few examples: A survey on few-shot learning.
\newblock \emph{ACM computing surveys (csur)}, 53\penalty0 (3):\penalty0 1--34, 2020{\natexlab{a}}.

\bibitem[Wang et~al.(2018)Wang, Wu, Herranz, Van~de Weijer, Gonzalez-Garcia, and Raducanu]{wang2018tgan}
Yaxing Wang, Chenshen Wu, Luis Herranz, Joost Van~de Weijer, Abel Gonzalez-Garcia, and Bogdan Raducanu.
\newblock Transferring gans: Generating images from limited data.
\newblock In \emph{Proceedings of the European Conference on Computer Vision}, 2018.

\bibitem[Wang et~al.(2020{\natexlab{b}})Wang, Gonzalez-Garcia, Berga, Herranz, Khan, and Weijer]{wang2020minegan}
Yaxing Wang, Abel Gonzalez-Garcia, David Berga, Luis Herranz, Fahad~Shahbaz Khan, and Joost van~de Weijer.
\newblock Minegan: Effective knowledge transfer from gans to target domains with few images.
\newblock In \emph{Proceedings of the IEEE/CVF Conference on Computer Vision and Pattern Recognition}, 2020{\natexlab{b}}.

\bibitem[Wang et~al.(2021)Wang, Gonzalez-Garcia, Wu, Herranz, Khan, Jui, and Van~de Weijer]{wang2021minegan++}
Yaxing Wang, Abel Gonzalez-Garcia, Chenshen Wu, Luis Herranz, Fahad~Shahbaz Khan, Shangling Jui, and Joost Van~de Weijer.
\newblock Minegan++: Mining generative models for efficient knowledge transfer to limited data domains.
\newblock \emph{arXiv preprint arXiv:2104.13742}, 2021.

\bibitem[Wang et~al.(2022{\natexlab{c}})Wang, Yi, Tai, Wang, and Ma]{wang2022ctlgan}
Yue Wang, Ran Yi, Ying Tai, Chengjie Wang, and Lizhuang Ma.
\newblock Ctlgan: Few-shot artistic portraits generation with contrastive transfer learning.
\newblock \emph{arXiv preprint arXiv:2203.08612}, 2022{\natexlab{c}}.

\bibitem[Wang et~al.(2023{\natexlab{c}})Wang, Jiang, Zheng, Wang, He, Wang, Chen, and Zhou]{wang2023patchdiffusion}
Zhendong Wang, Yifan Jiang, Huangjie Zheng, Peihao Wang, Pengcheng He, Zhangyang Wang, Weizhu Chen, and Mingyuan Zhou.
\newblock Patch diffusion: Faster and more data-efficient training of diffusion models.
\newblock \emph{arXiv preprint arXiv:2304.12526}, 2023{\natexlab{c}}.

\bibitem[Wang et~al.(2023{\natexlab{d}})Wang, Zheng, He, Chen, and Zhou]{wang2023diffusiongan}
Zhendong Wang, Huangjie Zheng, Pengcheng He, Weizhu Chen, and Mingyuan Zhou.
\newblock Diffusion-gan: Training gans with diffusion.
\newblock In \emph{International Conference on Learning Representations}, 2023{\natexlab{d}}.

\bibitem[Wei et~al.(2025{\natexlab{a}})Wei, Zeng, Li, Yin, Duan, and Li]{wei2025T2IRL}
Fanyue Wei, Wei Zeng, Zhenyang Li, Dawei Yin, Lixin Duan, and Wen Li.
\newblock Powerful and flexible: Personalized text-to-image generation via reinforcement learning.
\newblock In \emph{European Conference on Computer Vision}, pp.\  394--410. Springer, 2025{\natexlab{a}}.

\bibitem[Wei et~al.(2023)Wei, Zhang, Ji, Bai, Zhang, and Zuo]{Wei2023elite}
Yuxiang Wei, Yabo Zhang, Zhilong Ji, Jinfeng Bai, Lei Zhang, and Wangmeng Zuo.
\newblock Elite: Encoding visual concepts into textual embeddings for customized text-to-image generation.
\newblock In \emph{Proceedings of the IEEE/CVF International Conference on Computer Vision (ICCV)}, pp.\  15943--15953, October 2023.

\bibitem[Wei et~al.(2025{\natexlab{b}})Wei, Ji, Bai, Zhang, Zhang, and Zuo]{wei2025masterweaver}
Yuxiang Wei, Zhilong Ji, Jinfeng Bai, Hongzhi Zhang, Lei Zhang, and Wangmeng Zuo.
\newblock Masterweaver: Taming editability and face identity for personalized text-to-image generation.
\newblock In \emph{European Conference on Computer Vision}, pp.\  252--271. Springer, 2025{\natexlab{b}}.

\bibitem[Wen et~al.(2023)Wen, Kirchenbauer, Geiping, and Goldstein]{wen2023tree}
Yuxin Wen, John Kirchenbauer, Jonas Geiping, and Tom Goldstein.
\newblock Tree-ring watermarks: Fingerprints for diffusion images that are invisible and robust.
\newblock \emph{arXiv preprint arXiv:2305.20030}, 2023.

\bibitem[Whang et~al.(2023)Whang, Roh, Song, and Lee]{whang2023datacollection}
Steven~Euijong Whang, Yuji Roh, Hwanjun Song, and Jae-Gil Lee.
\newblock Data collection and quality challenges in deep learning: A data-centric ai perspective.
\newblock \emph{The VLDB Journal}, 32\penalty0 (4):\penalty0 791--813, 2023.

\bibitem[Wu et~al.(2023)Wu, Wang, Wu, and Li]{wu2023d3tgan}
Xintian Wu, Huanyu Wang, Yiming Wu, and Xi~Li.
\newblock D3t-gan: Data-dependent domain transfer gans for image generation with limited data.
\newblock \emph{ACM Transactions on Multimedia Computing, Communications and Applications}, 19\penalty0 (4):\penalty0 1--20, 2023.

\bibitem[Wu et~al.(2024)Wu, Li, Wang, Zheng, Zhao, Li, and Tao]{wu2024domain}
Yi~Wu, Ziqiang Li, Chaoyue Wang, Heliang Zheng, Shanshan Zhao, Bin Li, and Dacheng Tao.
\newblock Domain re-modulation for few-shot generative domain adaptation.
\newblock \emph{Advances in Neural Information Processing Systems}, 36, 2024.

\bibitem[Wu et~al.(2025)Wu, Huang, Wang, Wei, and Liu]{wu2025MultiGen}
Zhi-Fan Wu, Lianghua Huang, Wei Wang, Yanheng Wei, and Yu~Liu.
\newblock Multigen: Zero-shot image generation from multi-modal prompts.
\newblock In \emph{European Conference on Computer Vision}, pp.\  297--313. Springer, 2025.

\bibitem[Xiao et~al.(2024)Xiao, Yin, Freeman, Durand, and Han]{xiao2024fastcomposer}
Guangxuan Xiao, Tianwei Yin, William~T Freeman, Fr{\'e}do Durand, and Song Han.
\newblock Fastcomposer: Tuning-free multi-subject image generation with localized attention.
\newblock \emph{International Journal of Computer Vision}, pp.\  1--20, 2024.

\bibitem[Xiao et~al.(2022)Xiao, Li, Wang, Zha, and Huang]{xiao2022rssa}
Jiayu Xiao, Liang Li, Chaofei Wang, Zheng-Jun Zha, and Qingming Huang.
\newblock Few shot generative model adaption via relaxed spatial structural alignment.
\newblock In \emph{Proceedings of the IEEE/CVF Conference on Computer Vision and Pattern Recognition}, 2022.

\bibitem[Xiao et~al.(2025)Xiao, Cai, Guan, and Wang]{xiao2025semantic}
Ting Xiao, Yunjie Cai, Jiaoyan Guan, and Zhe Wang.
\newblock Semantic mask reconstruction and category semantic learning for few-shot image generation.
\newblock \emph{Neural Networks}, 183:\penalty0 106946, 2025.

\bibitem[Xie et~al.(2022)Xie, Fu, Tai, Cao, Zhu, and Wang]{xie2022learning}
Yu~Xie, Yanwei Fu, Ying Tai, Yun Cao, Junwei Zhu, and Chengjie Wang.
\newblock Learning to memorize feature hallucination for one-shot image generation.
\newblock In \emph{Proceedings of the IEEE/CVF Conference on Computer Vision and Pattern Recognition}, pp.\  9130--9139, 2022.

\bibitem[Yan et~al.(2024)Yan, Yan, Chai, Geng, Zhou, and Gao]{yan2024dm}
Longquan Yan, Ruixiang Yan, Bosong Chai, Guohua Geng, Pengbo Zhou, and Jian Gao.
\newblock Dm-gan: Cnn hybrid vits for training gans under limited data.
\newblock \emph{Pattern Recognition}, 156:\penalty0 110810, 2024.

\bibitem[Yang et~al.(2021{\natexlab{a}})Yang, Shen, Xu, and Zhou]{yang2021insgen}
Ceyuan Yang, Yujun Shen, Yinghao Xu, and Bolei Zhou.
\newblock Data-efficient instance generation from instance discrimination.
\newblock In \emph{Advances in Neural Information Processing Systems}, 2021{\natexlab{a}}.

\bibitem[Yang et~al.(2021{\natexlab{b}})Yang, Shen, Zhang, Xu, Zhu, Wu, and Zhou]{yang2021genda}
Ceyuan Yang, Yujun Shen, Zhiyi Zhang, Yinghao Xu, Jiapeng Zhu, Zhirong Wu, and Bolei Zhou.
\newblock One-shot generative domain adaptation.
\newblock \emph{arXiv preprint arXiv:2111.09876}, 2021{\natexlab{b}}.

\bibitem[Yang et~al.(2022{\natexlab{a}})Yang, Shen, Xu, Zhao, Dai, and Zhou]{yang2022dynamicd}
Ceyuan Yang, Yujun Shen, Yinghao Xu, Deli Zhao, Bo~Dai, and Bolei Zhou.
\newblock Improving gans with a dynamic discriminator.
\newblock In \emph{Advances in Neural Information Processing Systems}, 2022{\natexlab{a}}.

\bibitem[Yang et~al.(2022{\natexlab{b}})Yang, Wang, Chi, and Feng]{yang2022wavegan}
Mengping Yang, Zhe Wang, Ziqiu Chi, and Wenyi Feng.
\newblock Wavegan: Frequency-aware gan for high-fidelity few-shot image generation.
\newblock In \emph{Proceedings of the European Conference on Computer Vision}, 2022{\natexlab{b}}.

\bibitem[Yang et~al.(2022{\natexlab{c}})Yang, Wang, Chi, and Zhang]{yang2022fregan}
Mengping Yang, Zhe Wang, Ziqiu Chi, and Yanbing Zhang.
\newblock Fregan: Exploiting frequency components for training gans under limited data.
\newblock In \emph{Advances in Neural Information Processing Systems}, 2022{\natexlab{c}}.

\bibitem[Yang et~al.(2023{\natexlab{a}})Yang, Niu, Wang, Li, and Du]{yang2023dfsgan}
Mengping Yang, Saisai Niu, Zhe Wang, Dongdong Li, and Wenli Du.
\newblock Dfsgan: Introducing editable and representative attributes for few-shot image generation.
\newblock \emph{Engineering Applications of Artificial Intelligence}, 117:\penalty0 105519, 2023{\natexlab{a}}.

\bibitem[Yang et~al.(2023{\natexlab{b}})Yang, Wang, Chi, and Du]{yang2023protogan}
Mengping Yang, Zhe Wang, Ziqiu Chi, and Wenli Du.
\newblock Protogan: Towards high diversity and fidelity image synthesis under limited data.
\newblock \emph{Information Sciences}, 632:\penalty0 698--714, 2023{\natexlab{b}}.

\bibitem[Yang et~al.(2023{\natexlab{c}})Yang, Wang, Feng, Zhang, and Xiao]{yang2023sdtm}
Mengping Yang, Zhe Wang, Wenyi Feng, Qian Zhang, and Ting Xiao.
\newblock Improving few-shot image generation by structural discrimination and textural modulation.
\newblock In \emph{Proceedings of the 31st ACM International Conference on Multimedia}, pp.\  7837--7848, 2023{\natexlab{c}}.

\bibitem[Yang et~al.(2024)Yang, Peng, Kong, Zhang, Yao, and Jin]{yang2024fontdiffuser}
Zhenhua Yang, Dezhi Peng, Yuxin Kong, Yuyi Zhang, Cong Yao, and Lianwen Jin.
\newblock Fontdiffuser: One-shot font generation via denoising diffusion with multi-scale content aggregation and style contrastive learning.
\newblock In \emph{Proceedings of the AAAI conference on artificial intelligence}, 2024.

\bibitem[Yaniv et~al.(2019)Yaniv, Newman, and Shamir]{yaniv2019face}
Jordan Yaniv, Yael Newman, and Ariel Shamir.
\newblock The face of art: landmark detection and geometric style in portraits.
\newblock \emph{ACM Transactions on Graphics}, 38\penalty0 (4):\penalty0 1--15, 2019.

\bibitem[Yildiz et~al.(2024)Yildiz, Yuksel, and Sevgen]{yildiz2024sdsgan}
Eyyup Yildiz, Mehmet~Erkan Yuksel, and Selcuk Sevgen.
\newblock A single-image gan model using self-attention mechanism and densenets.
\newblock \emph{Neurocomputing}, pp.\  127873, 2024.

\bibitem[Yosinski et~al.(2014)Yosinski, Clune, Bengio, and Lipson]{yosinski2014transferable}
Jason Yosinski, Jeff Clune, Yoshua Bengio, and Hod Lipson.
\newblock How transferable are features in deep neural networks?
\newblock In \emph{Advances in Neural Information Processing Systems}, 2014.

\bibitem[Yu et~al.(2022)Yu, Han, Shen, Yu, Gong, Gong, and Liu]{yu2022understanding}
Chaojian Yu, Bo~Han, Li~Shen, Jun Yu, Chen Gong, Mingming Gong, and Tongliang Liu.
\newblock Understanding robust overfitting of adversarial training and beyond.
\newblock In \emph{Proceedings of the International Conference on Machine Learning}, 2022.

\bibitem[Yu et~al.(2015)Yu, Seff, Zhang, Song, Funkhouser, and Xiao]{yu2015lsun}
Fisher Yu, Ari Seff, Yinda Zhang, Shuran Song, Thomas Funkhouser, and Jianxiong Xiao.
\newblock Lsun: Construction of a large-scale image dataset using deep learning with humans in the loop.
\newblock \emph{arXiv preprint arXiv:1506.03365}, 2015.

\bibitem[Yu et~al.(2017)Yu, Zhang, Wang, and Yu]{yu2017seqgan}
Lantao Yu, Weinan Zhang, Jun Wang, and Yong Yu.
\newblock Seqgan: Sequence generative adversarial nets with policy gradient.
\newblock In \emph{Proceedings of the AAAI Conference on Artificial Intelligence}, 2017.

\bibitem[Yu et~al.(2024)Yu, Zhu, Cao, Xia, and Kang]{yu2024tf}
Qianzi Yu, Kai Zhu, Yang Cao, Feijie Xia, and Yu~Kang.
\newblock Tf 2: Few-shot text-free training-free defect image generation for industrial anomaly inspection.
\newblock \emph{IEEE Transactions on Circuits and Systems for Video Technology}, 2024.

\bibitem[Zeng \& Xiao(2024)Zeng and Xiao]{zeng2024few}
Wu~Zeng and Zheng-ying Xiao.
\newblock Few-shot learning based on deep learning: A survey.
\newblock \emph{Mathematical Biosciences and Engineering}, 21\penalty0 (1):\penalty0 679--711, 2024.

\bibitem[Zhai et~al.(2018)Zhai, Zhang, Chen, and He]{zhai2018autoencoder}
Junhai Zhai, Sufang Zhang, Junfen Chen, and Qiang He.
\newblock Autoencoder and its various variants.
\newblock In \emph{2018 IEEE international conference on systems, man, and cybernetics (SMC)}, pp.\  415--419. IEEE, 2018.

\bibitem[Zhai et~al.(2024)Zhai, Long, Pan, and Chen]{zhai2024micgan}
YiKui Zhai, ZhiHao Long, WenFeng Pan, and CL~Philip Chen.
\newblock Mutual information compensation for high-fidelity image generation with limited data.
\newblock \emph{IEEE Signal Processing Letters}, 2024.

\bibitem[Zhang et~al.(2023{\natexlab{a}})Zhang, Chen, Chai, Wu, Lagun, Beeler, and De~la Torre]{zhang2023itigen}
Cheng Zhang, Xuanbai Chen, Siqi Chai, Chen~Henry Wu, Dmitry Lagun, Thabo Beeler, and Fernando De~la Torre.
\newblock Iti-gen: Inclusive text-to-image generation.
\newblock In \emph{Proceedings of the IEEE/CVF International Conference on Computer Vision}, pp.\  3969--3980, 2023{\natexlab{a}}.

\bibitem[Zhang et~al.(2020)Zhang, Zhang, Odena, and Lee]{Zhang2020Consistency}
Han Zhang, Zizhao Zhang, Augustus Odena, and Honglak Lee.
\newblock Consistency regularization for generative adversarial networks.
\newblock In \emph{International Conference on Learning Representations}, 2020.

\bibitem[Zhang et~al.(2023{\natexlab{b}})Zhang, Xie, Sun, Huang, Wang, Lei, and Chen]{zhang2023clcr}
Jing Zhang, Yingpeng Xie, Dandan Sun, Ruidong Huang, Tianfu Wang, Baiying Lei, and Kuntao Chen.
\newblock Multi-national covid-19 ct image-label pairs synthesis via few-shot gans adaptation.
\newblock In \emph{2023 IEEE 20th International Symposium on Biomedical Imaging (ISBI)}, pp.\  1--4. IEEE, 2023{\natexlab{b}}.

\bibitem[Zhang et~al.(2018)Zhang, Isola, Efros, Shechtman, and Wang]{zhang2018lpips}
Richard Zhang, Phillip Isola, Alexei~A Efros, Eli Shechtman, and Oliver Wang.
\newblock The unreasonable effectiveness of deep features as a perceptual metric.
\newblock In \emph{Proceedings of the IEEE/CVF Conference on Computer Vision and Pattern Recognition}, 2018.

\bibitem[Zhang et~al.(2024{\natexlab{a}})Zhang, Huang, Huo, Yuan, Zhou, Chen, and Zhong]{zhang2024tage}
Ruicheng Zhang, Guoheng Huang, Yejing Huo, Xiaochen Yuan, Zhizhen Zhou, Xuhang Chen, and Guo Zhong.
\newblock Tage: Trustworthy attribute group editing for stable few-shot image generation.
\newblock \emph{arXiv preprint arXiv:2410.17855}, 2024{\natexlab{a}}.

\bibitem[Zhang et~al.(2023{\natexlab{c}})Zhang, Liu, Li, Xie, Huang, Li, Zheng, and Ghanem]{zhang2023dmd}
Wentian Zhang, Haozhe Liu, Bing Li, Jinheng Xie, Yawen Huang, Yuexiang Li, Yefeng Zheng, and Bernard Ghanem.
\newblock Dynamically masked discriminator for generative adversarial networks.
\newblock \emph{arXiv preprint arXiv:2306.07716}, 2023{\natexlab{c}}.

\bibitem[Zhang et~al.(2022{\natexlab{a}})Zhang, Wei, Ji, Bai, Zuo, et~al.]{zhang2022difa}
Yabo Zhang, Yuxiang Wei, Zhilong Ji, Jinfeng Bai, Wangmeng Zuo, et~al.
\newblock Towards diverse and faithful one-shot adaption of generative adversarial networks.
\newblock In \emph{Advances in Neural Information Processing Systems}, 2022{\natexlab{a}}.

\bibitem[Zhang et~al.(2024{\natexlab{b}})Zhang, Yang, Zhou, and Wang]{zhang2024disendiff}
Yanbing Zhang, Mengping Yang, Qin Zhou, and Zhe Wang.
\newblock Attention calibration for disentangled text-to-image personalization.
\newblock In \emph{Proceedings of the IEEE/CVF Conference on Computer Vision and Pattern Recognition}, pp.\  4764--4774, 2024{\natexlab{b}}.

\bibitem[Zhang et~al.(2023{\natexlab{d}})Zhang, Dong, Tang, Huang, Huang, Ma, Lee, Deussen, and Xu]{zhang2023prospect}
Yuxin Zhang, Weiming Dong, Fan Tang, Nisha Huang, Haibin Huang, Chongyang Ma, Tong-Yee Lee, Oliver Deussen, and Changsheng Xu.
\newblock Prospect: Prompt spectrum for attribute-aware personalization of diffusion models.
\newblock \emph{ACM Transactions on Graphics (TOG)}, 42\penalty0 (6):\penalty0 244:1--244:14, 2023{\natexlab{d}}.

\bibitem[Zhang et~al.(2024{\natexlab{c}})Zhang, Song, Liu, Wang, Yu, Tang, Li, Tang, Hu, Pan, et~al.]{zhang2024ssr}
Yuxuan Zhang, Yiren Song, Jiaming Liu, Rui Wang, Jinpeng Yu, Hao Tang, Huaxia Li, Xu~Tang, Yao Hu, Han Pan, et~al.
\newblock Ssr-encoder: Encoding selective subject representation for subject-driven generation.
\newblock In \emph{Proceedings of the IEEE/CVF Conference on Computer Vision and Pattern Recognition}, pp.\  8069--8078, 2024{\natexlab{c}}.

\bibitem[Zhang et~al.(2023{\natexlab{e}})Zhang, Li, Hong, Wang, Ma, Xiong, and Xu]{zhang2023fptgan}
Zeren Zhang, Xingjian Li, Tao Hong, Tianyang Wang, Jinwen Ma, Haoyi Xiong, and Cheng-Zhong Xu.
\newblock Overcoming catastrophic forgetting for fine-tuning pre-trained gans.
\newblock In \emph{Joint European Conference on Machine Learning and Knowledge Discovery in Databases}, pp.\  293--308. Springer, 2023{\natexlab{e}}.

\bibitem[Zhang et~al.(2024{\natexlab{d}})Zhang, Hua, Sun, Wang, and McLoone]{zhang2024anda}
Zhaoyu Zhang, Yang Hua, Guanxiong Sun, Hui Wang, and Se{\'a}n McLoone.
\newblock Improving the leaking of augmentations in data-efficient gans via adaptive negative data augmentation.
\newblock In \emph{Proceedings of the IEEE/CVF Winter Conference on Applications of Computer Vision}, pp.\  5412--5421, 2024{\natexlab{d}}.

\bibitem[Zhang et~al.(2024{\natexlab{e}})Zhang, Hua, Sun, Wang, and McLoone]{zhang2024dani}
Zhaoyu Zhang, Yang Hua, Guanxiong Sun, Hui Wang, and Se\'{a}n McLoone.
\newblock Improving the training of the gans with limited data via dual adaptive noise injection.
\newblock In \emph{Proceedings of the 32nd ACM International Conference on Multimedia}, 2024{\natexlab{e}}.

\bibitem[Zhang et~al.(2023{\natexlab{f}})Zhang, Han, Ghosh, Metaxas, and Ren]{zhang2023sine}
Zhixing Zhang, Ligong Han, Arnab Ghosh, Dimitris~N Metaxas, and Jian Ren.
\newblock Sine: Single image editing with text-to-image diffusion models.
\newblock In \emph{Proceedings of the IEEE/CVF Conference on Computer Vision and Pattern Recognition}, 2023{\natexlab{f}}.

\bibitem[Zhang et~al.(2021)Zhang, Han, and Guo]{zhang2021exsingan}
ZiCheng Zhang, CongYing Han, and TianDe Guo.
\newblock Exsingan: Learning an explainable generative model from a single image.
\newblock In \emph{32nd British Machine Vision Conference}, 2021.

\bibitem[Zhang et~al.(2022{\natexlab{b}})Zhang, Liu, Han, Guo, Yao, and Mei]{zhang2022generalizedoneshot}
Zicheng Zhang, Yinglu Liu, Congying Han, Tiande Guo, Ting Yao, and Tao Mei.
\newblock Generalized one-shot domain adaptation of generative adversarial networks.
\newblock In \emph{Advances in Neural Information Processing Systems}, 2022{\natexlab{b}}.

\bibitem[Zhang et~al.(2022{\natexlab{c}})Zhang, Liu, Han, Shi, Guo, and Zhou]{zhang2022petsgan}
Zicheng Zhang, Yinglu Liu, Congying Han, Hailin Shi, Tiande Guo, and Bowen Zhou.
\newblock Petsgan: Rethinking priors for single image generation.
\newblock In \emph{Proceedings of the AAAI Conference on Artificial Intelligence}, volume~36, pp.\  3408--3416, 2022{\natexlab{c}}.

\bibitem[Zhang et~al.(2024{\natexlab{f}})Zhang, Pan, Wei, Ji, Yang, and Deng]{zhang2024few}
Ziqi Zhang, Siduo Pan, Kun Wei, Jiapeng Ji, Xu~Yang, and Cheng Deng.
\newblock Few-shot generative model adaption via optimal kernel modulation.
\newblock \emph{IEEE Transactions on Circuits and Systems for Video Technology}, 2024{\natexlab{f}}.

\bibitem[Zhao et~al.(2020{\natexlab{a}})Zhao, Cong, and Carin]{zhao2020leveraging}
Miaoyun Zhao, Yulai Cong, and Lawrence Carin.
\newblock On leveraging pretrained gans for generation with limited data.
\newblock In \emph{Proceedings of the International Conference on Machine Learning}, 2020{\natexlab{a}}.

\bibitem[Zhao et~al.(2020{\natexlab{b}})Zhao, Liu, Lin, Zhu, and Han]{zhao2020diffaug}
Shengyu Zhao, Zhijian Liu, Ji~Lin, Jun-Yan Zhu, and Song Han.
\newblock Differentiable augmentation for data-efficient gan training.
\newblock In \emph{Advances in Neural Information Processing Systems}, 2020{\natexlab{b}}.

\bibitem[Zhao et~al.(2022{\natexlab{a}})Zhao, Chandrasegaran, Abdollahzadeh, and Cheung]{zhao2022adam}
Yunqing Zhao, Keshigeyan Chandrasegaran, Milad Abdollahzadeh, and Ngai-Man~Man Cheung.
\newblock Few-shot image generation via adaptation-aware kernel modulation.
\newblock In \emph{Advances in Neural Information Processing Systems}, 2022{\natexlab{a}}.

\bibitem[Zhao et~al.(2022{\natexlab{b}})Zhao, Ding, Huang, and Cheung]{zhao2022dcl}
Yunqing Zhao, Henghui Ding, Houjing Huang, and Ngai-Man Cheung.
\newblock A closer look at few-shot image generation.
\newblock In \emph{Proceedings of the IEEE/CVF Conference on Computer Vision and Pattern Recognition}, 2022{\natexlab{b}}.

\bibitem[Zhao et~al.(2023{\natexlab{a}})Zhao, Du, Abdollahzadeh, Pang, Lin, Yan, and Cheung]{zhao2023rick}
Yunqing Zhao, Chao Du, Milad Abdollahzadeh, Tianyu Pang, Min Lin, Shuicheng Yan, and Ngai-Man Cheung.
\newblock Exploring incompatible knowledge transfer in few-shot image generation.
\newblock In \emph{Proceedings of the IEEE/CVF Conference on Computer Vision and Pattern Recognition}, 2023{\natexlab{a}}.

\bibitem[Zhao et~al.(2023{\natexlab{b}})Zhao, Pang, Du, Yang, Cheung, and Lin]{zhao2023recipe}
Yunqing Zhao, Tianyu Pang, Chao Du, Xiao Yang, Ngai-Man Cheung, and Min Lin.
\newblock A recipe for watermarking diffusion models.
\newblock \emph{arXiv preprint arXiv:2303.10137}, 2023{\natexlab{b}}.

\bibitem[Zhao et~al.(2020{\natexlab{c}})Zhao, Zhang, Chen, Singh, and Zhang]{zhao2020imageaugmentation}
Zhengli Zhao, Zizhao Zhang, Ting Chen, Sameer Singh, and Han Zhang.
\newblock Image augmentations for gan training.
\newblock \emph{arXiv preprint arXiv:2006.02595}, 2020{\natexlab{c}}.

\bibitem[Zhao et~al.(2021)Zhao, Singh, Lee, Zhang, Odena, and Zhang]{zhao2021improved}
Zhengli Zhao, Sameer Singh, Honglak Lee, Zizhao Zhang, Augustus Odena, and Han Zhang.
\newblock Improved consistency regularization for gans.
\newblock In \emph{Proceedings of the AAAI Conference on Artificial Intelligence}, 2021.

\bibitem[Zheng et~al.(2023)Zheng, Liu, Zhang, Xu, and He]{zheng2023lso}
Chenxi Zheng, Bangzhen Liu, Huaidong Zhang, Xuemiao Xu, and Shengfeng He.
\newblock Where is my spot? few-shot image generation via latent subspace optimization.
\newblock In \emph{Proceedings of the IEEE/CVF Conference on Computer Vision and Pattern Recognition}, 2023.

\bibitem[Zhou et~al.(2017)Zhou, Lapedriza, Khosla, Oliva, and Torralba]{zhou2017places}
Bolei Zhou, Agata Lapedriza, Aditya Khosla, Aude Oliva, and Antonio Torralba.
\newblock Places: A 10 million image database for scene recognition.
\newblock \emph{IEEE Transactions on Pattern Analysis and Machine Intelligence}, 40\penalty0 (6):\penalty0 1452--1464, 2017.

\bibitem[Zhou et~al.(2024{\natexlab{a}})Zhou, Chen, and Huang]{zhou2024deformable}
Yang Zhou, Zichong Chen, and Hui Huang.
\newblock Deformable one-shot face stylization via dino semantic guidance.
\newblock In \emph{Proceedings of the IEEE/CVF Conference on Computer Vision and Pattern Recognition}, pp.\  7787--7796, 2024{\natexlab{a}}.

\bibitem[Zhou et~al.(2023)Zhou, Yue, Ye, Zhang, Wei, and Chen]{zhou2023eqgan}
Yingbo Zhou, Zhihao Yue, Yutong Ye, Pengyu Zhang, Xian Wei, and Mingsong Chen.
\newblock Eqgan: Feature equalization fusion for few-shot image generation.
\newblock \emph{arXiv preprint arXiv:2307.14638}, 2023.

\bibitem[Zhou et~al.(2024{\natexlab{b}})Zhou, Ye, Zhang, Wei, and Chen]{zhou2024f2dgan}
Yingbo Zhou, Yutong Ye, Pengyu Zhang, Xian Wei, and Mingsong Chen.
\newblock Exact fusion via feature distribution matching for few-shot image generation.
\newblock In \emph{Proceedings of the IEEE/CVF Conference on Computer Vision and Pattern Recognition}, pp.\  8383--8392, 2024{\natexlab{b}}.

\bibitem[Zhu et~al.(2022{\natexlab{a}})Zhu, Ma, Chen, and Yuan]{zhu2022few}
Jingyuan Zhu, Huimin Ma, Jiansheng Chen, and Jian Yuan.
\newblock Few-shot image generation via masked discrimination.
\newblock \emph{arXiv preprint arXiv:2210.15194}, 2022{\natexlab{a}}.

\bibitem[Zhu et~al.(2022{\natexlab{b}})Zhu, Ma, Chen, and Yuan]{zhu2022few_dm}
Jingyuan Zhu, Huimin Ma, Jiansheng Chen, and Jian Yuan.
\newblock Few-shot image generation with diffusion models.
\newblock \emph{arXiv preprint arXiv:2211.03264}, 2022{\natexlab{b}}.

\bibitem[Zhu et~al.(2023)Zhu, Ma, Chen, and Yuan]{zhu2023fewshot_3d}
Jingyuan Zhu, Huimin Ma, Jiansheng Chen, and Jian Yuan.
\newblock Few-shot 3d shape generation.
\newblock \emph{arXiv preprint arXiv:2305.11664}, 2023.

\bibitem[Zhu et~al.(2025)Zhu, Chen, Wang, Zhao, and Jia]{zhu2025logosticker}
Mingkang Zhu, Xi~Chen, Zhongdao Wang, Hengshuang Zhao, and Jiaya Jia.
\newblock Logosticker: Inserting logos into diffusion models for customized generation.
\newblock In \emph{European Conference on Computer Vision}, pp.\  363--378. Springer, 2025.

\bibitem[Zhu et~al.(2022{\natexlab{c}})Zhu, Abdal, Femiani, and Wonka]{zhu2022mindthegap}
Peihao Zhu, Rameen Abdal, John Femiani, and Peter Wonka.
\newblock Mind the gap: Domain gap control for single shot domain adaptation for generative adversarial networks.
\newblock In \emph{International Conference on Learning Representations}, 2022{\natexlab{c}}.

\bibitem[Zhu \& Yang(2024)Zhu and Yang]{zhu2024enhancing}
Yan-Lin Zhu and Peipei Yang.
\newblock Enhancing fine-tuning performance of text-to-image diffusion models for few-shot image generation through contrastive learning.
\newblock In \emph{Chinese Conference on Image and Graphics Technologies}, pp.\  133--147. Springer, 2024.

\end{thebibliography}
\bibliographystyle{tmlr}


\end{document}